\documentclass[journal]{IEEEtran}
\usepackage{graphicx}
\usepackage{bm}
\usepackage{amsmath}
\usepackage{algorithm}
\usepackage{algorithmic}
\usepackage{color}
\usepackage{amsfonts}
\usepackage{amsmath}
\usepackage{multirow}
\usepackage{threeparttable} 
\usepackage{chngpage}
\usepackage{epstopdf}
\usepackage{epsfig}
\usepackage{algorithmic}
\usepackage{Extpfeil}
\usepackage[table]{xcolor}
\usepackage{fixltx2e}
\usepackage{booktabs}

\begin{document}
%\title{Progressive Ensemble Symmetric Non-negative Matrix Factorization with Feedback}
%\title{Progressive Ensemble Symmetric Non-negative Matrix Factorization}
\title{Self-supervised  Symmetric Nonnegative Matrix Factorization}

\author{Yuheng~Jia,
	Hui~Liu,
	~Junhui~Hou,~\IEEEmembership{Senior Member,~IEEE,}
	Sam~Kwong,~\IEEEmembership{Fellow,~IEEE},~and
	Qingfu~Zhang,~\IEEEmembership{Fellow,~IEEE} 
%	\thanks{This work was supported in part by the Natural Science Foundation of China
%		under Grants 61871342, 61772344, 61672443  and in part by Hong Kong RGC General Research Funds 9042820 (CityU 11219019),
%		9042489 (CityU 11206317), 9042322 (CityU 11200116), and 9048123 (CityU 21211518). (Corresponding authors:  Junhui Hou and Sam Kwong.)}
	\thanks{Y. Jia is with the School of Computer Science and Engineering, Southeast University, Nanjing, (e-mail: yhjia@seu.edu.cn).}
	\thanks{H. Liu is with the Department
		of Computer Science, City University of Hong Kong, Kowloon, Hong Kong,
		(e-mail: hliu99-c@my.cityu.edu.hk).} \thanks{J. Hou, S. Kwong and Q. Zhang are with the Department of Computer Science,
		City University of Hong Kong, Kowloon, Hong Kong and also with the City University of Hong Kong Shenzhen Research Institute, Shenzhen, 51800, China, (e-mail: jh.hou@cityu.edu.hk; cssamk@cityu.edu.hk; qingfu.zhang@cityu.edu.hk).}}
\maketitle

\begin{abstract}
	Symmetric nonnegative matrix factorization (SNMF) has demonstrated to be a powerful method for data clustering. However, SNMF is mathematically formulated as a  non-convex optimization problem, making it sensitive to the initialization of variables.
	Inspired by ensemble clustering that aims to seek  a better clustering result from a set of clustering results, we propose self-supervised SNMF (S$^3$NMF), which is capable of boosting  clustering performance progressively by taking advantage of the sensitivity to initialization characteristic of SNMF, without relying on any additional information.
	Specifically, we first perform SNMF  repeatedly  with a random nonnegative matrix for initialization each time, leading to multiple decomposed matrices. Then, we rank the quality of the resulting matrices  with adaptively learned  weights, from which a  new similarity matrix 
	that is expected to be more discriminative  is reconstructed 
	for SNMF again. 
	These two steps are iterated  until the  stopping criterion/maximum number of iterations is achieved. We mathematically formulate S$^3$NMF as a constraint  optimization problem, and provide an alternative optimization algorithm to solve it  with the theoretical convergence guaranteed. 
	Extensive experimental results on $10$ commonly used benchmark datasets demonstrate the significant advantage of our S$^3$NMF  over $12$ state-of-the-art methods in terms of $5$ quantitative metrics. 
	The source code is publicly available at  \textcolor{magenta}{https://github.com/jyh-learning/SSSNMF}.
\end{abstract}

% Note that keywords are not normally used for peerreview papers.
\IEEEpeerreviewmaketitle

\section{Introduction}
Nonnegative matrix factorization (NMF) \cite{lee2001algorithms,lee1999learning,9208729} is a well-known  dimensionality reduction method for data representation. Technically,  it decomposes a nonnegative matrix as the product  of two smaller nonnegative matrices named the basis matrix and  the embedding matrix. Due to the nonnegative constraints on both the input matrix and the factorized matrices, NMF is able to learn a parts-based representation. Taking face images as an example, the basis matrix contains the meaningful parts of the input face, e.g., a nose or an eye, and the embedding matrix assembles the bases to recover the face. Such a unique feature enables the success of NMF in many  applications, e.g., topic modeling \cite{choo2013utopian}, hyperspectral image unmixing \cite{bioucas2012hyperspectral},  blind audio source separation \cite{9084229}, image classification \cite{8027109},  visual tracking \cite{7914620,6579656}, etc. See the comprehensive review of NMF in \cite{liu2017regularized} and \cite{6165290}.

In addition to data representation, NMF has been used to solve various clustering problems \cite{li2018nonnegative}, e.g., tumor clustering \cite{zheng2009tumor}, community detection \cite{wu2018nonnegative}, document clustering \cite{cai2010locally}, and so on. Particularly, Ding et al. \cite{ding2005equivalence} revealed the equivalence between NMF and K-means.  However,  NMF is only effective for partitioning linearly separable data and usually cannot exploit the non-linear relationship of the input  \cite{7167693}. 
To solve this drawback, symmetric NMF (SNMF) was proposed \cite{kuang2012symmetric,kuang2015symnmf}. Different from NMF that factorizes the sample matrix into two different matrices, SNMF decomposes an affinity matrix that records the pairwise similarity of samples as the product of a nonnegative matrix and its transpose. Kuang et al. \cite{kuang2015symnmf} showed  that SNMF is related to spectral clustering (SC) \cite{ng2002spectral}, and both of them share the same loss function only with different constraints. 
Therefore, SNMF can be regarded as a graph clustering method \cite{kuang2015symnmf}, and it is more effective for nonlinearly separable data than NMF.  
Another merit of SNMF is that it can directly generate the clustering indicator without post-processing, while SC needs extra post-processing like K-means to finalize clustering. 
%Those merits widen the application domains of SNMF in clustering. 

A number of advanced SNMF methods were proposed to enhance the clustering ability of SNMF. For example, to encode the geometry  structure of data samples, Gao et al. \cite{gao2018graph} proposed to use a local graph to regularize the decomposition of SNMF. Considering  the input affinity matrix affects the performance of SNMF severely, Jia et al. \cite{9072553} proposed to learn an adaptive affinity matrix from data samples rather than use an empirically defined one, and  perform SNMF simultaneously. 
Zhang et al. \cite{zhang2016multi} extended SNMF to solve the multi-view clustering problem. 
Besides, many semi-supervised SNMF models were proposed to encode the available supervisory information. 
For example, Yang et al. \cite{6985550} proposed to use the supervisory information to regularize SNMF. 
Wu et al. \cite{8361078} took advantage of the supervisory information by using them to construct a superior affinity matrix.  
Although different kinds of optimization methods were proposed, e.g., Newton method \cite{kuang2015symnmf},  alternating nonnegative least squares \cite{zhu2018dropping}, and multiplicative update \cite{9013063}, and block coordinate descent \cite{IBSymNMF-2017-TSP}, etc., 
SNMF is essentially a non-convex optimization problem, and its sensitivity to initialization is inevitable,  i.e., the clustering performance of SNMF heavily relies on the initialization, and a bad initialization matrix  will degrade clustering performance significantly.

Motivated by ensemble clustering \cite{li2017structured,9072553}, i.e., multiple clustering results obtained by  different kinds of clustering methods could generate a better one, 
we propose self-supervised SNMF (S$^3$NMF), which is able to  boost clustering performance progressively by taking advantage of the sensitivity to initialization characteristic of SNMF without relying on any additional information.
%Different from the existing SNMF-based methods that use a single random matrix to initialize the variable, we adopt a group of random initialization matrices, 
Specifically, we perform SNMF  repeatedly  with a random nonnegative matrix for initialization each time, 
leading to multiple  decomposed matrices.   
%\textcolor{blue}{Motivated by the underlying spirit of ensemble clustering \cite{li2017structured,9072553}, i.e., multiple clustering partitions could generate a better one,} 
Then, we rank  the quality of those matrices adaptively, and form a new
affinity matrix from their binarizations (named clustering partition) under the guidance of their ranking.
As the  clustering partition of an affinity matrix is usually more
discriminative than the relationships between samples encoded in the affinity matrix (because an imperfect affinity matrix could produce a perfect partition  \cite{9072553,li2017structured}),
it is  expected that the newly constructed affinity matrix is superior to the original one.
Then, we further learn a set of new clustering partitions from the newly  constructed affinity matrix via SNMF. 
Such two steps  are repeated until the proposed stopping criterion is satisfied. During the iteration, the clustering performance grows progressively. Mathematically, we explicitly cast S$^3$NMF as a constrained optimization problem, and provide an efficient and effective algorithm  to solve it with the theoretical convergence guaranteed. Moreover, we give empirical values for the hyper-parameters, making it easy to use. We compare the proposed model with  12 state-of-the-art methods on 10 datasets in terms of 5 clustering metrics. The extensive experimental comparisons substantiate the superior performance of the proposed model, and also validates the effectiveness of the stopping criterion as well as the empirical values.

The rest of this paper is organized as follows. We first introduce the background of NMF and SNMF in Section II. In Section III, we present the proposed model and its numerical solution as well as the theoretical analyses.
In Section IV, we experimentally validate the advantages of the proposed method. 
Finally, we conclude this paper in Section V. 

\section{Related Work}
In this section, we first introduce NMF and the basic SNMF, followed by advanced SNMF-based mehtods.

\subsection{Nonnegative Matrix Factorization}
Let $\mathbf{X}=[\mathbf{x}_1,\cdots,\mathbf{x}_n]\in\mathbb{R}^{d\times n}$ denote the input nonnegative  matrix with $n$ samples, where $\mathbf{x}_i\in\mathbb{R}^{d}$ is the vectorial representation of the $i$-th sample of dimension  $d$. 
As a powerful and popular dimensionality reduction method,   
NMF factorizes $\mathbf{X}$ as the product of  two smaller nonnegative  matrices \cite{7428887}, i.e.,
\begin{equation}
\begin{split}
\min_{\mathbf{U},\mathbf{V}}\left\|\mathbf{X}-\mathbf{U}\mathbf{V}^\mathsf{T}\right\|_F^2,{\rm s.t.~}\mathbf{U}\geq 0,\mathbf{V}\geq 0, 
\end{split}
\label{NMF}
\end{equation}
where $\mathbf{U}=[\mathbf{u}_1,\cdots,\mathbf{u}_k]\in\mathbb{R}^{d\times k}$ is the basis matrix, $\mathbf{V}=[\mathbf{v}_1,\cdots,\mathbf{v}_n]^\mathsf{T}\in\mathbb{R}^{n\times k}$ is the low-dimensional representation, $k$ is the dimension of the new representation, $\mathbf{U}\geq 0,\mathbf{V}\geq 0$ mean that each element of $\mathbf{U}$ and $\mathbf{V}$ is nonnegative, and $\|\cdot\|_F$ and $\cdot^\mathsf{T}$ indicate the Frobenius norm and the transpose of a matrix, respectively. 
The conventional NMF requires the input matrix $\mathbf{X}$ to be  nonnegative. This assumption is reasonable for many real-world signals as they can be naturally represented with nonnegative values, such as pixels of an image, the acoustic signals, and the spectral signature of hyperspectral images. Together with the fact that the outputs $\mathbf{U}$ and $\mathbf{V}$ are also nonnegative, NMF only allows additive operations, i.e., each sample $\mathbf{x}_i$ is reconstructed by the conical combination of columns of $\mathbf{U}$, where $\mathbf{v}_i$ is the combination coefficient vector. This means that each column of $\mathbf{U}$ should only be part of $\mathbf{X}$, leading to an interpretable parts-based representation. 
Owing to the parts-based representation, NMF has been applied in many real-world applications. Especially, NMF has acted as a clustering method \cite{li2018nonnegative}. Ding et al. \cite{ding2005equivalence} pointed out that NMF is a soft version of K-means clustering.   Together with the good data representation ability, the clustering performance of NMF is excellent in many problems, like tumor clustering \cite{zheng2009tumor}, community detection \cite{wu2018nonnegative}, face clustering \cite{8612941} and document clustering \cite{cai2010locally}.

\subsection{Symmetric Nonnegative Matrix Factorization}
The conventional NMF is effective for linear-spreadable data and usually fails to cope  with  non-linear data \cite{7167693}. To this end, as a variant of NMF,  SNMF \cite{kuang2015symnmf} was proposed to play as a graph clustering method, which is effective for handling non-linear data \cite{7167693}. Different from NMF, SNMF decomposes an affinity matrix as the product of a nonnegative  matrix and  its transpose, i.e., 
\begin{equation}
\begin{split}
\min_{\mathbf{V}}\left\|\mathbf{W}-\mathbf{V}\mathbf{V}^\mathsf{T}\right\|_F^2, {\rm s.t.~}\mathbf{V}\geq 0, 
\end{split}
\label{SymNMF}
\end{equation}
where $\mathbf{W}\in\mathbb{R}^{n\times n}$ is the affinity  matrix recording the pairwise relationship between samples, and $\mathbf{V}\in\mathbb{R}^{n\times k}$ is the factorized matrix.   
SNMF is highly related to the well-known spectral clustering (SC) \cite{ng2002spectral}. Concretely, the objective of SC can be formulated as
\begin{equation}
\begin{split}
\min_{\mathbf{V}}\left\|\mathbf{W}-\mathbf{V}\mathbf{V}^\mathsf{T}\right\|_F^2, {\rm s.t.~}\mathbf{V}^\mathsf{T}\mathbf{V}=\mathbf{I}, 
\end{split}
\label{SC}
\end{equation}
 where $\mathbf{I}\in\mathbb{R}^{k\times k}$ is the identity matrix. Comparing  Eq. \eqref{SymNMF} with Eq.\eqref{SC}, we can see that SC and SNMF share the same objective function but with different constraints. Particularity, SC seeks an orthogonal decomposition, while SNMF learns a nonnegative embedding. 
 The same objective function reveals why SNMF is capable of acting as a general graph clustering method, while the different constraints increase the interpretability of the factorized matrix of SNMF, i.e., the location of the largest value of $\mathbf{v}_i$ can indicate the cluster membership of sample $\mathbf{x}_i$ \cite{kuang2015symnmf}. 
% Specifically, the clustering result is indicated by the clustering membership matrix 
% $\mathcal{B}(V)$, i.e.,
% \begin{equation}
% \mathcal{B}(\mathbf{V})_{ij}=\begin{cases}
% 1, ~{\rm if~\mathbf{V}_{ij}~is~the~largest~value~in~the}~i{\rm th~row~ \mathbf{V}_{i:} }\\
% 0, ~{\rm others},
% \end{cases}
% \end{equation}
% where $\mathcal{B}(\cdot)$ is an operator to generate the clustering membership matrix from SNMF. 
% In SNMF, we could set $k=c$, where $c$ denotes the number of clusters.  This is one of the major advantages of SNMF over SC since SC needs to perform the K-means algorithm on $\mathbf{V}$ to obtain the final clustering result. 
Specifically, for SNMF, the partition matrix (or clustering membership matrix)  $\mathbf{M}\in\mathbb{R}^{n\times k}$ is obtained by
%$\mathcal{B}(V)$, i.e.,
%\begin{small}
%\begin{equation}
%\mathcal{B}(\mathbf{V})_{ij}=\begin{cases}
%1, ~{\rm if}~\mathbf{V}_{ij}{\rm~is~the~largest~value~in~the}~i{\rm th~row~of~\mathbf{V} }\\
%0, ~{\rm others}.
%\end{cases}
%\end{equation}\end{small}
\begin{equation}
	\mathbf{M}_{ij}=\begin{cases}
	1, ~{\rm if}~\mathbf{V}_{ij}=\max_{j}\mathbf{V}_{ij} \\
	0, ~{\rm others},
	\end{cases}
	\label{clusteringMembership}
\end{equation}
%where $\mathcal{B}(\cdot)$ is an operator to generate the clustering membership matrix from SNMF. 
where $\mathbf{M}_{ij}$ and $\mathbf{V}_{ij}$ are the $(i,j)$-th entries of $\mathbf{M}$ and $\mathbf{V}$, respectively. In SNMF, we could set $k=c$ with $c$ being  the number of clusters.  This is one of the main advantages of SNMF over SC that needs to perform the K-means algorithm on $\mathbf{V}$ to obtain the final clustering result.
 
%SNMF has been developed as an important clustering technique.
Plenty of  SNMF-based  clustering methods were proposed. For example,  Gao et al. \cite{gao2018graph} proposed graph regularized SNMF for clustering, which is formulated as  
\begin{equation}
\begin{split}
\min_{\mathbf{V}}\left\|\mathbf{W}-\mathbf{V}\mathbf{V}^\mathsf{T}\right\|_F^2+\lambda{\rm tr}\left(\mathbf{V}^\mathsf{T}\mathbf{LV}\right), {\rm s.t.~}\mathbf{V}\geq 0, 
\end{split}
\label{GSymNMF}
\end{equation}
where the Laplacian graph $\mathbf{L}\in\mathbb{R}^{n\times n}$ is utilized  to incorporate the local structure of the input into SNMF. Being aware of the importance of the input affinity matrix, Jia et al. \cite{9072553} proposed to adaptively learn an affinity matrix from the samples rather than use a predefined one, i.e.,
\begin{equation}
\begin{split}
&\min_{\mathbf{V},\mathbf{S}}\|\mathbf{X}-\mathbf{XS}\|_F^2+\lambda\left\|\mathbf{S}-\mathbf{V}\mathbf{V}^\mathsf{T}\right\|_F^2+\mu\|\mathbf{S}-\mathbf{W}\|_F^2\\
&{\rm s.t.~}\mathbf{V}\geq 0, \mathbf{S}\geq 0, {\rm diag}(\mathbf{S})=0,
\end{split}
\label{caSymNMF}
\end{equation}
where $\mathbf{S}\in\mathbb{R}^{n\times n}$ is the learned affinity matrix, and $\mathbf{W}$ is the empirically built affinity matrix. Eq. \eqref{caSymNMF} could simultaneously perform clustering and learn the affinity  matrix to achieve a mutual enhancement. 
SNMF was also extended to solve the multi-view clustering problem \cite{zhang2016multi,ni2015flexible} and the multi-task clustering problem \cite{al2014multi,zhang2018multi}. Since SNMF is performed on an affinity matrix, SNMF-based methods have been developed as an essential tool in correlation modeling \cite{shi2018short}, network analysis \cite{gligorijevic2018non}, link prediction \cite{zhang2018beyond}, and community detection \cite{ganji2018lagrangian,6985550}. 
Moreover, many semi-supervised SNMF models were proposed to incorporate the available supervisory information \cite{6985550,9013063,8361078,7167693}, which can generate better clustering performance than the unsupervised one. 

Both the basic  SNMF and the above-mentioned advanced models involve solving a 4-th order non-convex optimization problem of $\mathbf{V}$ with the nonnegative constraint iteratively, and there is no closed-form solutions for them. In general, taking the basic SNMF as an example, $\mathbf{V}$ is updated iteratively to optimize Eq. \eqref{SymNMF}  from a random initialization matrix   $\mathbf{V}^0\in\mathbb{R}^{n\times c}$, i.e., $\mathbf{V}^0\rightarrow\cdots\mathbf{V}^{t-1}\rightarrow\mathbf{V}^t\cdots$, where $t$ is the iteration index. In each iteration, the loss function is non-increased, i.e., $\|\mathbf{W}-\mathbf{V}^{t}\mathbf{V}^{{t}^\mathsf{T}}\|\leq \|\mathbf{W}-\mathbf{V}^{t-1}\mathbf{V}^{{t-1}^\mathsf{T}}\|$. 
However, the objective function  of SNMF is non-convex, and thus, the initialization matrix $\mathbf{V}^0$ will severely affect the final solution $\mathbf{V}^*$.  This means that some poor initialization matrices will  lead to inferior clustering results, which limits the application of SNMF.

%However, as shown in Eq. \eqref{SymNMF} SNMF needs to solve a non-convex optimization. This means that the SNMF-based model, including both the basic one in Eq. \eqref{SymNMF} and the ones mentioned above are all sensitive to the initialization of $\mathbf{V}$.  This may limit the performance of SNMF.

%
%[] proposed to factorize an affinity matrix, making it a graph clustering method.
%
%The relation between SymNMF and spectral clustering can be found in \cite{9072553}.
%
%The main drawback of SymNMF is sensitive to the initialization.  
%%%%%%%%%%%%%%%%%%%%%%%%%%%%%%%%%%
\section{Proposed model}

As earlier mentioned, SNMF needs to solve a non-convex optimization problem, which is sensitive to the initialization of variables. 
By utilizing such a sensitivity characteristic of SNMF, we propose self-supervised SNMF (S$^3$NMF), which is capable of progressively boosting clustering performance without relying on any additional information. 
%In this paper, we turns the stone into gold, which regards the drawback of SNMF into a possible start point to improve the performance of SNMF.
%In this paper, we do not treat this as a drawback. Instead, we regard it as a possible start point to improve the performance of SNMF. 
To be specific, S$^3$NMF is inspired  by the success of ensemble clustering that aims to seek a consensus and better clustering result from a number of clustering results obtained by different kinds of clustering methods \cite{SEC-2017-TKDE} via exploiting the coherence among them.
%Given a number of different clusterings for a particular data set, ensemble clustering seeks a consensus clustering, which may produce a better partition than the existing clusterings for that data set.  
%An underlying assumption of ensemble clustering is that the different input clustering partitions should diverse to each other, such that a better clustering result could be finalized  by exploiting the information from different clusterings.  
Due to the sensitivity of SNMF to the initialization of variables, we can treat the decomposed matrices of SNMF with various random initialization as diverse  clustering results. Thus, our S$^3$NMF boils down to  how to improve the clustering performance of SNMF with different initialization.

%As shown in Eq. (2), SNMF fulfills clustering by learning graph embedding $\mathbf{V}$ from $\mathbf{W}$. Clustering is a data mining process, which implies that the clustering result is more discriminative than the input. Therefore,  we argue that  $\mathbf{V}$ is more discriminative than the original similarity matrix $\mathbf{W}$.  Based on this observation, we take advantage of the embedding $\mathbf{V}$ with multiple initializations to form a superior similarity matrix $\mathbf{S}$. Assuming we have $b$ different initializations, the proposed model is formulated as
Unlike  the SNMF in Eq. \eqref{SymNMF}, where the variable is initialized  with a single  random nonnegative matrix $\mathbf{V}^0$,
we first generate a set of random nonnegative matrices denoted as  $\{\mathbf{V}^0_m\in\mathbb{R}^{n\times c}\}|_{m=1}^b$, where $b$ is the size of the set, 
and thus we can obtain $b$ clustering partitions  denoted as  $\{\mathbf{M}_m\}|_{m=1}^b$ after  SNMF. Considering that each  $\mathbf{M}_m$ is usually  more discriminative than the original similarity matrix $\mathbf{W}$, we could form a superior affinity matrix $\mathbf{S}$ as
%\begin{equation}
%\mathbf{S}=\sum_{m=1}^b\alpha_m\mathcal{B}(\mathbf{V}_m)\mathcal{B}(\mathbf{V}_m^\mathsf{T}),
%\label{learnS}
%\end{equation}
\begin{equation}
\mathbf{S}=\sum_{m=1}^b\alpha_m\mathbf{M}_m\mathbf{M}_m^\mathsf{T},
\label{learnS}
\end{equation}
where  $\alpha_m$ is the $m$-th element of $\bm{\alpha}\in\mathbb{R}^{b\times 1}$, the weight vector balancing the contribution of each partition whose derivation will be introduced later. With  $\mathbf{S}$ obtained, we can generate a group of new and better clustering partitions under multiple initialization again. This process is repeated until the stopping criterion or maximum number of iterations is achieved.

Explicitly, we formulate the above procedure as a constrained  optimization model, i.e.,
\begin{equation}
\begin{split}
&\min_{\mathbf{V}_m,\mathbf{S},\bm{\alpha}}\sum_{m=1}^b\mathbf{\alpha}_m\left\|\mathbf{S}-\mathbf{V}_m\mathbf{V}_m^\mathsf{T}\right\|_F^2\\
&{\rm s.t.~}\mathbf{V}_m\geq 0, \forall m, \bm{\alpha}\mathbf{ 1}=1, \bm{\alpha}\geq0,
\end{split}
\label{proposed1}
\end{equation}
where  $\mathbf{1}\in\mathbb{R}^{b\times 1}$ denotes the all one vector,
%$\mathbf{\alpha}=[\alpha_1,\cdots,\alpha_b]\in\mathbb{R}^{1\times b}$ is a weight vector to measure the quality of each $\mathbf{V}_m$, 
the constraint 
$\bm{\alpha}\mathbf{1}=1$ avoids the trivial solution of $\bm{\alpha}$ (i.e., $\bm{\alpha}=0$), and $\bm{\alpha} \geq 0$ guarantees that each ${\alpha}_m$ is a valid weight.  
Intuitively, by minimizing Eq. \eqref{proposed1}, a good (resp. poor) $\mathbf{V}_m$ will produce a smaller (resp. larger)  $\|\mathbf{S}-\mathbf{V}_m\mathbf{V}_m^\mathsf{T}\|_F^2$ and likewise  a  larger (resp. smaller)  $\mathbf{\alpha}_m$. 
%Therefore, the value of $\mathbf{\alpha}_m$ could measure the quality of $\mathbf{V}_m$. The self-learning quality measure is further used to construct $\mathbf{S}$ in Eq. \eqref{learnS}. 
Therefore, the value of $\mathbf{\alpha}_m$ could measure the quality of $\mathbf{V}_m$, and be further employed  to construct $\mathbf{S}$ in Eq. \eqref{learnS}.
However, due to the nonnegative constraint on $\bm{\alpha}$, Eq. \eqref{proposed1} imposes an implicit weighted $\ell_1$ norm on $\bm{\alpha}$. This may result in a quite sparse solution when optimizing Eq. \eqref{proposed1}, i.e., the majority of $\bm{\alpha}$ will be or quite close to $0$. As we aim to leverage the power of multiple clustering partitions, an extreme sparse $\bm{\alpha}$ is not a perfect choice. To this end, we  introduce a hyper-parameter $\tau$ to control the distribution of the entries of  $\bm{\alpha}$, and the final model is written as 
\begin{equation}
\begin{split}
&\min_{\mathbf{V}_m,\mathbf{S},\bm{\alpha}}\sum_{m=1}^b(\mathbf{\alpha}_m)^\tau\left\|\mathbf{S}-\mathbf{V}_m\mathbf{V}_m^\mathsf{T}\right\|_F^2\\
&{\rm s.t.~}\mathbf{V}_m\geq 0, \forall m, \bm{\alpha}\mathbf{ 1}=1, \bm{\alpha}\geq0,
\end{split}
\label{proposed}
\end{equation}
where $\tau$ lies in the range of $(1,+\infty)$. When $\tau$ is close to $1$, only a few entries of $\bm{\alpha}$ will dominate the  vector, while, when  $\tau$ tends to  $+\infty$, minimizing Eq. \eqref{proposed} will assign equal weights to $\bm{\alpha}$. Therefore, $\tau$ should be neither too large or too small. Thus, the value of $\tau$ is empirically set to $2$, and the influence of $\tau$ is experimentally investigated in Section IV.D.  After solving Eq. (9), we can obtain a set of clustering partitions $\{\mathbf{M}_m\}|_{m=1}^b$, a better affinity matrix $\mathbf{S}$, and an estimated quality measure vector $\bm{\alpha}$.  

\textit{Remark.}  It is worth pointing out that S$^3$NMF is different from the existing  ensemble clustering methods, which require a number of  clustering partitions by different kinds of clustering methods as the input. The proposed model only needs an affinity matrix as the input, which is identical to the basic SNMF. 
%has the identical input as the basic SNMF: an affinity matrix, and it adopt 
%, and as will be shown later in the experiments, it can significantly outperform SNMF and  state-of-the-art graph clustering  methods and ensemble clustering methods.  

To solve Eq \eqref{proposed}, we propose an alternating iterative   method. First, with a fixed $\mathbf{S}$\footnote{In the first iteration, $\mathbf{S}=\mathbf{W}$.} as well as multiple random nonnegative initialization matrices $\mathbf{V}_m^0$s, we solve 
$\mathbf{V}_m, \forall m$ and $\bm{\alpha}$ by
\begin{equation}
\begin{split}
&\min_{\mathbf{V}_m,\bm{\alpha}}\sum_{m=1}^b(\mathbf{\alpha}_m)^\tau\left\|\mathbf{S}-\mathbf{V}_m\mathbf{V}_m^\mathsf{T}\right\|_F^2\\
&{\rm s.t.~}\mathbf{V}_m\geq 0,\forall m, \bm{\alpha }\mathbf{1}=1, \bm{\alpha}\geq0.
\end{split}
\label{fixS}
\end{equation}
See the detailed solution of Eq. \eqref{fixS} in Section III.B. 
Then, fixing $\mathbf{V}_m,\forall m$ and $\bm{\alpha}$ as those  previously obtained, we update $\mathbf{S}$ via Eq. \eqref{learnS}.
Those two steps are performed  alternatively and iteratively. See Algorithm 1 for the detailed description.

%\subsection{Optimization of Eq. \eqref{proposed}}
%
%
%With Fixed $\mathbf{S}$

\begin{algorithm}[!t]
	\caption{The Proposed S$^3$NMF}
	\begin{algorithmic}[1]
		\renewcommand{\algorithmicrequire}{\textbf{Input:}}
		\renewcommand{\algorithmicensure}{\textbf{Initialize:}}
		\REQUIRE  $\mathbf{W}$, $\tau=2$, $b$;
		\ENSURE iter=$1$, maxIter=$10$, $\mathbf{S}=\mathbf{W}$;
		\WHILE{iter $<$ maxIter}
%		\STATE Initialize $\mathbf{V}$ with $b$ random non-negative matrices;  
		\STATE Update $\mathbf{V}_m,\forall m$ and $\bm{\alpha}$ via  Algoithm 2;
		\STATE Update $\mathbf{S}$ by Eq. \eqref{learnS};
		\IF {the stopping criterion is meet}
		%		\BREAK % intending to break
		\STATE \textbf{break};	
		\ENDIF
		\STATE iter= iter +1;
		\ENDWHILE
		\STATE Generate clustering membership matrces $\{\mathbf{M}_m\}|_{m=1}^b$ by Eq. \eqref{clusteringMembership};
	\end{algorithmic}
\textbf{Output:} a group of clustering results $\{\mathbf{M}_m\}_{m=1}^b$.
%	\hspace*{\algorithmicindent} \textbf{Output} dd
%	\renewcommand{\algorithmicensure}{\textbf{Initialize:}}
%	\ENSURE d
	\label{PESnmf}
\end{algorithm}

\subsection{Stopping Criterion of Algorithm 1}
%Algorithm 1 needs to alternatively update $\mathbf{S}$ and $\mathbf{V}$ iteratively. 
We propose a simple yet effective criterion to terminate  Algorithm 1 adaptively. 
%As the proposed model would generate multiple clustering partitions characterized by $\mathbf{V}_m, \forall m$. 
It is reasonable to assume that at the first a few iterations, the clustering performance of all the partitions could be improved progressively, and the consensus across them can also be increased. When the largest consensus is achieved, the consensus across them will keep at such a high level or even might decrease and fluctuate due to the randomness in the initialization of variables at that iteration. Therefore, we leverage the level of consensus across different partitions to construct the stopping criterion. 

Mutual information is a quantity that measures the relationship between two random variables. We treat the clustering results of two partitions as two sets of random variables, and then use the normalized mutual information (NMI) to assess their correlation, i.e., 
%\begin{equation}
%{\rm NMI}_{ij}=\frac{{\rm I}\left(\mathcal{B}(\mathbf{V}_j),\mathcal{B}(\mathbf{V}_j)\right)}{({\rm H}(\mathcal{B}(\mathbf{V}_i))+{\rm H}(\mathcal{B}(\mathbf{V}_j)))/2},
%\end{equation}
\begin{equation}
{\rm NMI}_{ij}=\frac{\mathcal{I}\left(\mathbf{M}_i,\mathbf{M}_j\right)}{\left(\mathcal{H}(\mathbf{M}_i)+\mathcal{H}(\mathbf{M}_j)\right)/2},
\end{equation}
%where ${\rm I}\left(\mathcal{B}(\mathbf{V}_j),\mathcal{B}(\mathbf{V}_j)\right)$ denotes the mutual information between partitions $i$ and $j$, and ${\rm H}(\mathcal{B}(\mathbf{V}_i))$ is the entropy of $\mathcal{B}(\mathbf{V}_i)$. The normalized version is adopted as the value of  ${\rm NMI}_{ij}$ will be in the range of  $[0,1]$ for all the sets of random variables. 
where $\mathcal{I}\left(\mathbf{M}_i,\mathbf{M}_j\right)$ denotes the mutual information between partitions $i$ and $j$, and $\mathcal{H}(\mathbf{M}_i)$ is the entropy of $\mathbf{M}_i$. The normalized version is adopted as the value of  ${\rm NMI}_{ij}$ will be in the range of  $[0,1]$ for all the sets of random variables. 
Based on NMI, we use the average NMI (ANMI) to measure the consensus level  of all the partitions, i.e., 
\begin{equation}
{\rm ANMI}=\frac{1}{b(b-1)}\sum_{i=1}^b\sum_{j=i+1}^b {\rm NMI}_{ij}.
\end{equation}
ANMI lies in the range of $[0,1]$, and a larger ANMI indicates a higher level of consensus across different partitions. We terminate Algorithm 1 when the value of ANMI begins to drop, and take the outputs of  the iteration with the highest ANMI as the final clustering result.

\subsection{Numerical  Solution to Eq. \eqref{fixS} }
%
%Due to the existence of the non-negative constraints, Eq. \eqref{Valpha} is non-convex and quite challenging to solve. To solve it, we propose a 
Eq. \eqref{fixS} is non-convex, and thus quite challenging to solve. To tackle this challenge, we propose to optimize  $\mathbf{V}_m, \forall m$ and $\bm{\alpha}$ alternatively and iteratively, i.e., update  $\mathbf{V}_m, \forall m$ with a fixed $\bm{\alpha}$, and then update  $\bm{\alpha}$ with a fixed $\mathbf{V}_m, \forall m$. 

\subsubsection{Update  $\mathbf{V}$}
With a fixed $\bm{\alpha}$, the $\mathbf{V}$-subproblem is expressed as
\begin{equation}
\begin{split}
&\min_{\mathbf{V}_m}\sum_{m=1}^b(\mathbf{\alpha}_m)^\tau\left\|\mathbf{S}-\mathbf{V}_m\mathbf{V}_m^\mathsf{T}\right\|_F^2,~{\rm s.t.~}\mathbf{V}_m \geq 0, \forall m.
\end{split}
\end{equation}
As the $b$ decomposed matrices  are independent of each other, we could solve each $\mathbf{V}_m$ separately. Specifically, for a typical one, the subproblem is written as 
\begin{equation}
\begin{split}
&\min_{\mathbf{V}_m}(\mathbf{\alpha}_m)^\tau\left\|\mathbf{S}-\mathbf{V}_m\mathbf{V}_m^\mathsf{T}\right\|_F^2,~{\rm s.t.~}\mathbf{V}_m \geq 0.
\end{split}
\label{V-sub}
\end{equation}
Eq. \eqref{V-sub} is a nonnegativity constrained 4-th order non-convex optimization problem, and there is no closed-form solution. To solve it, 
we first construct an auxiliary function \cite{lee1999learning}, which is a tight upper bound function of the original function and defined as
%we use the method in \cite{lee1999learning} that first constructs an auxiliary function and then decreases the auxiliary function. Specifically, 
%the auxiliary function is defined as:

\noindent
\textbf{Definition 1} \textit{$g(x)$ is an auxiliary function of $f(x)$, if the following conditions hold}
\begin{equation}
g(x)\geq f(x),~g(x=x^t)=f(x=x^t),
\end{equation}
\textit{where $g(x=x^t)=f(x=x^t)$ means that at point $x^t$, $f$ and $g$ have the same value.} 

%The auxiliary function is a tight upper bound function of the original function. 
If we decrease $g(x)$ at each iteration, i.e.,  $g(x^{t+1})<g(x^t)$, the original function $f(x)$ will also be decreased, i.e., $f(x^{t+1})<f(x^t)$ because we have 
$f(x^t)=g(x^t)> g(x^{t+1})\geq f(x^{t+1})$. Therefore, the optimization of Eq. \eqref{V-sub} becomes constructing an appropriate auxiliary function and decreasing it at each iteration.

Specifically, at the $t$-th iteration, we build the auxiliary function of Eq. \eqref{V-sub} as
%\begin{equation}
%\begin{split}
%&g(\mathbf{V}_{m})=\mathbf{\alpha}_m^\tau\sum_{ijk}\left(\mathbf{V}^t_m\mathbf{V}_m^{t^\mathsf{T}}\right)_{ik}\mathbf{V}_{m_{kj}}\frac{\mathbf{V}_{m_{ij}}^4}{\mathbf{V}_{m_{ij}}^{t^3}}\\
%&~-2\mathbf{\alpha}_m^\tau\sum_{ijk}\mathbf{S}_{ik}\mathbf{V}^t_{m_{ij}}\mathbf{V}^t_{m_{kj}}\left(1+{\rm log}\frac{\mathbf{V}_{m_{ij}}\mathbf{V}_{m_{kj}}}{\mathbf{V}^t_{m_{ij}}\mathbf{V}^t_{m_{kj}}}\right),
%\end{split}
%\label{auxiliary-V}
%\end{equation}
\begin{equation}
\begin{split}
&g(\mathbf{V}_{m})=(\mathbf{\alpha}_m)^\tau\sum_{ijk}\left(\mathbf{V}^t_m\mathbf{V}_m^{t^\mathsf{T}}\right)_{ik}\mathbf{V}_{m_{kj}}\frac{(\mathbf{V}_{m_{ij}})^4}{(\mathbf{V}_{m_{ij}}^{t})^3}\\
&~-2(\mathbf{\alpha}_m)^\tau\sum_{ijk}\mathbf{S}_{ik}\mathbf{V}^t_{m_{ij}}\mathbf{V}^t_{m_{kj}}\left(1+{\rm log}\frac{\mathbf{V}_{m_{ij}}\mathbf{V}_{m_{kj}}}{\mathbf{V}^t_{m_{ij}}\mathbf{V}^t_{m_{kj}}}\right),
\end{split}
\label{auxiliary-V}
\end{equation}
where $\mathbf{V}^t_m$ denotes the value of $\mathbf{V}_m$ at the $t$-th iteration, and  $\mathbf{V}_{m_{ij}}$ denotes the element of $\mathbf{V}_m$ at the $i$-th row and $j$-th column. To avoid the confusion between iteration index (e.g., $t$) and the exponentiation operator,  for an exponentiation operator we put the base in parentheses, i.e., $(\mathbf{V}_{m_{ij}})^4$ denotes the 4-th power of $\mathbf{V}_{m_{ij}}$. 
Appendix A proves that Eq. \eqref{auxiliary-V} is a valid auxiliary function of Eq. \eqref{V-sub}. 
The first order  and second order  derivatives of Eq. \eqref{auxiliary-V} are
%\begin{equation}
%\begin{split}
%\frac{\partial g}{\partial\mathbf{V}_{m_{ij}}}&=4\left(\mathbf{V}'_m\mathbf{V'}_m^\mathsf{T}\mathbf{V'}_m\right)_{ij}\frac{\mathbf{V}_{m_{ij}}^3}{\mathbf{V'}_{m_{ij}}^3}\\
%&-2\frac{\left(\mathbf{SV'}_m\right)_{ij}\mathbf{V'}_{m_{ij}}}{\mathbf{V}_{m_{ij}}}-2\frac{\left(\mathbf{S}^\mathsf{T}\mathbf{V'}_m\right)_{ij}\mathbf{V'}_{m_{ij}}}{\mathbf{V}_{m_{ij}}}\\
%&=4\left(\mathbf{V}'_m\mathbf{V'}_m^\mathsf{T}\mathbf{V'}_m\right)_{ij}\frac{\mathbf{V}_{m_{ij}}^3}{\mathbf{V'}_{m_{ij}}^3}-4\frac{\left(\mathbf{SV'}_m\right)_{ij}\mathbf{V'}_{m_{ij}}}{\mathbf{V}_{m_{ij}}}\\
%\end{split}
%\end{equation}
\begin{equation}
\begin{split}
\frac{\partial g(\mathbf{V}_{m})}{\partial\mathbf{V}_{m_{ij}}}=&4(\mathbf{\alpha}_m)^\tau\left(\mathbf{V}^t_m\mathbf{V}_m^{t^\mathsf{T}}\mathbf{V}^t_m\right)_{ij}\frac{(\mathbf{V}_{m_{ij}})^3}{(\mathbf{V}_{m_{ij}}^{t})^3}\\
&-4(\mathbf{\alpha}_m)^\tau\frac{\left(\mathbf{SV}^t_m\right)_{ij}\mathbf{V}^t_{m_{ij}}}{\mathbf{V}_{m_{ij}}}\\
\end{split}
\end{equation}
and%\begin{small}
\begin{equation}
\begin{split}
\frac{\partial^2g(\mathbf{V}_{m})}{\partial\mathbf{V}_{m_{ij}}\partial\mathbf{V}_{m_{kl}}}=&12(\mathbf{\alpha}_m)^\tau\delta_{ik}\delta_{jl}\left(\mathbf{V}^t_m\mathbf{V}_m^{t^\mathsf{T}}\mathbf{V}^t_m\right)_{ij}\frac{(\mathbf{V}_{m_{ij}})^2}{(\mathbf{V}_{m_{ij}}^{t})^3}\\
&+4(\mathbf{\alpha}_m)^\tau\frac{\left(\mathbf{SV}^t_m\right)_{ij}\mathbf{V}^t_{m_{ij}}}{(\mathbf{V}_{m_{ij}})^2},
\label{2th-orderD}
\end{split}
\end{equation}%\end{small}
respectively, where $\delta_{ij}=1$ (resp. $0$) if $i=j$ (resp. $i\neq j$).  
Since  the second order derivative $\frac{\partial^2g(\mathbf{V}_{m})}{\partial\mathbf{V}_{m_{ij}}\partial\mathbf{V}_{m_{kl}}}>0$, the constructed auxiliary function $g(\mathbf{V}_m)$ is a convex function. The global minimization of $g(\mathbf{V}_m)$ is achieved when $\frac{\partial g(\mathbf{V}_m)}{\partial\mathbf{V}_{m_{ij}}}=0$, i.e., 
\begin{equation}
\mathbf{V}_{m_{ij}}^{t+1}=\mathbf{V}_{m_{ij}}^t\left(\frac{\left(\mathbf{SV}_m^t\right)_{ij}}{\left(\mathbf{V}^t_m\mathbf{V}_m^{t^\mathsf{T}}\mathbf{V}_m^t\right)_{ij}}\right)^\frac{1}{4}, \forall i,j. 
\label{sol:V}
\end{equation}
According to the property of the auxiliary function, the updating in Eq. \eqref{sol:V} also decreases the original function in Eq. \eqref{V-sub}.

\subsubsection{Solve $\bm{\alpha}$}
With the fixed $\mathbf{V}_m, \forall m$, the $\bm{\alpha}$-subproblem is rewritten as  
%\begin{equation}
%\begin{split}
%&\min_{\mathbf{\alpha}}\sum_i^m\mathbf{\alpha}_i^\tau\mathbf{h}_i\\
%&{\rm s.t.~} \mathbf{\alpha 1}=1, \mathbf{\alpha}\geq0
%\end{split}
%\end{equation}
%where $\mathbf{h}_i=\left\|\mathbf{S}-\mathbf{V}_i\mathbf{V}_i^\mathsf{T}\right\|_F^2$,
\begin{equation}
\begin{split}
&\min_{\bm{\alpha}}\sum_m^b({\alpha}_m)^\tau{h}_m, ~{\rm s.t.~} \mathbf{\alpha}\mathbf{1}=1, \bm{\alpha}\geq0,
\end{split}
\label{alpha-sub}
\end{equation}
where ${h}_m=\left\|\mathbf{S}-\mathbf{V}_m^{t+1}\mathbf{V}_m^{{t+1}^\mathsf{T}}\right\|_F^2$. The  Lagrange  function of Eq. \eqref{alpha-sub} is 
%\begin{equation}
%\mathcal{L}=\sum_i^m\mathbf{\alpha}_i^\tau\mathbf{h}_i-\lambda \left(\sum_i^m\mathbf{\alpha}_i-1\right)~{\rm s.t.,}~\mathbf{\alpha}\geq0
%\end{equation} 
\begin{equation}
\mathcal{L}=\sum_m^b({\alpha}_m)^\tau{h}_m-\lambda \left(\sum_m^b{\alpha}_m-1\right)~{\rm s.t.,}~\bm{\alpha}\geq0.
\label{Lagrange}
\end{equation}
Taking the first order  derivative of Eq. \eqref{Lagrange} $\frac{\partial{\mathcal{L}}}{\partial\mathbf{\alpha}_m}=\tau(\alpha_m)^{\tau-1}{h}_m-\lambda$, 
and setting it to zero, we have
\begin{equation}
\mathbf{\alpha}_m=\left(\frac{\lambda}{\tau\mathbf{h}_m}\right)^{\frac{1}{\tau-1}},\forall m.
\label{alpha1}
\end{equation}
Based on  the constraint $\sum_m^b\mathbf{\alpha}_m=1$, $\lambda$ can be obtained as
%\begin{equation}
%\sum_i^m\mathbf{\alpha}_i=\sum_i^m\left(\frac{\lambda}{\tau\mathbf{h}_i}\right)^{\frac{1}{\tau-1}}=\lambda^{\frac{1}{\tau-1}}\sum_{i}^{m}(\tau\mathbf{h}_i)^{\frac{1}{1-\tau}}=1
%\end{equation}
%\begin{equation}
%\lambda^{\frac{1}{\tau-1}}=\frac{1}{\sum_{i}^{m}(\tau\mathbf{h}_i)^{\frac{1}{1-\tau}}}.
%\end{equation}
\begin{equation}
\lambda=\left(\frac{1}{\sum_{m}^{b}(\tau{h}_m)^{\frac{1}{1-\tau}}}\right)^{\tau-1}.
\end{equation}
Substituting $\lambda$ into Eq. \eqref{alpha1}, $\alpha_m$ is obtained:
\begin{equation}
\mathbf{\alpha}_m=\frac{(\tau{h}_m)^\frac{1}{1-\tau}}{\sum_{m}^{b}(\tau{h}_m)^{\frac{1}{1-\tau}}}.
\label{sol:alpha}
\end{equation}
Since both  the numerator and denominator of Eq. \eqref{sol:alpha} are larger than $0$, we have $\alpha_m>0,\forall m$, and the nonnegative constraint for $\bm{\alpha}$ is satisfied. 
The solution in Eq. \eqref{sol:alpha} satisfies the Karush-Kuhn-Tucker (KKT)  conditions  of Eq. \eqref{alpha-sub}, and thus it is a local optimum. Moreover, as Eq. \eqref{alpha-sub} is a convex problem, Eq. \eqref{sol:alpha} is the global optimum of Eq. \eqref{alpha-sub}.

The detailed optimization process of Eq. \eqref{fixS} is summarized in Algorithm 2. If the maximum difference of the variables between two iterations are less than $10^{-3}$, i.e., 
${\rm max}(\|\mathbf{V}_m^{t}-\mathbf{V}_m^{t-1}\|_\infty, \|\bm{\alpha}^t-\bm{\alpha}^{t-1}\|_\infty)<10^{-3}$, 
the iteration will stop.

\begin{algorithm}[!t]
	\caption{Numerical solution of Eq. \eqref{fixS}}
	\begin{algorithmic}[1]
		\renewcommand{\algorithmicrequire}{\textbf{Input:}}
		\renewcommand{\algorithmicensure}{\textbf{Initialize:}}
		\REQUIRE  $\mathbf{S}$, $\tau=2$, $b$
		\ENSURE $t=1$, a set of radnom nonnegative matrices $\mathbf{V}_m^0$; %$\mathbf{F}=\mathbf{F}_0$
		\WHILE{$t<500$}
%		\For{\texttt{<some condition>}}
%		\State \texttt{<do stuff>}
%		\EndFor
		\FOR{$m\in \{1,\dots,b\}$}  %% should include the algorithmic package.
			\STATE Update $\mathbf{V}_m$ by Eq. \eqref{sol:V};
		\ENDFOR
%		\STATE \textcolor{red}{use a loop, for m=1:b7}
%		\STATE Update $\mathbf{V}_m,
%		\forall m$ by Eq. \eqref{sol:V};
%		\ENDFor
		\STATE Update $\bm{\alpha}$ by Eq. \eqref{sol:alpha};
		\IF {converged}
		%		\BREAK % intending to break
		\STATE \textbf{break};	
		\ENDIF
		\STATE $t=t+1$;
		\ENDWHILE
		%		\STATE iter= iter +1;
	\end{algorithmic}
	\label{1block}
\end{algorithm}

According to the property of the auxiliary function, we can prove that updating $\{\mathbf{V}_m\}|_{m=1}^b$ will not increase  the objective of Eq. \eqref{fixS}, i.e. $\mathcal{O}(\mathbf{V}_m^{t+1},\bm{\alpha}^t)\leq \mathcal{O}(\mathbf{V}_m^{t},\bm{\alpha}^t)$, where $\mathcal{O}(\mathbf{V}_m,\bm{\alpha})$ denotes  the objective function  of Eq. (10). 
Updating $\bm{\alpha}$ via Eq. \eqref{sol:alpha} achieves the global minimization of Eq. \eqref{fixS} with a fixed $\{\mathbf{V}_m\}|_{m=1}^b$, and thus we have $\mathcal{O}(\mathbf{V}_m^{t+1},\bm{\alpha}^{t+1})\leq \mathcal{O}(\mathbf{V}_m^{{t+1}},\bm{\alpha}^t)$. 
Therefore,  in each iteration of Algorithm 2, we have $\mathcal{O}(\mathbf{V}_m^{t+1},\bm{\alpha}^{t+1})\leq \mathcal{O}(\mathbf{V}_m^{{t}},\bm{\alpha}^t)$. 
Moreover, since both $\bm{\alpha}$ and $\left\|\mathbf{S}-\mathbf{V}_m\mathbf{V}_m^\mathsf{T}\right\|_F^2,\forall m$ are nonnegative, the objective of Eq. \eqref{fixS} is lower-bounded, i.e., $\mathcal{O}(\mathbf{V}_m,\bm{\alpha})\geq 0$.  %$\sum_m^b\mathbf{\alpha}_m^\tau\left\|\mathbf{S}-\mathbf{V}_m\mathbf{V}_m^\mathsf{T}\right\|_F^2\geq 0$. 
Therefore the convergence of Algorithm 2 is guaranteed.

\subsection{Complexity Analysis of Our S$^3$NMF}
We first analyze the computational complexity of Algorithm 2. Algorithm 2 solves the $\mathbf{V}$-subproblem and the $\bm{\alpha}$-subproblem alternatively and iteratively. For the $\mathbf{V}$-subproblem, the computational complexity is $\mathsf{O}(n^2cb)$ and the   $\bm{\alpha}$-subproblem  has a complexity of $\mathsf{O}(n)$. Therefore, the complexity of each iteration of Algorithm 2 is  $\mathsf{O}(n^2cb)$. 

Algorithm 1 involves repeatedly solving Algorithm 2 with the computational complexity of $\mathsf{O}(n^2cbr)$, where $r$ is the maximum iteration number of Algorithm 2, 
and constructing $\mathbf{S}$ with the computational complexity of $\mathsf{O}(n^2c)$. Therefore, the complexity of each iteration of Algorithm 1 is $\mathsf{O}(n^2cbr)$.

\section{Experiments}

\begin{table}
	\label{tab:dataset}
	\begin{center}
		\caption{Details of Datasets}
		\begin{tabular}{lccc}  
%			\toprule
			\hline\hline
			Dataset  & \# Sample $(n)$ & \# Dimension $(d)$ & \# Cluster $(c)$ \\
			\midrule
			
			CHART & 600  & 60 &6       \\
			USPST & 2007  & 256 & 10       \\
			SEEDS & 210  & 7 & 3       \\
			MSRA & 1799  & 256 & 12       \\
			
%			Semeion & 1593  & 256 & 10    \\
			SEMEION & 1593  & 256 & 10    \\
			PALM & 2000  & 256 & 100       \\
			
			IRIS & 150  & 4 & 3       \\
			
			COIL20 & 1440  & 1024 & 20       \\
			
			MNIST*   & 1000 & 784 & 10      \\
			
			USPS* & 1000  & 256 & 10    \\
			
%			\bottomrule
			\hline\hline
		\end{tabular}
	\end{center}
	\begin{tablenotes}
		\item[*] \tiny{Both USPS and MNIST were generated by 
		random selection from the original datasets. }\
		%	\item[a] Another footnote.
	\end{tablenotes}
\end{table}

To evaluate the proposed method,  we compared it with three types of methods. The first type refers to basic graph clustering, including SNMF  \cite{kuang2015symnmf}; and the classic SC \cite{ng2002spectral}. 
	%	\item [3)] NMF
%\end{itemize}
%\begin{itemize}
%	\item[1)] SNMF  \cite{kuang2015symnmf} is the conventional symmetric NMF model for graph clustering.
%	\item[2)] SC denotes the spectral clustering method proposed in \cite{ng2002spectral}. 
%%	\item [3)] NMF
%\end{itemize}
The second type  is the advanced graph clustering methods based on SNMF/SC, including  sparse SC (SSC) \cite{CSC-2016-TIP}, a convex SC method with a sparse regularizer, which aims to learn a block diagonal representation of the spectral embedding; GSNMF \cite{gao2018graph}, a Laplacian graph regularized SNMF model; spectral rotation (SR) \cite{huang2013spectral}, a variant of SC, which achieves clustering by rotating the spectral embedding of SC;   ANLS \cite{zhu2018dropping}, which  relaxes the symmetric constraint of SNMF and uses alternating nonnegative least squares to solve; HALS \cite{zhu2018dropping}, a more efficient version of ANLS;  sBSUM \cite{IBSymNMF-2017-TSP}, an entry-wise block coordinate descent method for SNMF; and   vBSUM \cite{IBSymNMF-2017-TSP}, a vector-wise block coordinate descent method for SNMF. 
Since our method is related to ensemble clustering, we also compared it with several state-of-the-art ensemble clustering models, including  spectral ensemble clustering (SEC) \cite{SEC-2017-TKDE}, which  efficiently uses a co-association matrix to achieve ensemble clustering;   LWCA \cite{7932479}, which  performs the SC on a locally weighted co-association matrix that takes the quality of each base partition into account; and  	LWGP \cite{7932479}, which performs  graph partitioning on the local weighted co-association matrix. 
For a fair comparison, the base partitions of the ensemble clustering methods under comparison  were generated by SNMF.  
We evaluated all the methods on $10$ commonly used datasets. 
%The sample sizes of the \textcolor{red}{data sets} vary from hundreds to thousands. The dimensions of the data sets range from  $7$ to $1024$. The numbers of clusters vary from  $3$ to $100$. 
The detailed information about the adopted datasets is summarized in Table I .% \ref{tab:dataset}. 

To compare the clustering performance of all the methods quantitatively, we adopted five commonly used metrics, i.e., clustering accuracy (ACC), normalized mutual information (NMI), purity (PUR), adjust rand index (ARI) and F1-score. The detailed definitions of these  metrics can be found in \cite{9072553}. All the metrics except ARI lie in the range of $[0,1]$, while ARI has a value ranging in $[-1, 1]$. For all the metrics, the larger, the better.

The initial  affinity matrix of all the methods was generated by the same $k$-nearest-neighbor graph for a fair comparison, where $k$ was empirically set to ${\rm log}_2n+1$ \cite{von2007tutorial}. 
For the compared methods, we exhaustively searched their hyper-parameters and selected the ones producing the best clustering  performance. Then, we reported their average performance  and the associated standard deviation (std) of $20$ repetitions.  
For the proposed method, we fixed the hyper-parameters as $\tau=2$, $b=20$. According to Algorithm $1$, the proposed model could generate $20$ different clustering results when $b=20$, and we reported the average performance as well as the std.

\begin{table}[t]
	\caption{Clustering Results on MNIST}%\smallskip
	\centering
	\resizebox{1\columnwidth}{!}{
		\smallskip\begin{tabular}{l c c c c c}
			\hline\hline
			Method & ACC &  NMI & PUR & ARI & F1-score\\
			\hline		
			sBSUM & $0.551\pm0.055$ &	$0.578\pm0.024$&	$0.608\pm0.030$&	$0.435\pm0.036$&	$0.502\pm0.030$ \\
			vBSUM & $0.572\pm0.035$ &	$0.575\pm0.017$&	$0.613\pm0.024$&	$0.434\pm0.025$&	$0.500\pm	0.020$ \\
			ALS & $0.565\pm0.049$	&$0.575\pm0.023$	&$0.609\pm0.037$&	$0.435\pm0.035$	&$0.502\pm0.028$\\
			HALS & $0.564\pm0.056$&	$0.576\pm0.027$&	$0.607\pm0.043$&	$0.436\pm0.031$&	$0.501\pm	0.025$ \\
			ECS &  $0.557\pm0.038$ &	$0.548\pm0.021$&	$0.593\pm0.030$&	$0.418\pm0.027$&	$0.484	\pm0.023$\\
			GSNMF	&$0.546\pm	0.049$ &	$0.566\pm 	0.028$ &	$0.590\pm 	0.033$ &	$0.409\pm 	0.036$ &	$0.474\pm 	0.030$\\ 
			LWCA &	$0.544\pm 	0.034$ &	$0.546\pm 	0.023$ &	$0.579\pm 	0.029$ &	$0.417\pm 	0.023$ &	$0.486\pm 	0.019$ \\

			LWGP &	$0.549\pm 	0.036$ &	$0.544\pm 	0.028$ &	$0.580\pm 	0.034$ &	$0.420\pm 	0.035$ &	$0.488\pm 	0.029$\\ 
			SC	& \underline{$0.574\pm 	0.051$} &	\underline{$0.586\pm 	0.021$} &	\underline{$0.622\pm 	0.035$} &	\underline{$0.445\pm 	0.033$} &	\underline{$0.504\pm 	0.028$} \\ 
			
			SR	&$0.544\pm 	0.011$ &	$0.582\pm 	0.010$ &	$0.616\pm 	0.010$& 	$0.423\pm 	0.009$ &	$0.488\pm 	0.007$\\ 
			
			SSC	&$0.551\pm 	0.033$ &	$0.555\pm 	0.020$ &	$0.583\pm 	0.031$& 	$0.431\pm 	0.021$ &	$0.496\pm 	0.018$\\ 
			
			SNMF	&$0.518\pm 	0.042$ &	$0.523\pm 	0.031$ &	$0.569\pm 	0.036$ &	$0.376\pm 	0.041$ &	$0.447\pm 	0.035$\\

			PESNMF &	$\mathbf{0.680\pm 	0.026}$ &	$\mathbf{0.606\pm 	0.011}$ &	$\mathbf{0.684\pm 	0.020}$ &	$\mathbf{0.506\pm 	0.009}$ & 	$\mathbf{0.558\pm 	0.008}$\\ 
			
			\hline\hline
		\end{tabular}
	}
	\label{table-MNIST}
\end{table}

\begin{table}[t]
	\caption{Clustering Results on CHART}%\smallskip
	\centering
	\resizebox{1\columnwidth}{!}{
		\smallskip\begin{tabular}{l c c c c c}
			\hline\hline
			Method & ACC &  NMI & PUR & ARI & F1-score\\
			\hline	
			sBSUM		&$0.616\pm		0.093$&		$0.746\pm		0.062$		&$0.710\pm		0.068$		&$0.607\pm		0.094$		&$0.684\pm		0.071$\\
			vBSUM		&$0.635\pm		0.076$&		$0.758\pm		0.043$		&$0.717\pm		0.057$		&$0.633\pm		0.055$		&$0.705\pm		0.041$\\
			ALS		&$0.651\pm		0.073$&		$0.732\pm		0.046$		&$0.714\pm		0.050$		&$0.611\pm		0.053$		&$0.686\pm		0.042$\\
			HALS		&$0.574\pm		0.136$&		$0.702\pm		0.132$		&$0.670\pm		0.118$		&$0.564\pm		0.151$		&$0.655\pm		0.106$\\
			SEC	&$0.701\pm		0.067$&		$0.752\pm		0.051$		&$0.743\pm		0.051$		&$0.628\pm		0.066$		&$0.699\pm		0.050$\\
			
			GSNMF		&$0.671\pm		0.087$&		$0.675\pm		0.068$		&$0.705\pm		0.069$		&$0.555\pm		0.089$		&$0.639\pm		0.069$\\
			LWCA		&\underline{$0.717\pm		0.050$}&		$0.749\pm		0.043$	&\underline{$0.749\pm		0.041$}		&$0.632\pm		0.058$		&$0.702\pm		0.045$\\

			LWGP		&$0.655\pm		0.077$&		$0.731\pm		0.059$		&$0.714\pm		0.053$		&$0.606\pm		0.077$		&$0.682\pm		0.06$\\
			SC		&$0.675\pm		0.077$&		$0.765\pm		0.026$		&$0.726\pm		0.052$		&$0.653\pm		0.035$		&$0.720\pm		0.026$\\
			
			SR		&$0.568\pm		0.000$&		$0.738\pm		0.000$		&$0.667\pm		0.000$		&$0.615\pm		0.000$		&$0.693\pm		0.000$\\
			
			SSC		&$0.702\pm		0.066$&		\underline{$0.772\pm		0.020$}		&$0.747\pm		0.043$		&\underline{$0.66\pm		0.030$}		&\underline{$0.725\pm		0.023$}\\
			
			SNMF		&$0.671\pm		0.088$&		$0.678\pm		0.065$		&$0.707\pm		0.068$		&$0.557\pm		0.085$		&$0.641\pm		0.067$\\

			Proposed		&$\mathbf{0.869\pm		0.035}$&		$\mathbf{0.828\pm		0.016}$		&$\mathbf{0.87\pm		0.033}$		&$\mathbf{0.763\pm		0.016}$		&$\mathbf{0.803\pm		0.011}$\\

			\hline\hline
		\end{tabular}
	}
	\label{table-chart}
\end{table}

\begin{table}[t]
	\caption{Clustering Results on USPST}%\smallskip
	\centering
	\resizebox{1\columnwidth}{!}{
		\smallskip\begin{tabular}{l c c c c c}
			\hline\hline
			Method & ACC &  NMI & PUR & ARI & F1-score\\
			\hline		
			sBSUM		&$0.477\pm		0.058$		&$0.561\pm		0.050$		&$0.588\pm		0.045$		&$0.333\pm		0.069$		&$0.428\pm		0.054$\\
			vBSUM		&$0.592\pm		0.088$		&$0.660\pm		0.056$		&$0.673\pm		0.069$		&$0.497\pm		0.080$		&$0.555\pm		0.070$\\
			ALS		&$0.658\pm		0.062$		&$0.769\pm		0.035$		&$0.774\pm		0.046$		&$0.632\pm		0.064$		&$0.674\pm		0.055$\\
			HALS		&$0.676\pm		0.086$		&$0.760\pm		0.038$		&$0.773\pm		0.059$		&$0.636\pm		0.079$		&$0.678\pm		0.067$\\
			SEC	&$0.693\pm		0.069$		&$0.760\pm		0.030$		&$0.774\pm		0.043$		&$0.644\pm		0.063$		&$0.680\pm		0.055$\\
			
			GSNMF		&$0.632\pm		0.071$		&$0.659\pm		0.042$		&$0.687\pm		0.057$		&$0.518\pm		0.071$		&$0.572\pm		0.062$\\
			LWCA		&$0.671\pm		0.069$		&$0.728\pm		0.041$		&$0.736\pm		0.054$		&$0.606\pm		0.079$		&$0.652\pm		0.068$\\

			LWGP		&$0.656\pm		0.075$		&$0.734\pm		0.035$		&$0.752\pm		0.045$		&$0.603\pm		0.066$		&$0.649\pm		0.057$\\
			SC		&$0.701\pm		0.067$		&$0.778\pm		0.030$		&$0.795\pm		0.046$		&$0.669\pm		0.062$		&$0.706\pm		0.054$\\
			
			SR		&$0.680\pm		0.003$		&\underline{$0.792\pm		0.006$}		&\underline{$0.806\pm		0.009$}		&$0.660\pm		0.004$		&$0.697\pm		0.003$\\
			
			SSC		&\underline{$0.718\pm		0.073$}		&$0.783\pm		0.029$		&$0.803\pm		0.043$		&\underline{$0.682\pm		0.068$}		&\underline{$0.718\pm		0.059$}\\
			
			SNMF		&$0.587\pm		0.059$		&$0.644\pm		0.036$		&$0.667\pm		0.042$		&$0.487\pm		0.053$		&$0.546\pm		0.046$\\

			Proposed		&$\mathbf{0.838\pm		0.042}$		&$\mathbf{0.805\pm		0.018}$		&$\mathbf{0.857\pm		0.028}$		&$\mathbf{0.777\pm		0.038}$		&$\mathbf{0.802\pm		0.034}$\\

			\hline\hline
		\end{tabular}
	}
	\label{table-USPST}
\end{table}

\begin{table}[t]
	\caption{Clustering Results on SEEDS}%\smallskip
	\centering
	\resizebox{1\columnwidth}{!}{
		\smallskip\begin{tabular}{l c c c c c}
			\hline\hline
			Method & ACC &  NMI & PUR & ARI & F1-score\\
			\hline		
			sBSUM		&$0.729\pm		0.162$		&$0.515\pm		0.192$		&$0.730\pm		0.161$		&$0.502\pm		0.182$		&$0.697\pm		0.072$\\
			vBSUM		&$0.761\pm		0.107$		&$0.568\pm		0.089$		&$0.771\pm		0.093$		&$0.545\pm		0.079$		&$0.712\pm		0.033$\\
			ALS		&$0.730\pm		0.114$		&$0.512\pm		0.112$		&$0.741\pm		0.102$		&$0.497\pm		0.106$		&$0.686\pm		0.051$\\
			HALS		&$0.651\pm		0.133$		&$0.394\pm		0.159$		&$0.663\pm		0.127$		&$0.362\pm		0.172$		&$0.612\pm		0.087$\\
			SEC	&$0.805\pm		0.030$		&$0.579\pm		0.035$		&$0.805\pm		0.030$		&$0.538\pm		0.037$		&$0.694\pm		0.024$\\
			
			GSNMF		&$0.709\pm		0.117$		&$0.468\pm		0.104$		&$0.731\pm		0.089$		&$0.445\pm		0.108$		&$0.643\pm		0.065$\\
			LWCA		&$0.815\pm		0.011$		&$0.585\pm		0.003$		&$0.815\pm		0.011$		&$0.557\pm		0.017$		&$0.706\pm		0.011$\\

			LWGP		&$0.814\pm		0.004$		&$0.549\pm		0.012$		&$0.814\pm		0.004$		&$0.536\pm		0.013$		&$0.693\pm		0.008$\\
			SC		&$0.828\pm		0.031$		&$0.637\pm		0.029$		&$0.829\pm		0.027$		&$0.593\pm		0.022$		&$0.73\pm		0.009$\\
			
			SR		&\underline{$0.833\pm		0.000$}		&\underline{$0.641\pm		0.000$}		&$0.833\pm		0.000$		&$0.597\pm		0.000$		&$0.733\pm		0.000$\\
			
			SSC		&$0.831\pm		0.095$		&$0.617\pm		0.079$		&\underline{$0.836\pm		0.083$}		&\underline{$0.621\pm		0.084$}		&\underline{$0.755\pm		0.041$}\\
			
			SNMF		&$0.602\pm		0.072$		&$0.311\pm		0.103$		&$0.616\pm		0.065$		&$0.279\pm		0.100$		&$0.531\pm		0.068$\\

			Proposed		&$\mathbf{0.881\pm		0.000}$		&$\mathbf{0.667\pm		0.000}$		&$\mathbf{0.881\pm		0.000}$		&$\mathbf{0.688\pm		0.000}$		&$\mathbf{0.792\pm		0.000}$\\
			
			\hline\hline
		\end{tabular}
	}
	\label{table-seeds}
\end{table}

\begin{table}[t]
	\caption{Clustering Results on MSRA}%\smallskip
	\centering
	\resizebox{1\columnwidth}{!}{
		\smallskip\begin{tabular}{l c c c c c}
			\hline\hline
			Method & ACC &  NMI & PUR & ARI & F1-score\\
			\hline		
			sBSUM		&$0.248\pm		0.036$		&$0.218\pm		0.056$		&$0.298\pm		0.044$		&$0.038\pm		0.020$		&$0.182\pm		0.015$\\
			vBSUM		&$0.475\pm		0.031$		&$0.591\pm		0.026$		&$0.517\pm		0.025$		&$0.354\pm		0.027$		&$0.413\pm		0.025$\\
			ALS		&$0.539\pm		0.037$		&\underline{$0.682\pm		0.027$}		&$0.584\pm		0.036$		&$0.433\pm		0.043$		&\underline{$0.487\pm		0.038$}\\
			HALS		&$0.476\pm		0.050$		&$0.573\pm		0.068$		&$0.527\pm		0.057$		&$0.272\pm		0.074$		&$0.358\pm		0.059$\\
			SEC 	&\underline{$0.548\pm		0.037$}		&$0.674\pm		0.025$		&\underline{$0.588\pm		0.035$}		&\underline{$0.435\pm		0.042$}		&\underline{$0.487\pm		0.037$}\\
			
			GSNMF		&$0.489\pm		0.039$		&$0.608\pm		0.026$		&$0.527\pm		0.033$		&$0.370\pm		0.036$		&$0.429\pm		0.032$\\
			LWCA		&$0.521\pm		0.047$		&$0.637\pm		0.039$		&$0.564\pm		0.042$		&$0.400\pm		0.044$		&$0.457\pm		0.038$\\

			LWGP		&$0.511\pm		0.040$		&$0.640\pm		0.029$		&$0.569\pm		0.028$		&$0.395\pm		0.038$		&$0.454\pm		0.031$\\
			SC		&$0.484\pm		0.025$		&$0.634\pm		0.034$		&$0.529\pm		0.024$		&$0.378\pm		0.044$		&$0.439\pm		0.037$\\
			
			SR		&$0.454\pm		0.016$		&$0.597\pm		0.021$		&$0.506\pm		0.017$		&$0.347\pm		0.023$		&$0.414\pm		0.019$\\
			
			SSC		&$0.478\pm		0.031$		&$0.601\pm		0.031$		&$0.517\pm		0.030$		&$0.340\pm		0.034$		&$0.406\pm		0.028$\\
			
			SNMF		&$0.467\pm		0.031$		&$0.567\pm		0.032$		&$0.502\pm		0.033$		&$0.337\pm		0.043$		&$0.400\pm		0.038$\\

			Proposed		&$\mathbf{0.595\pm		0.016}$		&$\mathbf{0.697\pm		0.012}$		&$\mathbf{0.629\pm		0.017}$		&$\mathbf{0.488\pm		0.015}$		&$\mathbf{0.533\pm		0.014}$\\
			
			\hline\hline
		\end{tabular}
	}
	\label{table-MSRA}
\end{table}

\begin{table}[t]
	\caption{Clustering Results on USPS}%\smallskip
	\centering
	\resizebox{1\columnwidth}{!}{
		\smallskip\begin{tabular}{l c c c c c}
			\hline\hline
			Method & ACC &  NMI & PUR & ARI & F1-score\\
			\hline		
			sBSUM		&$0.626\pm		0.088$		&$0.704\pm		0.059$		&$0.704\pm		0.073$		&$0.567\pm		0.085$		&$0.618\pm		0.071$\\
			vBSUM		&$0.674\pm		0.084$		&$0.734\pm		0.043$		&$0.744\pm		0.060$		&$0.607\pm		0.076$		&$0.653\pm		0.065$\\
			ALS		&$0.717\pm		0.053$		&$0.762\pm		0.021$		&$0.777\pm		0.036$		&$0.655\pm		0.040$		&$0.693\pm		0.034$\\
			HALS		&$0.653\pm		0.086$		&$0.718\pm		0.052$		&$0.714\pm		0.065$		&$0.591\pm		0.093$		&$0.640\pm		0.077$\\
			SEC		&$0.707\pm		0.055$		&$0.717\pm		0.035$		&$0.744\pm		0.046$		&$0.602\pm		0.054$		&$0.645\pm		0.047$\\
			
			GSNMF		&$0.643\pm		0.060$		&$0.661\pm		0.047$		&$0.687\pm		0.054$		&$0.514\pm		0.073$		&$0.569\pm		0.064$\\
			LWCA		&$0.687\pm		0.055$		&$0.717\pm		0.033$		&$0.725\pm		0.044$		&$0.602\pm		0.051$		&$0.647\pm		0.044$\\

			LWGP		&$0.741\pm		0.061$		&$0.779\pm		0.021$		&$0.795\pm		0.039$		&$0.687\pm		0.042$		&$0.721\pm		0.037$\\
			SC		&$0.755\pm		0.064$		&$0.757\pm		0.030$		&$0.791\pm		0.046$		&$0.663\pm		0.052$		&$0.699\pm		0.045$\\
			
			SR		&$0.763\pm		0.022$		&\underline{$0.773\pm		0.017$}		&\underline{$0.808\pm		0.021$}		&$0.683\pm		0.022$		&$0.717\pm		0.019$\\
			
			SSC		&\underline{$0.764\pm		0.058$}		&$0.769\pm		0.031$		&$0.788\pm		0.046$		&\underline{$0.688\pm		0.049$}		&\underline{$0.722\pm		0.043$}\\
			
			SNMF		&$0.616\pm		0.077$		&$0.653\pm		0.049$		&$0.674\pm		0.058$		&$0.505\pm		0.074$		&$0.560\pm		0.064$\\

			Proposed		&$\mathbf{0.847\pm		0.019}$		&$\mathbf{0.794\pm		0.013}$		&$\mathbf{0.847\pm		0.019}$		&$\mathbf{0.731\pm		0.016}$		&$\mathbf{0.758\pm		0.014}$\\
			
			\hline\hline
		\end{tabular}
	}
	\label{table-USPS}
\end{table}

\begin{table}[t]
	\caption{Clustering Results on SEMEION}%\smallskip
	\centering
	\resizebox{1\columnwidth}{!}{
		\smallskip\begin{tabular}{l c c c c c}
			\hline\hline
			Method & ACC &  NMI & PUR & ARI & F1-score\\
			\hline	
			sBSUM		&$0.577\pm		0.061$		&$0.609\pm		0.041$		&$0.649\pm		0.046$		&$0.459\pm		0.062$		&$0.521\pm		0.053$\\
			vBSUM		&$0.636\pm		0.051$		&$0.641\pm		0.025$		&$0.690\pm		0.034$		&$0.508\pm		0.043$		&$0.562\pm		0.037$\\
			ALS		&$0.621\pm		0.051$		&$0.634\pm		0.024$		&$0.683\pm		0.031$		&$0.484\pm		0.039$		&$0.541\pm		0.033$\\
			HALS		&$0.621\pm		0.056$		&$0.628\pm		0.027$		&$0.674\pm		0.039$		&$0.487\pm		0.040$		&$0.544\pm		0.033$\\
			SEC	&$0.666\pm		0.050$		&$0.642\pm		0.030$		&$0.693\pm		0.038$		&$0.514\pm		0.043$		&$0.567\pm		0.036$\\
			
			GSNMF		&$0.614\pm		0.054$		&$0.598\pm		0.037$		&$0.651\pm		0.046$		&$0.456\pm		0.054$		&$0.515\pm		0.048$\\
			LWCA		&\underline{$0.679\pm		0.030$}		&$0.644\pm		0.021$		&$0.705\pm		0.027$		&$0.519\pm		0.027$		&$0.571\pm		0.022$\\

			LWGP		&$0.671\pm		0.067$		&$0.646\pm		0.034$		&$0.700\pm		0.046$		&$0.528\pm		0.053$		&$0.580\pm		0.044$\\
			SC		&$0.641\pm		0.059$		&$0.636\pm		0.032$		&$0.682\pm		0.042$		&$0.494\pm		0.052$		&$0.549\pm		0.045$\\
			
			SR		&$0.565\pm		0.004$		&$0.609\pm		0.005$		&$0.644\pm		0.006$		&$0.433\pm		0.007$		&$0.497\pm		0.006$\\
			
			SSC		&\underline{$0.679\pm		0.057$}		&\underline{$0.660\pm		0.030$}		&\underline{$0.713\pm		0.037$}		&\underline{$0.535\pm		0.050$}		&\underline{$0.585\pm		0.043$}\\
			
			SNMF		&$0.591\pm		0.06$		&$0.579\pm		0.036$		&$0.639\pm		0.039$		&$0.435\pm		0.047$		&$0.497\pm		0.040$\\

			Proposed		&$\mathbf{0.72\pm		0.028}$		&$\mathbf{0.667\pm		0.022}$		&$\mathbf{0.731\pm		0.023}$		&$\mathbf{0.573\pm		0.026}$		&$\mathbf{0.619\pm		0.022}$\\

			\hline\hline
		\end{tabular}
	}
	\label{table-Semeion}
\end{table}

\begin{table}[t]
	\caption{Clustering Results on PALM}%\smallskip
	\centering
	\resizebox{1\columnwidth}{!}{
		\smallskip\begin{tabular}{l c c c c c}
			\hline\hline
			Method & ACC &  NMI & PUR & ARI & F1-score\\
			\hline		
			sBSUM		&$0.341\pm		0.074$		&$0.519\pm		0.089$		&$0.392\pm		0.085$		&$0.048\pm		0.021$		&$0.065\pm		0.021$\\
			vBSUM		&$0.173\pm		0.103$		&$0.439\pm		0.114$		&$0.184\pm		0.113$		&$0.057\pm		0.109$		&$0.067\pm		0.108$\\
			ALS		&\underline{$0.830\pm		0.017$}		&$\mathbf{0.951\pm		0.004}$		&\underline{$0.871\pm		0.011$}		&\underline{$0.817\pm		0.016$}		&\underline{$0.819\pm		0.016$}\\
			HALS		&$0.748\pm		0.027$		&$0.919\pm		0.011$		&$0.815\pm		0.019$		&$0.613\pm		0.069$		&$0.618\pm		0.068$\\
			SEC 	&$0.682\pm		0.023$		&$0.910\pm		0.008$		&$0.749\pm		0.019$		&$0.679\pm		0.026$		&$0.683\pm		0.026$\\
			
			GSNMF		&$0.742\pm		0.020$		&$0.893\pm		0.018$		&$0.789\pm		0.013$		&$0.721\pm		0.021$		&$0.724\pm		0.020$\\
			LWCA		&$0.712\pm		0.030$		&$0.907\pm		0.013$		&$0.771\pm		0.023$		&$0.653\pm		0.057$		&$0.657\pm		0.056$\\

			LWGP		&$0.783\pm		0.028$		&\underline{$0.936\pm		0.007$}		&$0.825\pm		0.023$		&$0.780\pm		0.022$		&$0.783\pm		0.021$\\
			SC		&$0.558\pm		0.027$		&$0.836\pm		0.014$		&$0.625\pm		0.023$		&$0.477\pm		0.038$		&$0.483\pm		0.038$\\
			
			SR		&$0.608\pm		0.014$		&$0.846\pm		0.017$		&$0.676\pm		0.010$		&$0.522\pm		0.007$		&$0.528\pm		0.007$\\
			
			SSC		&$0.639\pm		0.038$		&$0.863\pm		0.018$		&$0.708\pm		0.031$		&$0.467\pm		0.063$		&$0.474\pm		0.062$\\
			
			SNMF		&$0.733\pm		0.017$		&$0.922\pm		0.005$		&$0.793\pm		0.013$		&$0.723\pm		0.016$		&$0.726\pm		0.016$\\
			Proposed		&$\mathbf{0.881\pm		0.008}$		&$0.917\pm		0.003$		&$\mathbf{0.886\pm		0.008}$		&$\mathbf{0.854\pm		0.008}$		&$\mathbf{0.856\pm		0.008}$\\
			
			\hline\hline
		\end{tabular}
	}
	\label{table-PALM}
\end{table}

\begin{table}[t]
	\caption{Clustering Results on IRIS}%\smallskip
	\centering
	\resizebox{1\columnwidth}{!}{
		\smallskip\begin{tabular}{l c c c c c}
			\hline\hline
			Method & ACC &  NMI & PUR & ARI & F1-score\\
			\hline	
			sBSUM		&$0.783\pm		0.174$		&$0.680\pm		0.190$		&$0.798\pm		0.156$		&$0.641\pm		0.190$		&$0.781\pm		0.092$\\
			vBSUM		&\underline{$0.885\pm		0.066$}		&$\mathbf{0.776\pm		0.061}$		&\underline{$0.885\pm		0.066$}		&$\mathbf{0.739\pm		0.061}$		&$\mathbf{0.83\pm		0.032}$\\
			ALS		&$0.863\pm		0.089$		&$0.755\pm		0.083$		&$0.863\pm		0.089$		&$0.719\pm		0.082$		&\underline{$0.820\pm		0.043$}\\
			HALS		&$0.826\pm		0.099$		&$0.717\pm		0.094$		&$0.826\pm		0.099$		&$0.676\pm		0.099$		&$0.796\pm		0.052$\\
			SEC	&$0.717\pm		0.026$		&$0.575\pm		0.065$		&$0.717\pm		0.026$		&$0.499\pm		0.064$		&$0.669\pm		0.032$\\
			
			GSNMF		&$0.751\pm		0.121$		&$0.602\pm		0.100$		&$0.765\pm		0.103$		&$0.576\pm		0.118$		&$0.732\pm		0.071$\\
			LWCA		&$0.710\pm		0.040$		&$0.565\pm		0.091$		&$0.711\pm		0.038$		&$0.490\pm		0.083$		&$0.664\pm		0.042$\\

			LWGP		&$0.720\pm	0.000$		&$0.588\pm		0.000$		&$0.720\pm		0.000$		&$0.511\pm		0.000$		&$0.675\pm		0.000$\\
			SC		&$0.467\pm		0.019$		&$0.339\pm		0.007$		&$0.582\pm		0.004$		&$0.229\pm		0.015$		&$0.556\pm		0.008$\\
			
			SR		&$0.463\pm		0.003$		&$0.322\pm		0.000$		&$0.573\pm		0.000$		&$0.210\pm		0.000$		&$0.547\pm		0.000$\\
			
			SSC		&$0.587\pm		0.040$		&$0.554\pm		0.015$		&$0.662\pm		0.003$		&$0.470\pm		0.038$		&$0.683\pm		0.025$\\
			
			SNMF		&$0.645\pm		0.121$		&$0.388\pm		0.141$		&$0.664\pm		0.103$		&$0.346\pm		0.164$		&$0.579\pm		0.102$\\

			Proposed		&$\mathbf{0.886\pm		0.031}$		&\underline{$0.769\pm		0.040$}		&$\mathbf{0.886\pm		0.031}$		&\underline{$0.722\pm		0.045$}		&$0.816\pm		0.031$\\

			\hline\hline
		\end{tabular}
	}
	\label{table-iris}
\end{table}

\begin{table}[t]
	\caption{Clustering Results on COIL20}%\smallskip
	\centering
	\resizebox{1\columnwidth}{!}{
		\smallskip\begin{tabular}{l c c c c c}
			\hline\hline
			Method & ACC &  NMI & PUR & ARI & F1-score\\
			\hline		
			sBSUM		&$0.256\pm		0.055$		&$0.406\pm		0.079$		&$0.338\pm		0.047$		&$0.110\pm		0.052$		&$0.187\pm		0.044$ \\
			vBSUM		&$0.639\pm		0.044$		&$0.787\pm		0.023$		&$0.691\pm		0.037$		&$0.573\pm		0.040$		&$0.597\pm		0.037$ \\
			ALS		&$0.754\pm		0.053$		&$\mathbf{0.875\pm		0.022}$		&$0.810\pm		0.043$		&$0.711\pm		0.049$		&$0.728\pm		0.045$\\
			HALS		&$0.502\pm		0.064$		&$0.728\pm		0.045$		&$0.604\pm		0.051$		&$0.419\pm		0.080$		&$0.460\pm		0.073$ \\
			SEC	&$0.706\pm		0.065$		&$0.829\pm		0.030$		&$0.753\pm		0.051$		&$0.665\pm		0.060$		&$0.683\pm		0.056$ \\
			
			GSNMF		&$0.542\pm		0.040$		&$0.660\pm		0.025$		&$0.562\pm		0.037$		&$0.440\pm		0.039$		&$0.474\pm		0.036$ \\
			LWCA		&$0.692\pm		0.044$		&$0.816\pm		0.025$		&$0.739\pm		0.038$		&$0.639\pm		0.047$		&$0.659\pm		0.044$ \\

			LWGP		&\underline{$0.789\pm		0.024$}		&\underline{$0.872\pm		0.009$}		&$\mathbf{0.832\pm		0.015}$		&\underline{$0.737\pm		0.022$}		&\underline{$0.752\pm		0.020$} \\
			SC		&$0.564\pm		0.052$		&$0.749\pm		0.028$		&$0.628\pm		0.039$		&$0.503\pm		0.051$		&$0.532\pm		0.046$\\
			
			SR		&$0.586\pm		0.031$		&$0.758\pm		0.015$		&$0.652\pm		0.024$		&$0.518\pm		0.027$		&$0.546\pm		0.024$\\
			
			SSC		&$0.619\pm		0.050$		&$0.769\pm		0.031$		&$0.671\pm		0.040$		&$0.536\pm		0.059$		&$0.563\pm		0.055$ \\
			
			SNMF		&$0.491\pm		0.030$		&$0.657\pm		0.019$		&$0.532\pm		0.031$		&$0.386\pm		0.030$		&$0.421\pm		0.029$ \\

			Proposed		&$\mathbf{0.805\pm		0.018}$		&$0.857\pm		0.012$		&\underline{$0.816\pm		0.018$}		&$\mathbf{0.76\pm		0.015}$		&$\mathbf{0.773\pm		0.014}$ \\
			
			\hline\hline
		\end{tabular}
	}
	\label{table-coil20}
\end{table}

\begin{figure*}[t]
	\begin{minipage}[b]{0.195\linewidth}
		\centering
		\centerline{\epsfig{figure=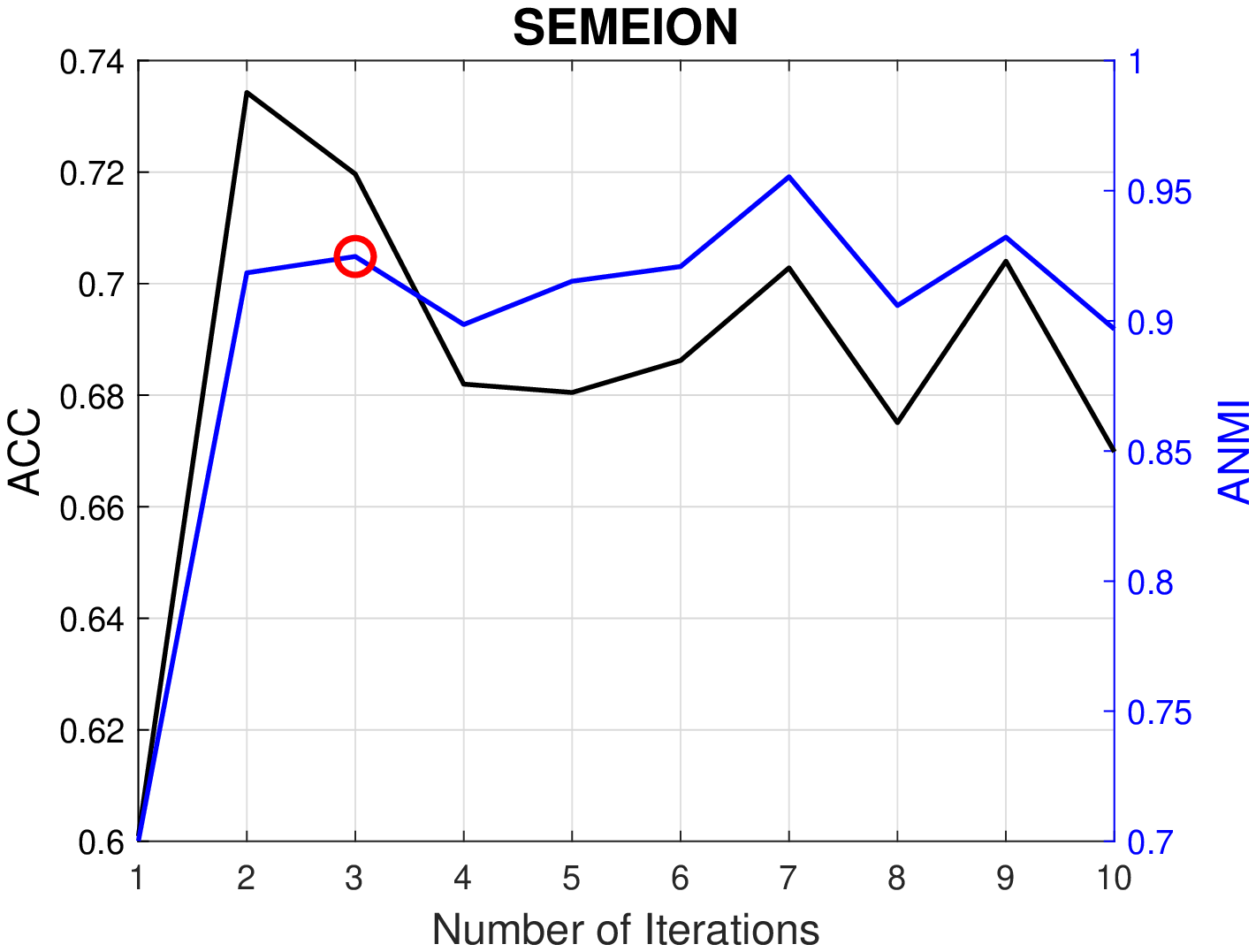,width=3.7cm}}
		%		\vspace{-0.1cm}
		%       \centerline{(a) Result 1}\medskip
	\end{minipage}
	\begin{minipage}[b]{0.195\linewidth}
		\centering
		\centerline{\epsfig{figure=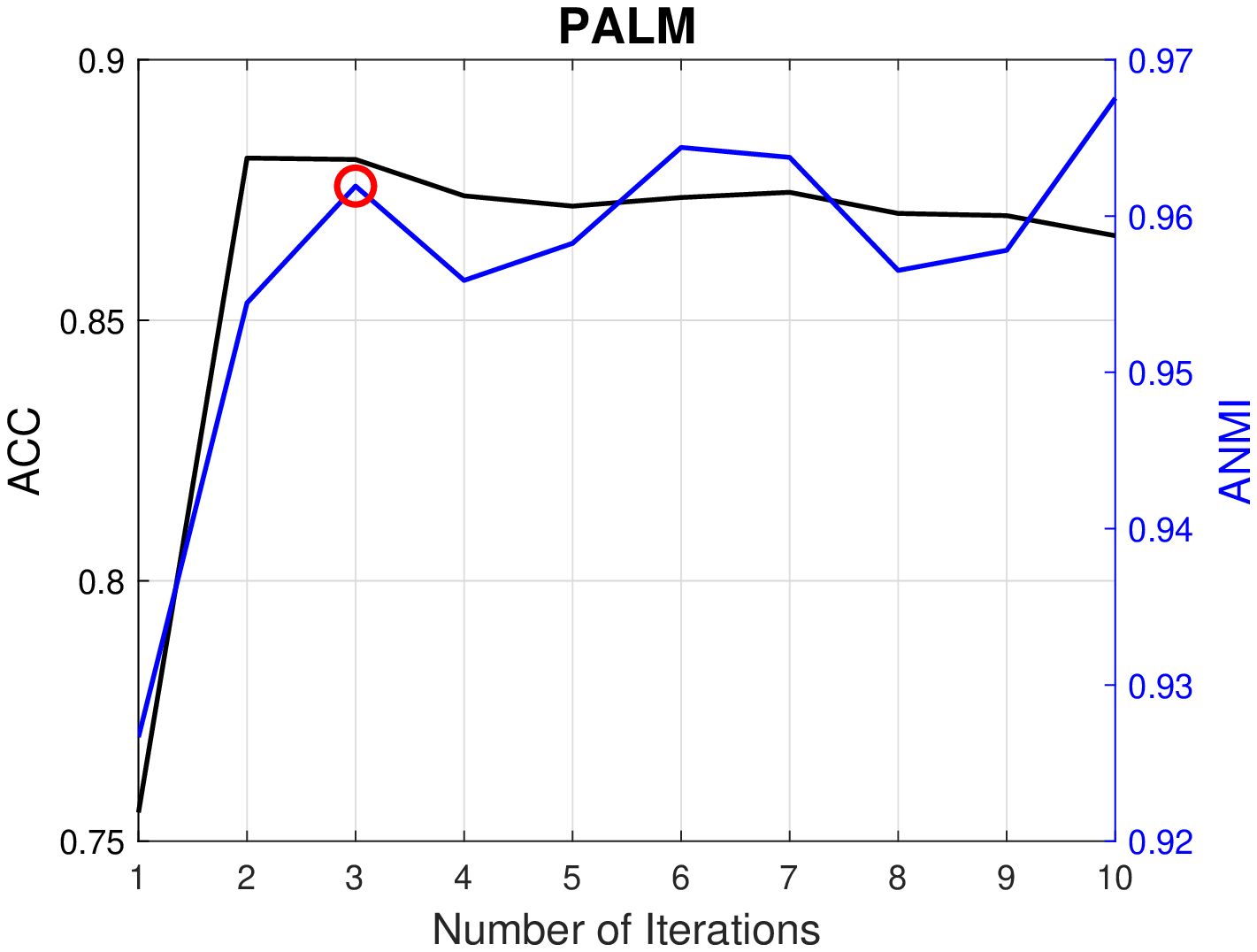,width=3.7cm}}
		%		\vspace{-0.1cm}
		%       \centerline{(a) Result 1}\medskip
	\end{minipage}
	\begin{minipage}[b]{0.195\linewidth}
		\centering
		\centerline{\epsfig{figure=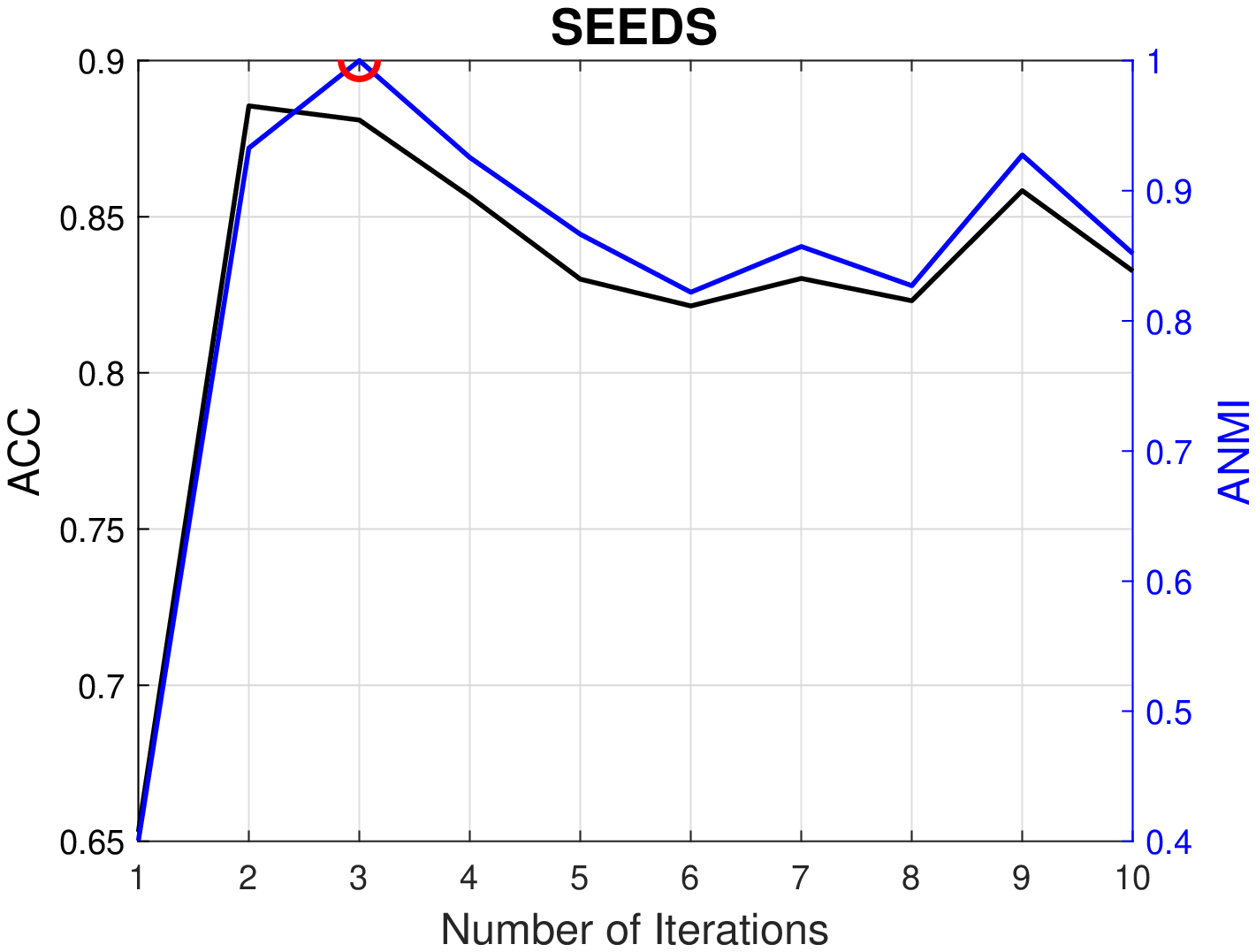,width=3.7cm}}
		%		\vspace{-0.1cm}
		%       \centerline{(a) Result 1}\medskip
	\end{minipage}
	\begin{minipage}[b]{0.195\linewidth}
		\centering
		\centerline{\epsfig{figure=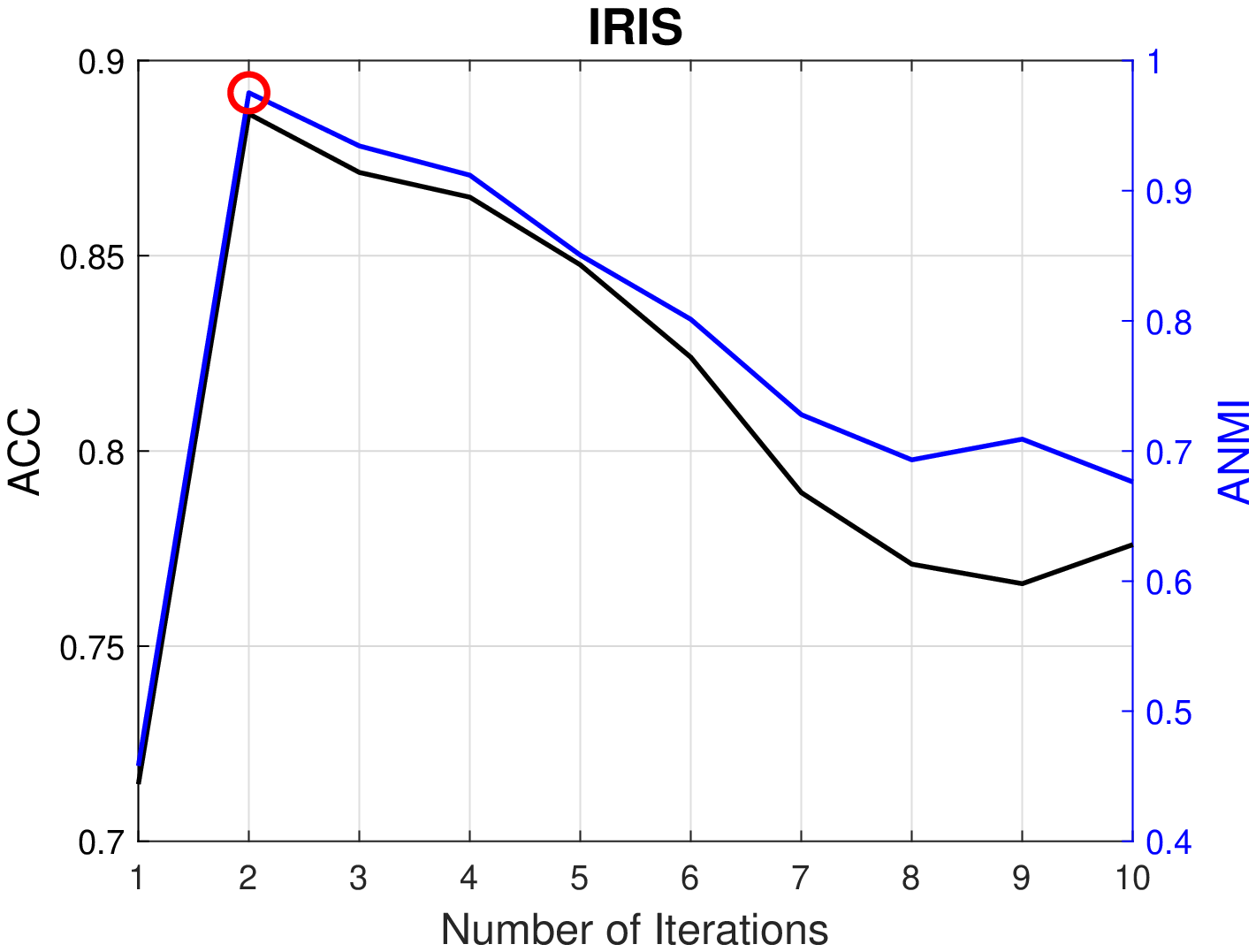,width=3.7cm}}
		%		\vspace{-0.5cm}
		%       \centerline{(a) Result 1}\medskip
	\end{minipage}
	\begin{minipage}[b]{0.195\linewidth}
		\centering
		\centerline{\epsfig{figure=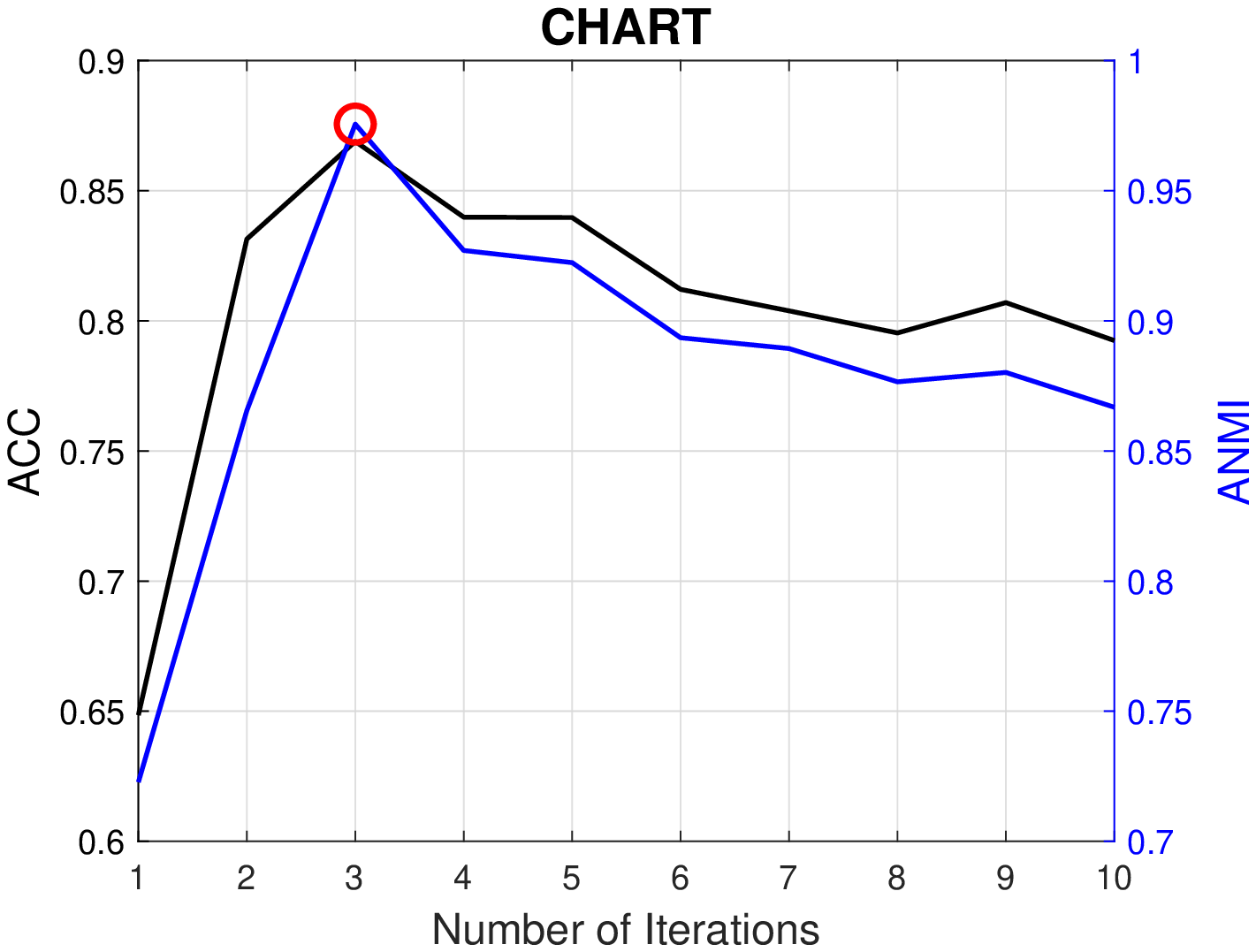,width=3.7cm}}
		%		\vspace{-0.5cm}
		%       \centerline{(a) Result 1}\medskip
	\end{minipage}\\
	\begin{minipage}[b]{0.195\linewidth}
		\centering
		\centerline{\epsfig{figure=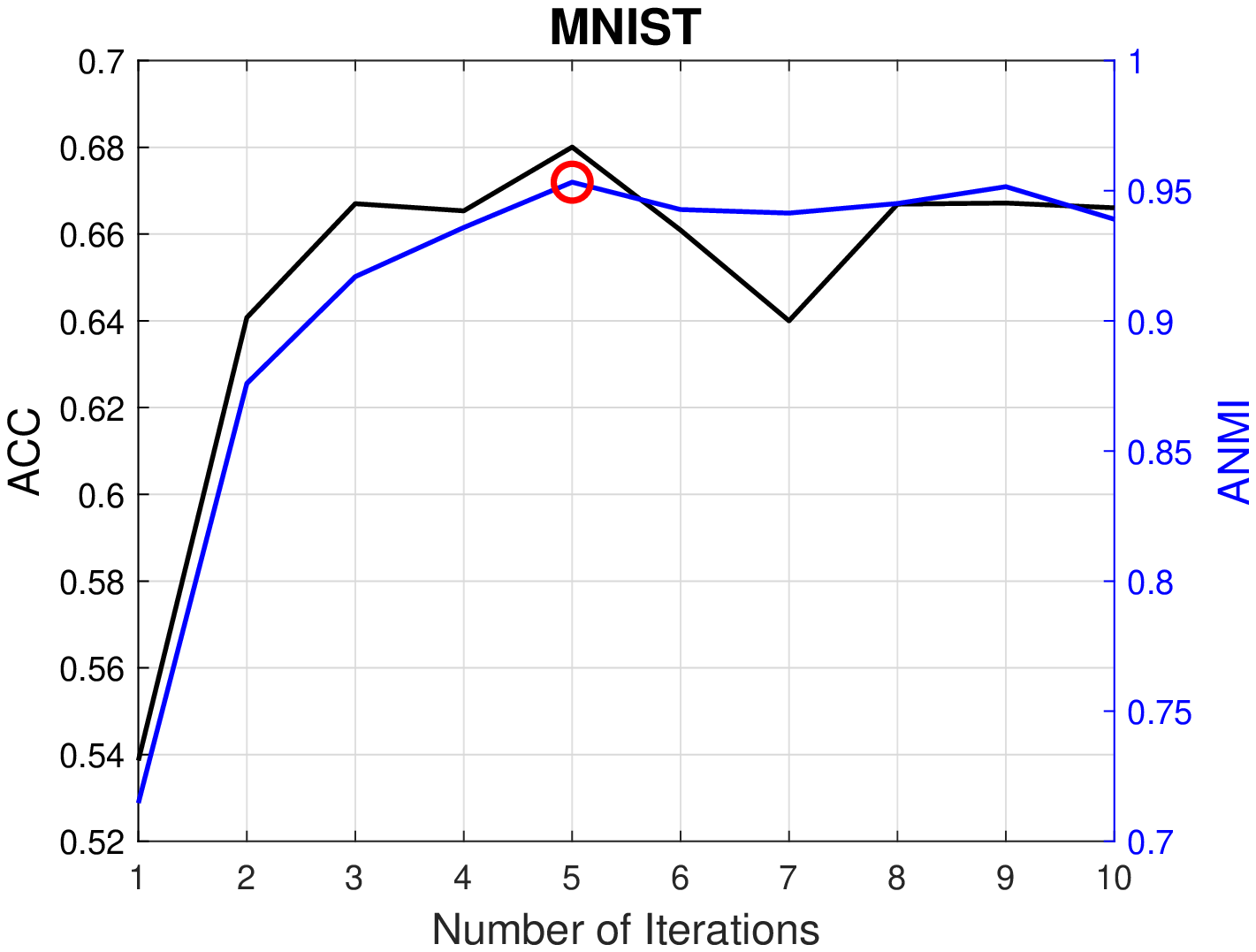,width=3.7cm}}
		%		\vspace{-0.5cm}
		%       \centerline{(a) Result 1}\medskip
	\end{minipage}
	\begin{minipage}[b]{0.195\linewidth}
		\centering
		\centerline{\epsfig{figure=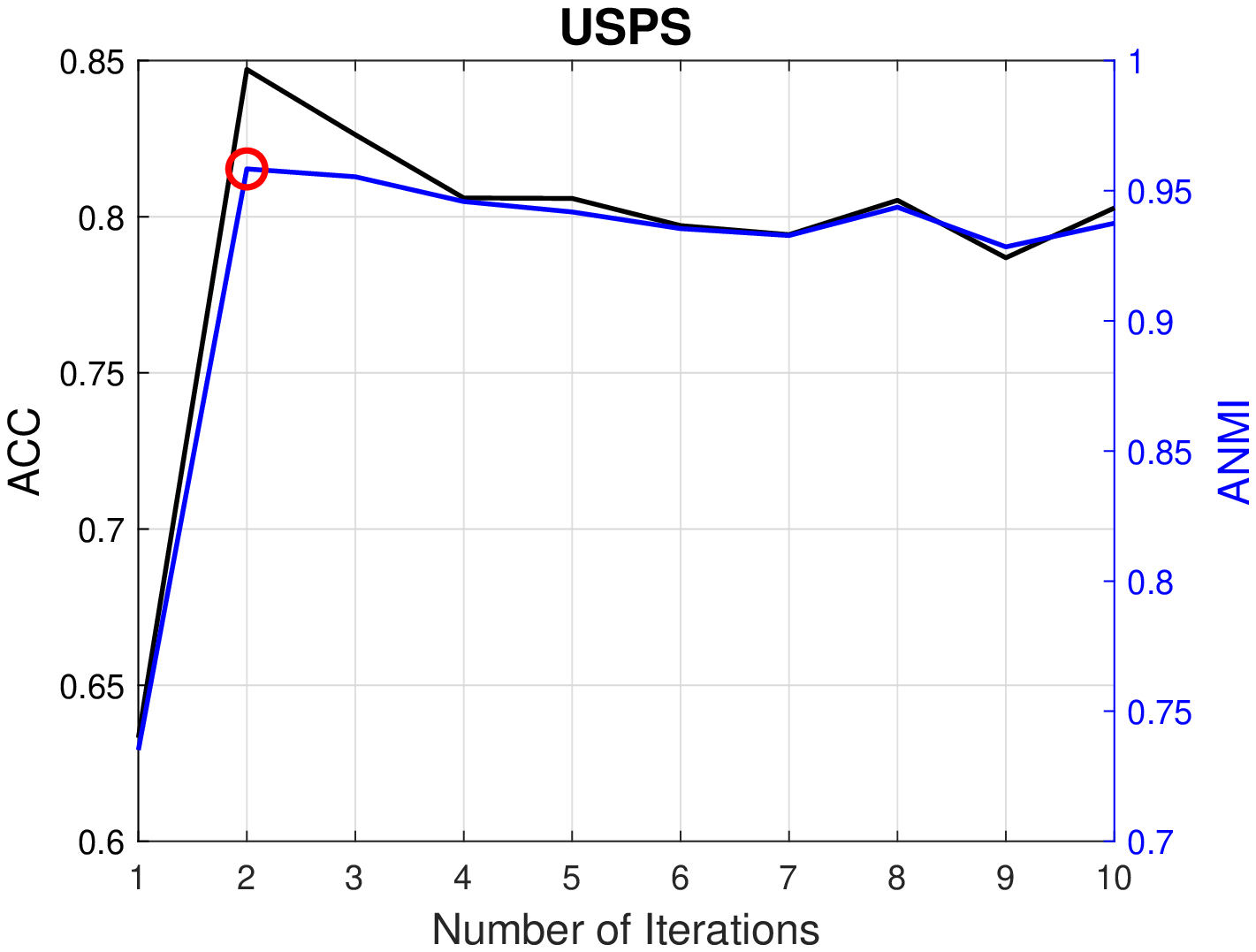,width=3.7cm}}
		%		\vspace{-0.5cm}
		%       \centerline{(a) Result 1}\medskip
	\end{minipage}
	\begin{minipage}[b]{0.195\linewidth}
		\centering
		\centerline{\epsfig{figure=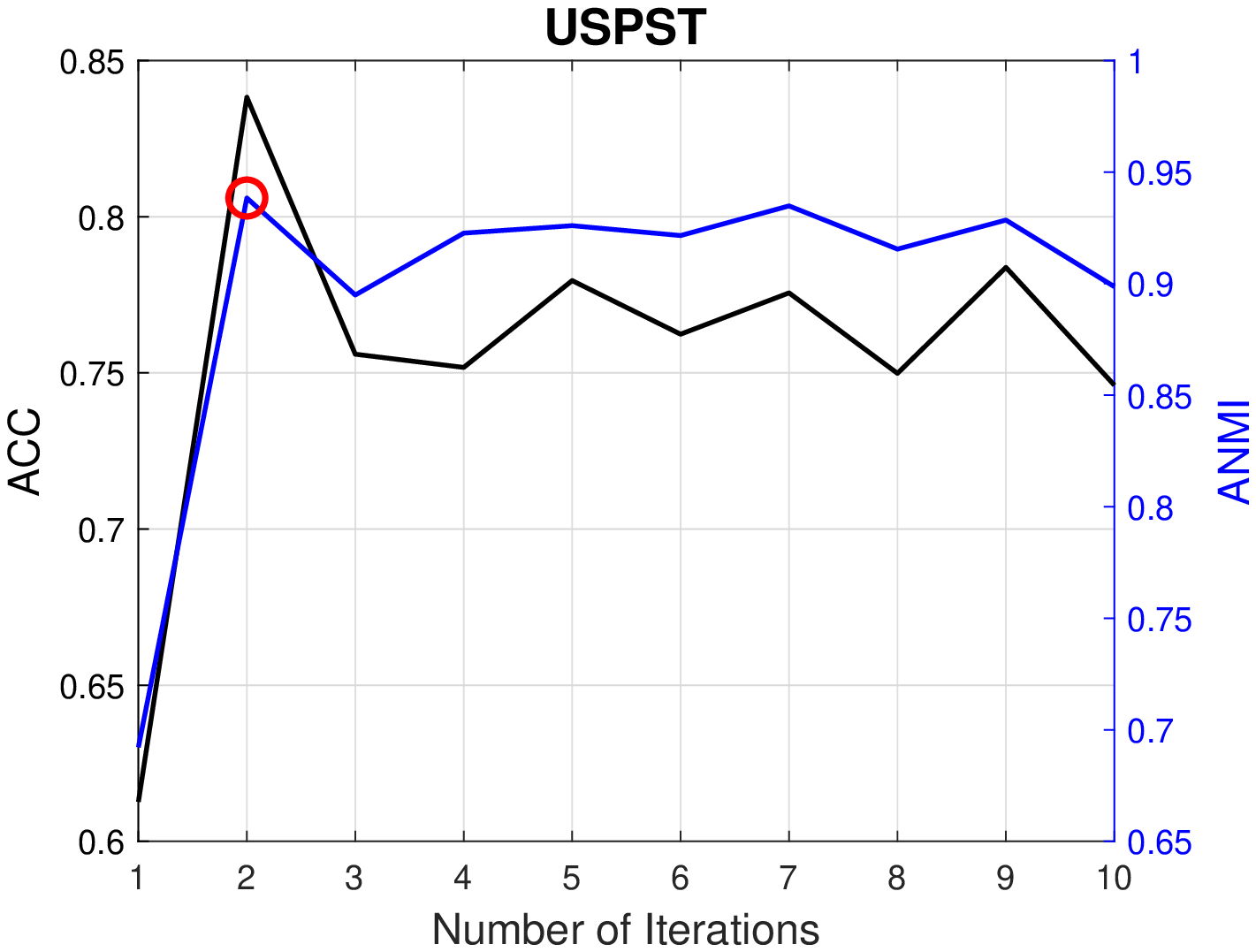,width=3.7cm}}
		%		\vspace{-0.5cm}
		%       \centerline{(a) Result 1}\medskip
	\end{minipage}
	\begin{minipage}[b]{0.195\linewidth}
		\centering
		\centerline{\epsfig{figure=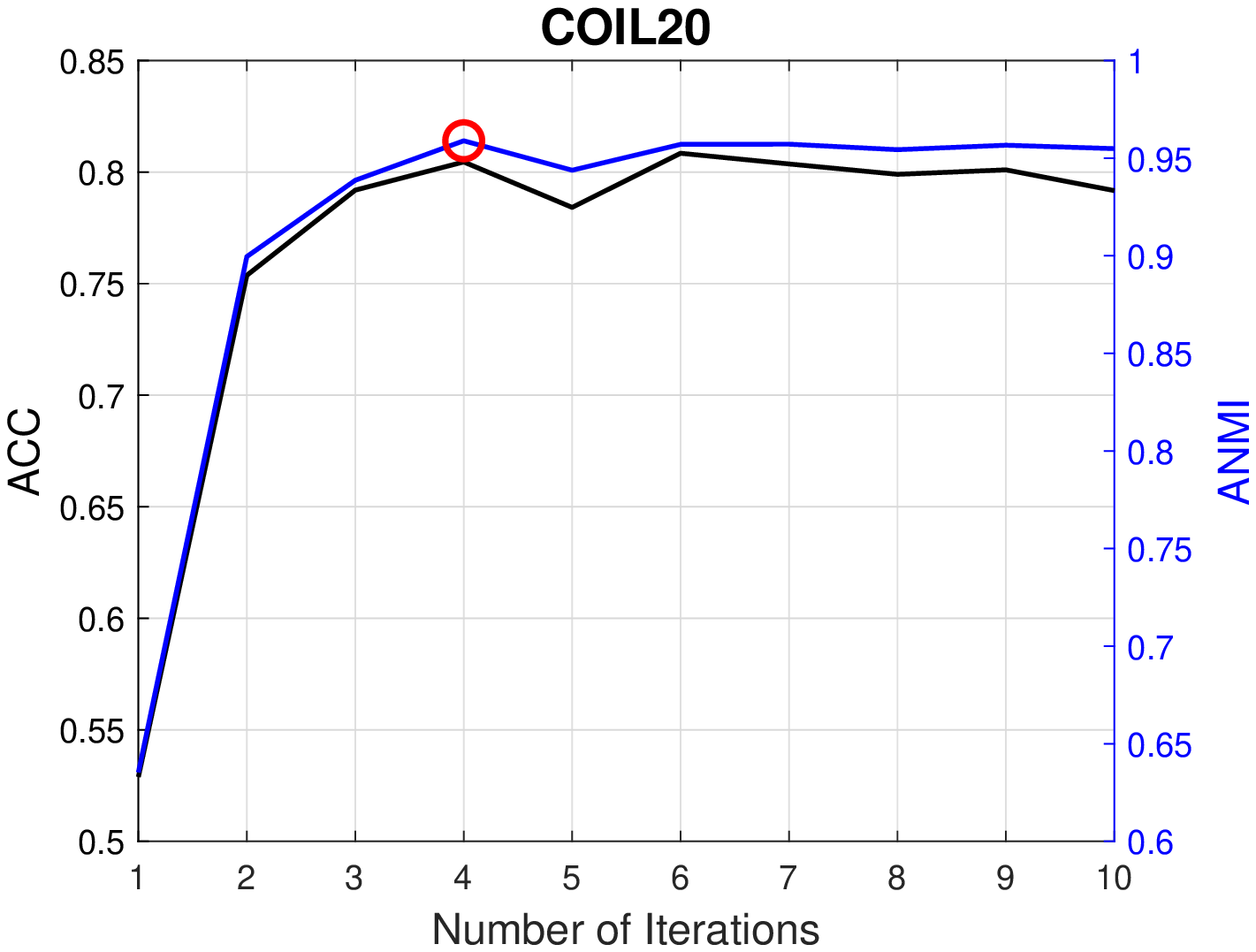,width=3.7cm}}
		%		\vspace{-0.5cm}
		%       \centerline{(a) Result 1}\medskip
	\end{minipage}
	\begin{minipage}[b]{0.195\linewidth}
		\centering
		\centerline{\epsfig{figure=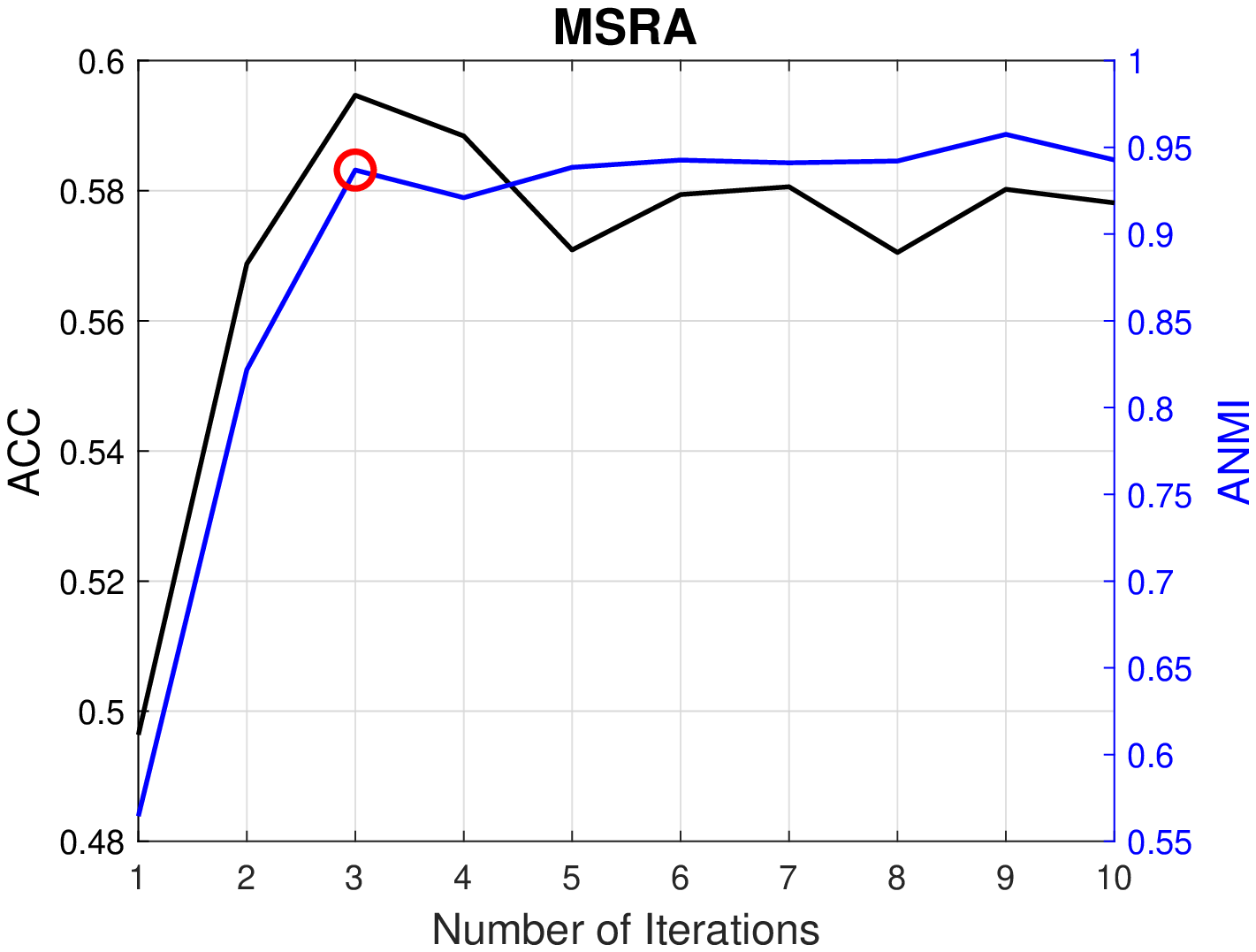,width=3.7cm}}
		%		\vspace{-0.5cm}
		%       \centerline{(a) Result 1}\medskip
	\end{minipage}
	\caption{The black curves indicate the ACC of the proposed model against different iteration number of Algorithm 1. The blue curves draw the ANMI, i.e., the value of the proposed stopping criterion of Algorithm 1.   
		The red circle on the blue curve indicates that the proposed stopping criterion is achieved at that point.  }
	\label{fig:ANMI}
\end{figure*}

\begin{figure*}[t]
	\begin{minipage}[b]{0.195\linewidth}
		\centering
		\centerline{\epsfig{figure=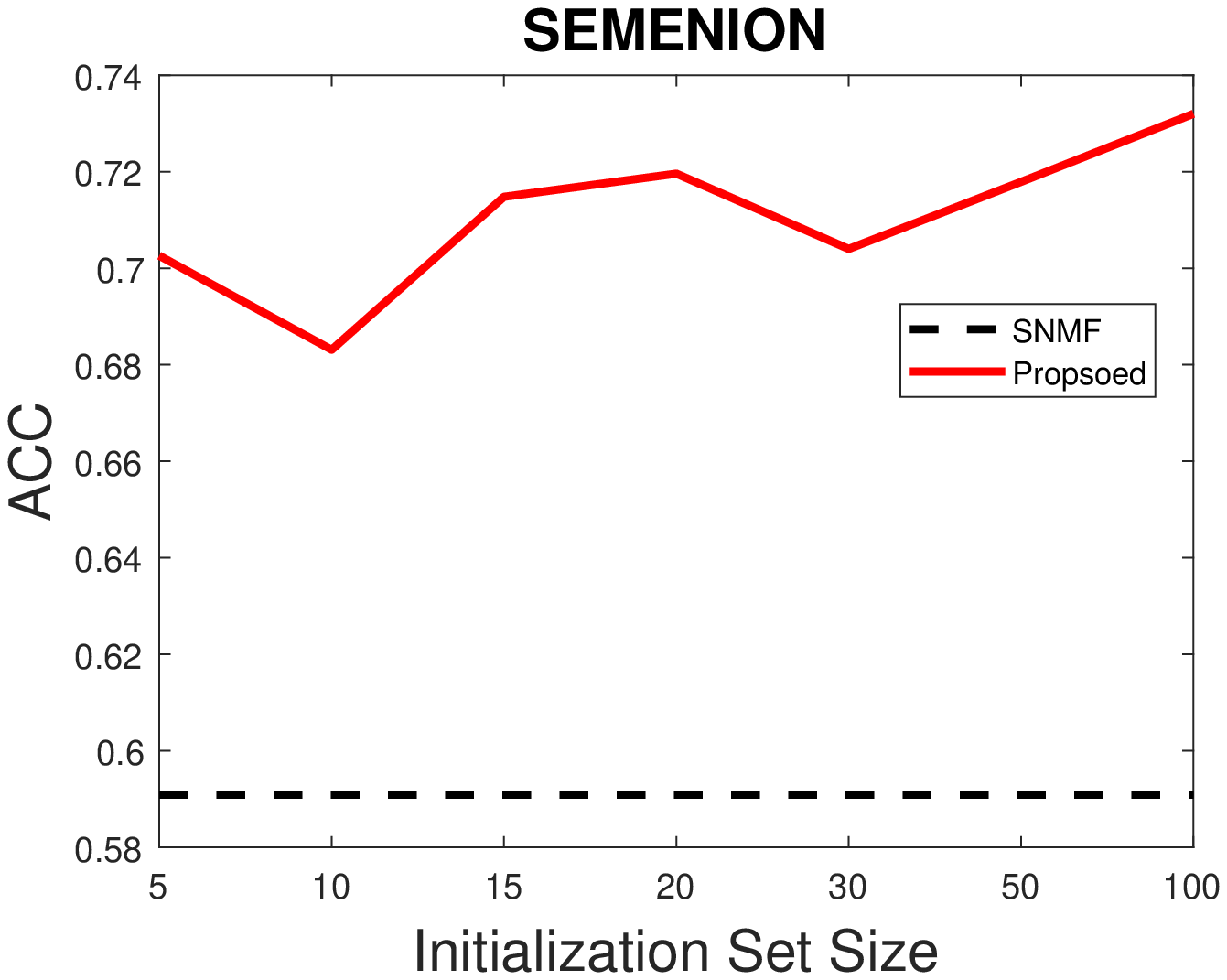,width=3.7cm}}
		%		\vspace{-0.1cm}
		%       \centerline{(a) Result 1}\medskip
	\end{minipage}
	\begin{minipage}[b]{0.195\linewidth}
		\centering
				\centerline{\epsfig{figure=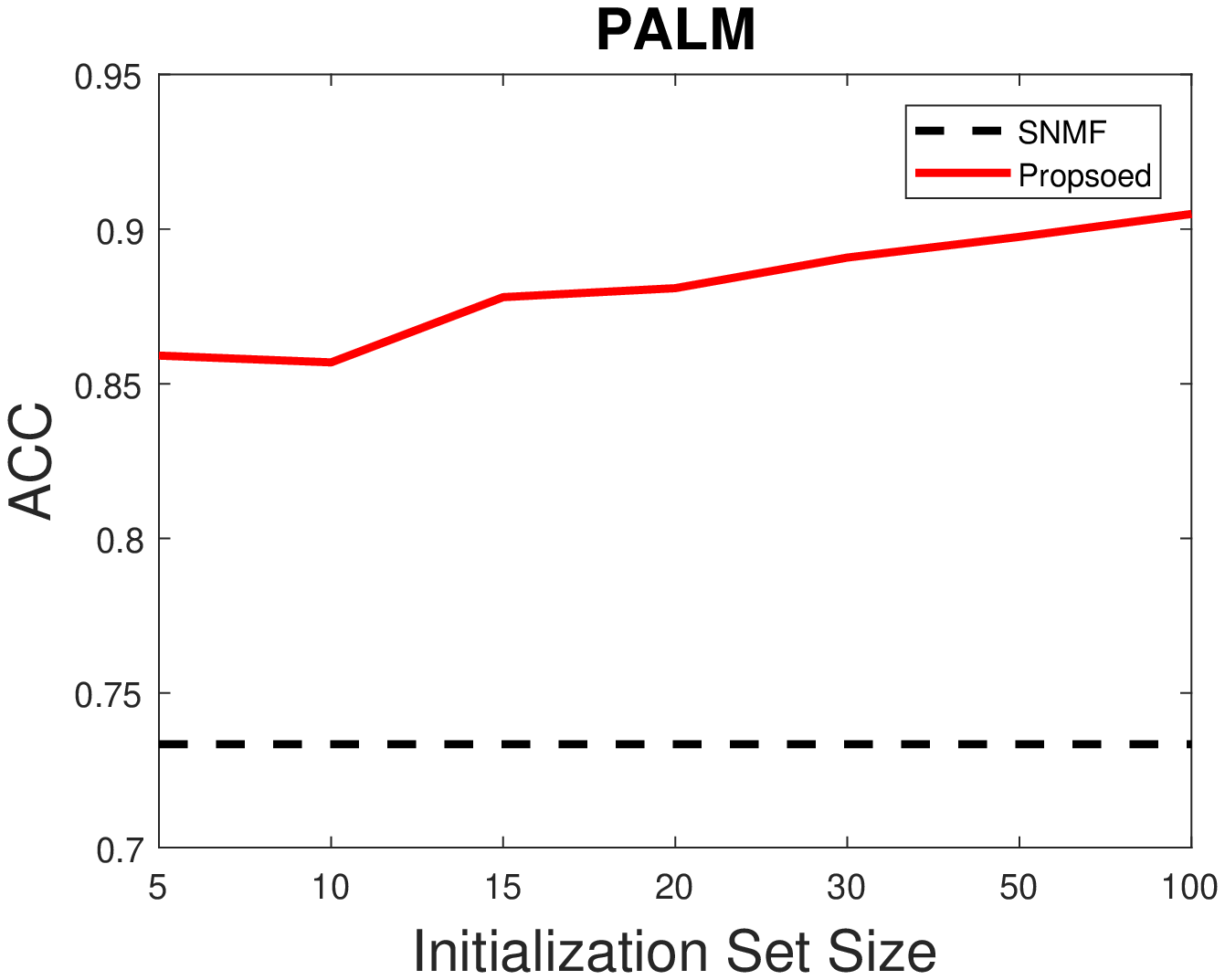,width=3.7cm}}
		%		\vspace{-0.1cm}
		%       \centerline{(a) Result 1}\medskip
	\end{minipage}
	\begin{minipage}[b]{0.195\linewidth}
		\centering
		\centerline{\epsfig{figure=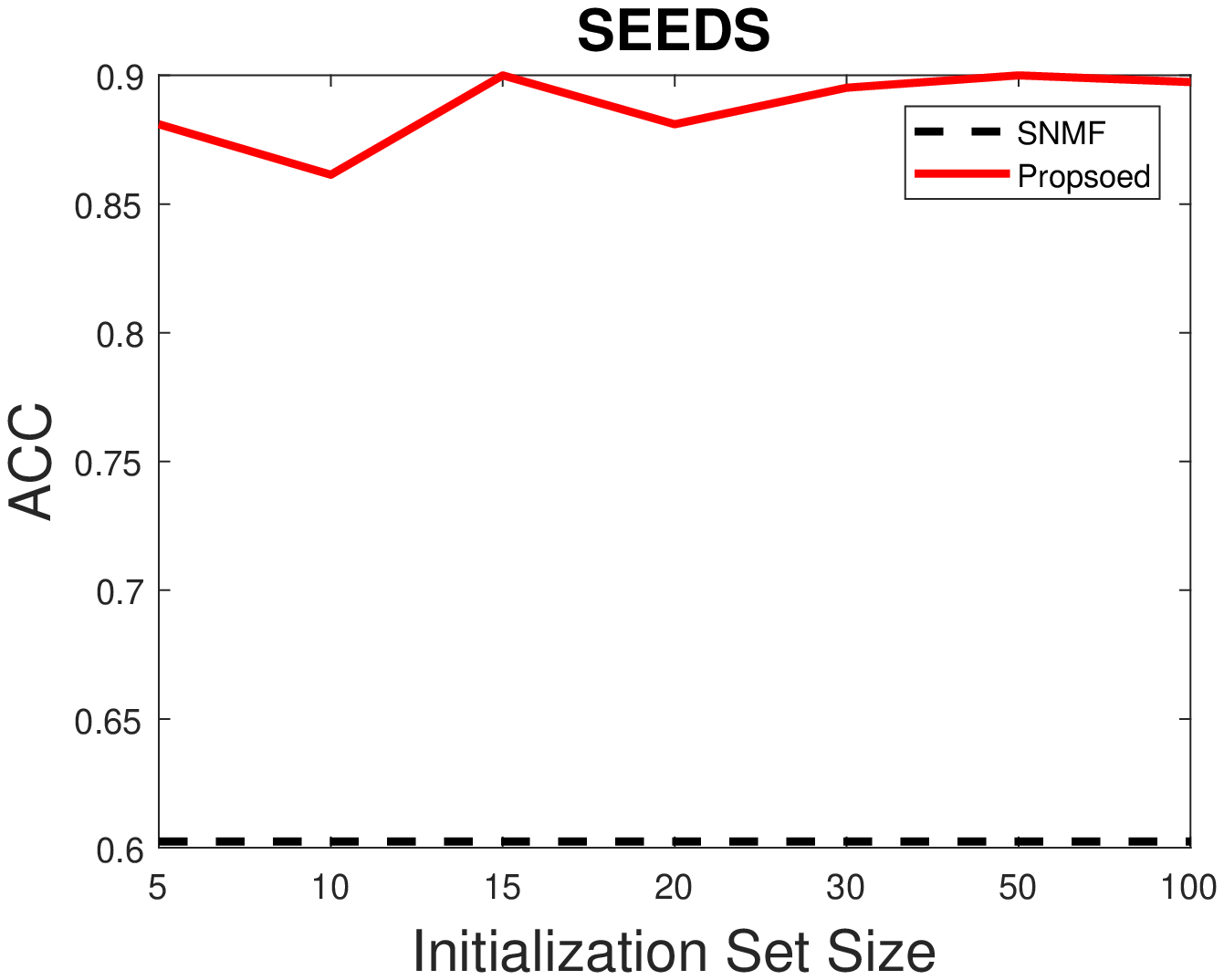,width=3.7cm}}
		%		\vspace{-0.1cm}
		%       \centerline{(a) Result 1}\medskip
	\end{minipage}
	\begin{minipage}[b]{0.195\linewidth}
		\centering
		\centerline{\epsfig{figure=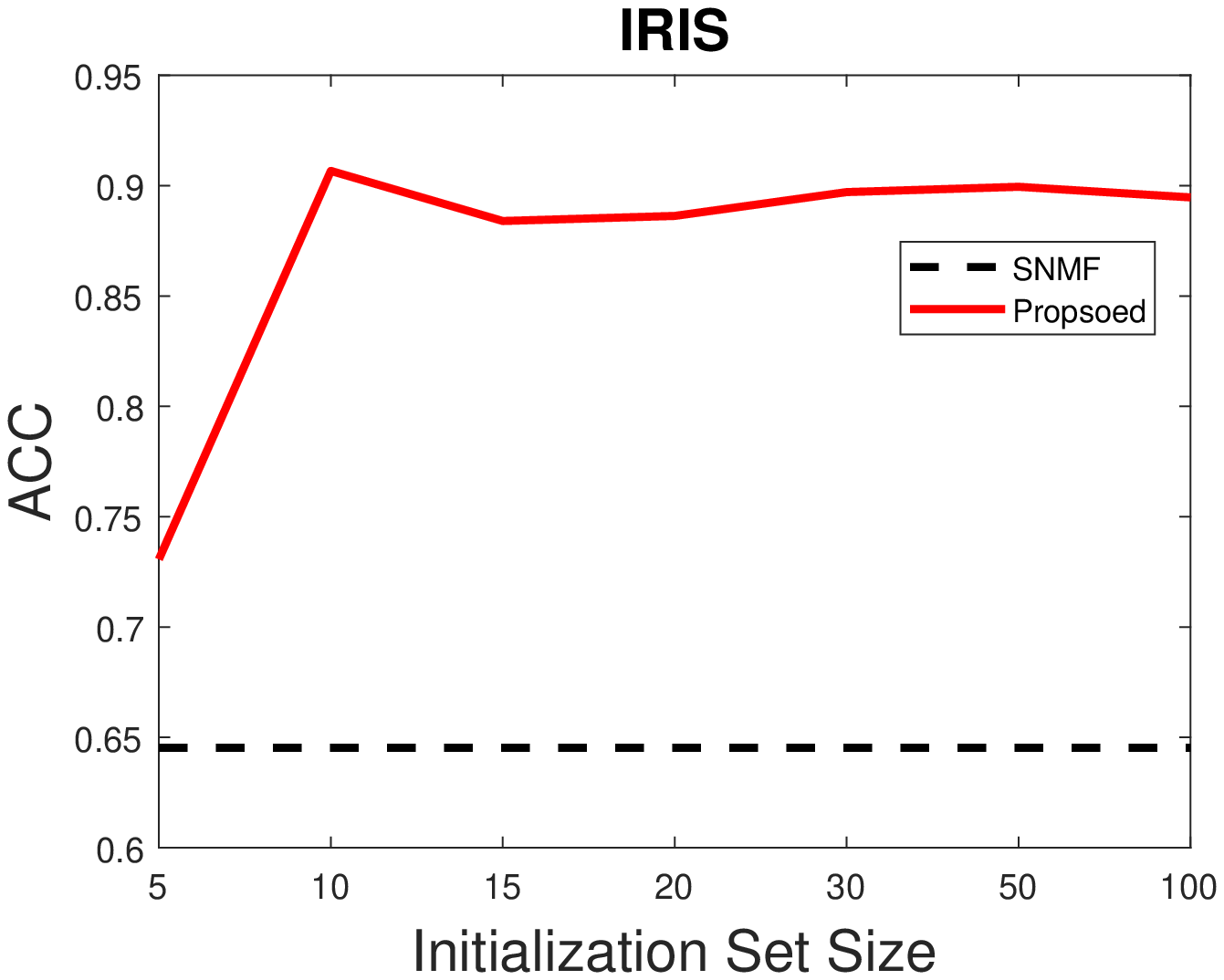,width=3.7cm}}
		%		\vspace{-0.5cm}
		%       \centerline{(a) Result 1}\medskip
	\end{minipage}
	\begin{minipage}[b]{0.195\linewidth}
		\centering
		\centerline{\epsfig{figure=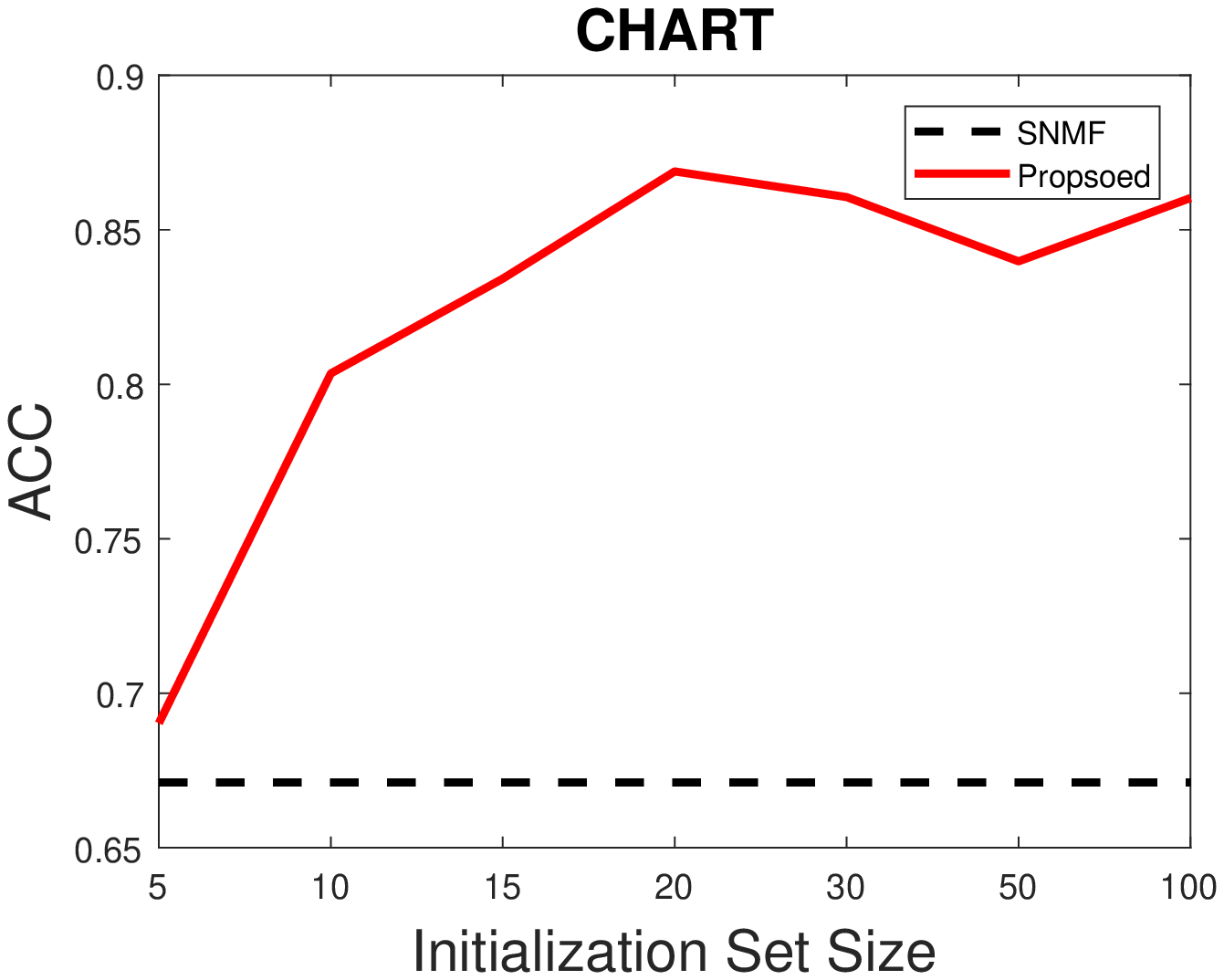,width=3.7cm}}
		%		\vspace{-0.5cm}
		%       \centerline{(a) Result 1}\medskip
	\end{minipage}\\
	\begin{minipage}[b]{0.195\linewidth}
		\centering
		\centerline{\epsfig{figure=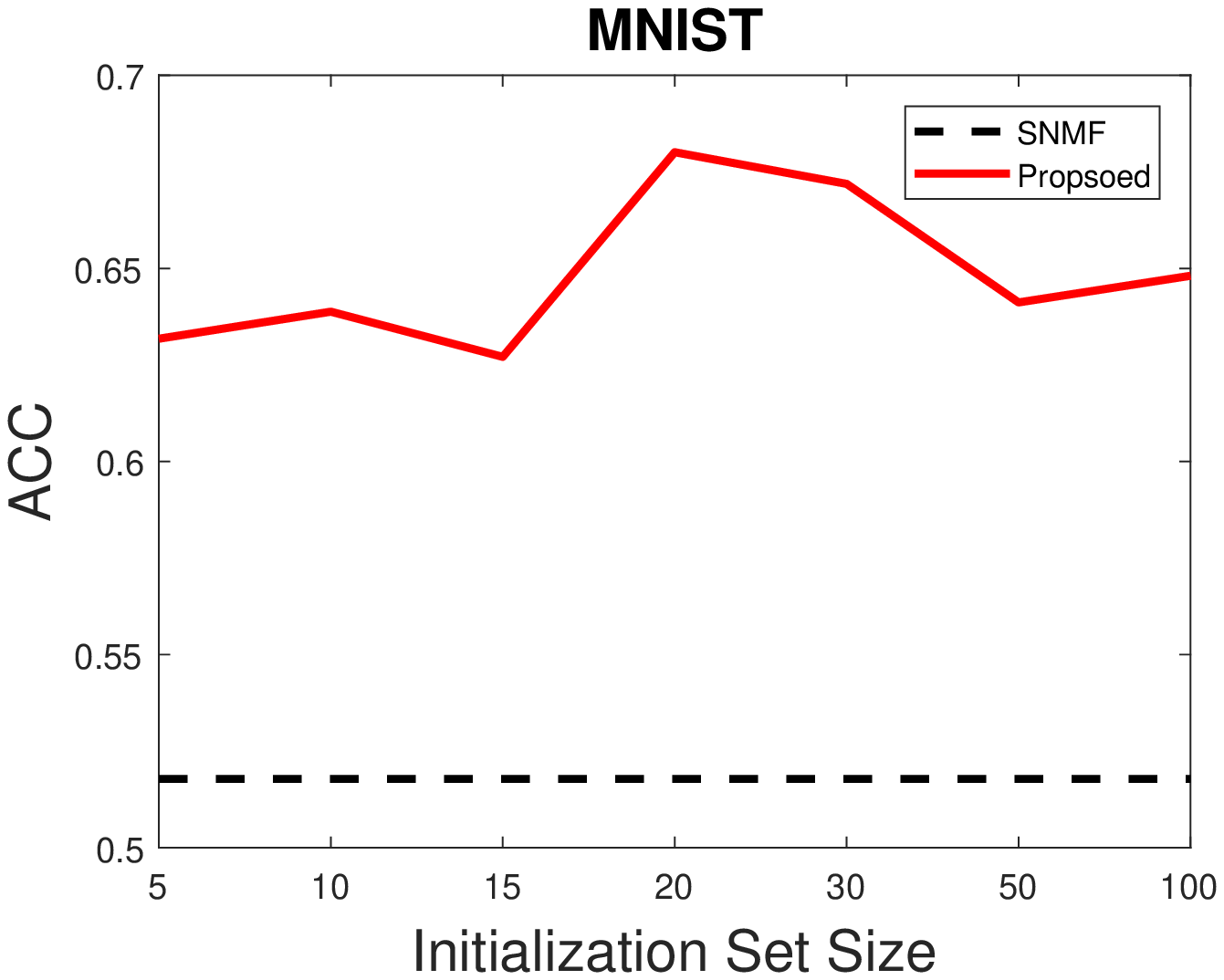,width=3.7cm}}
		%		\vspace{-0.5cm}
		%       \centerline{(a) Result 1}\medskip
	\end{minipage}
	\begin{minipage}[b]{0.195\linewidth}
		\centering
		\centerline{\epsfig{figure=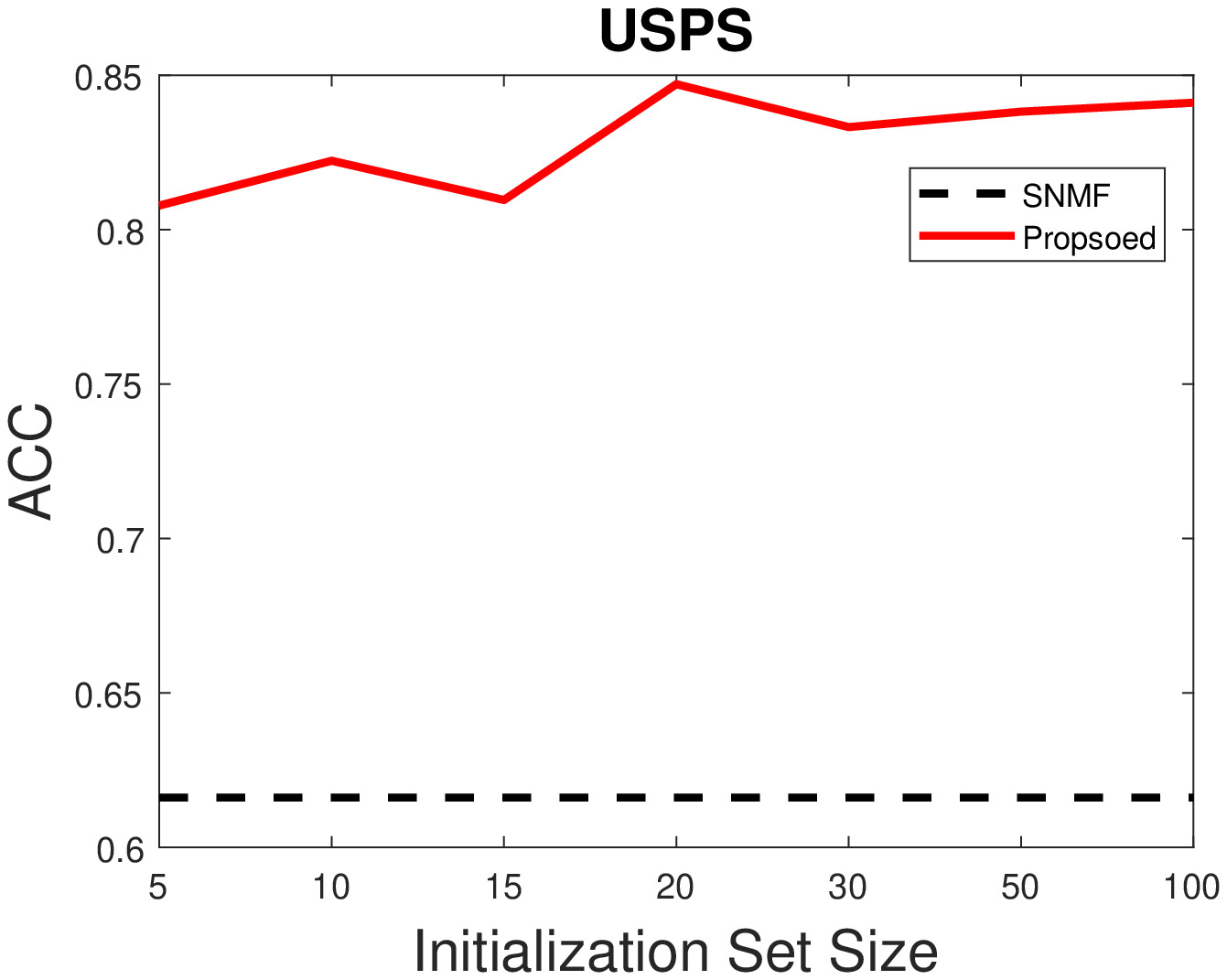,width=3.7cm}}
		%		\vspace{-0.5cm}
		%       \centerline{(a) Result 1}\medskip
	\end{minipage}
	\begin{minipage}[b]{0.195\linewidth}
		\centering
				\centerline{\epsfig{figure=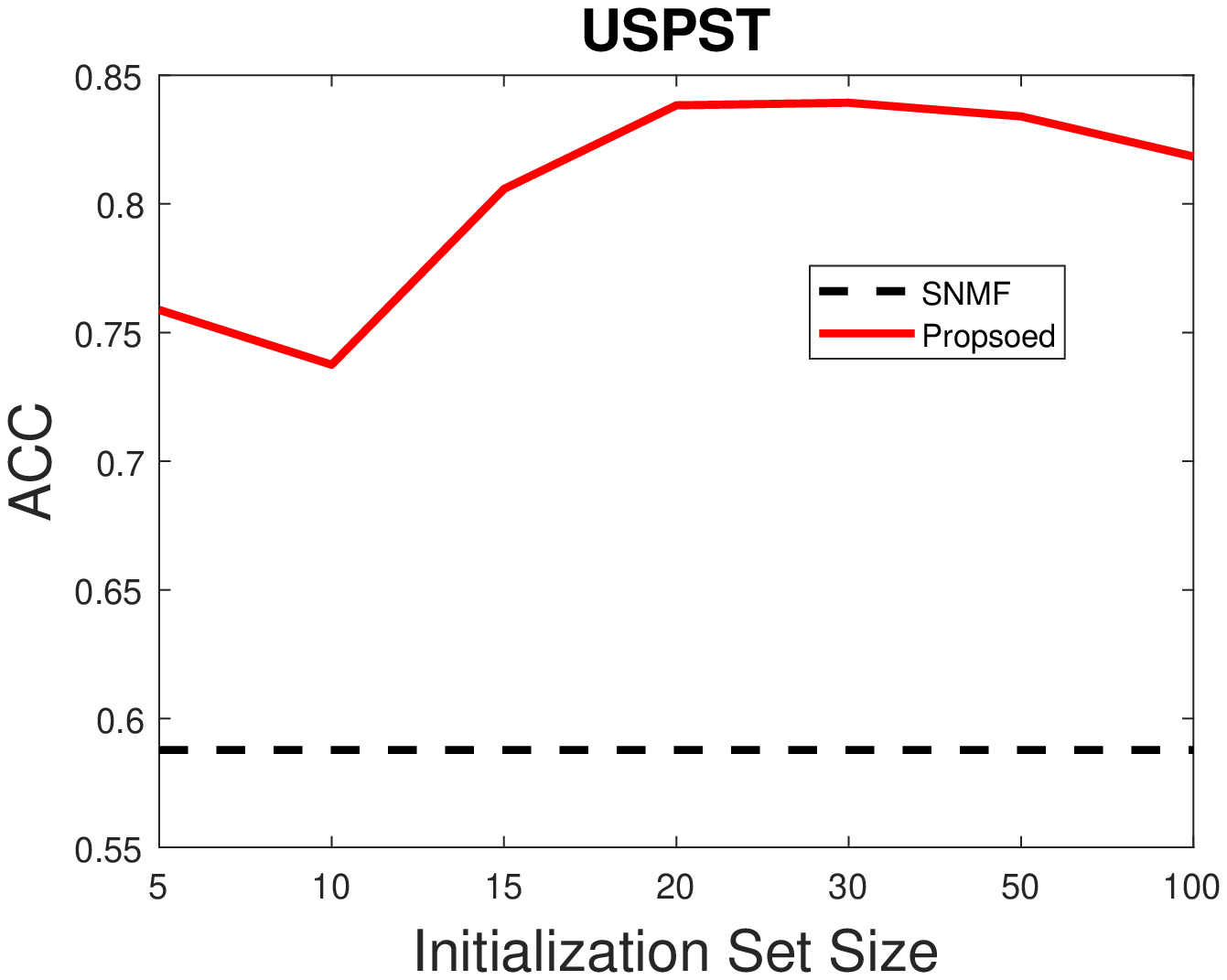,width=3.7cm}}
		%		\vspace{-0.5cm}
		%       \centerline{(a) Result 1}\medskip
	\end{minipage}
	\begin{minipage}[b]{0.195\linewidth}
		\centering
		\centerline{\epsfig{figure=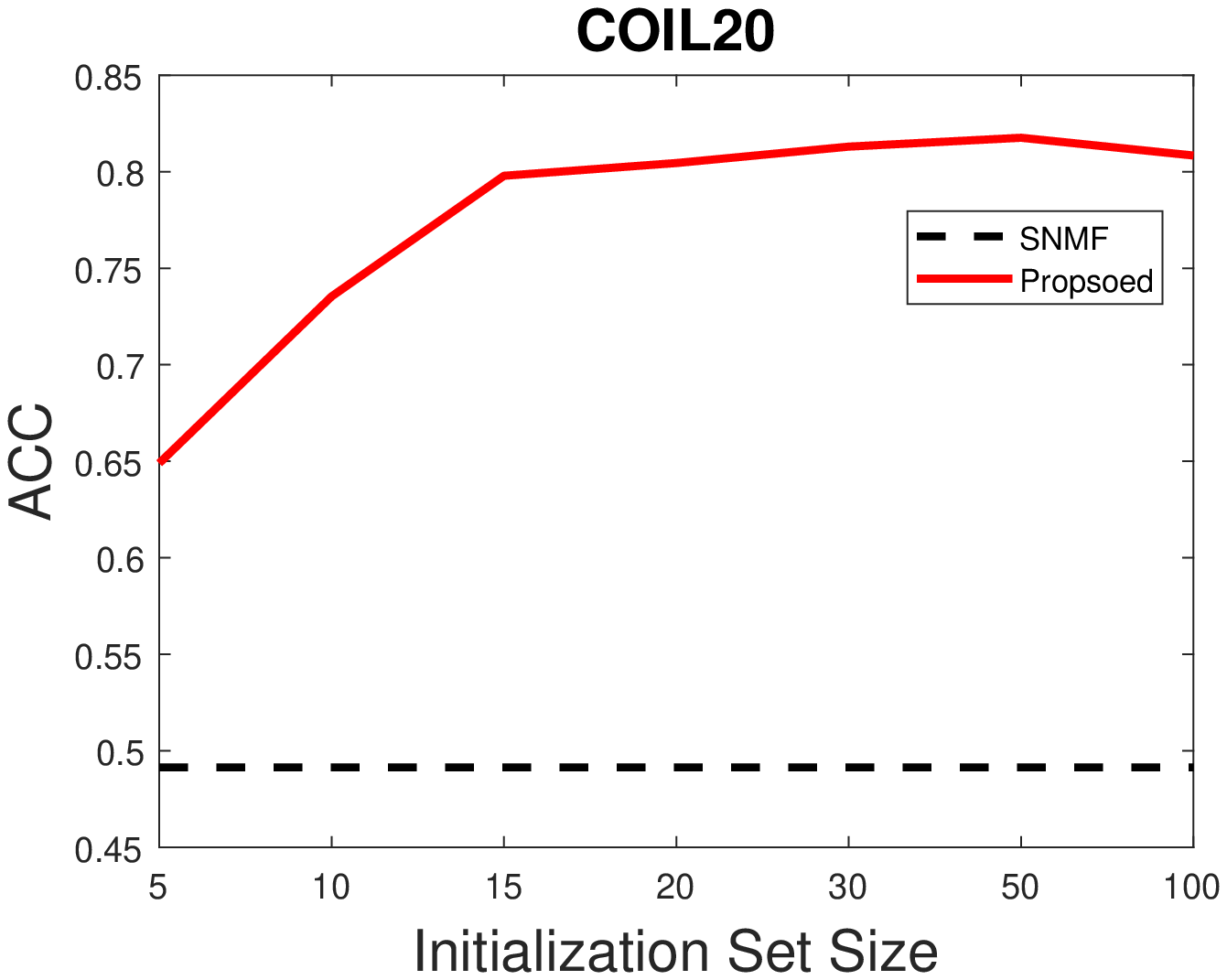,width=3.7cm}}
%				\vspace{-0.5cm}
%		       \centerline{(a) Result 1}\medskip
	\end{minipage}
	\begin{minipage}[b]{0.195\linewidth}
		\centering
		\centerline{\epsfig{figure=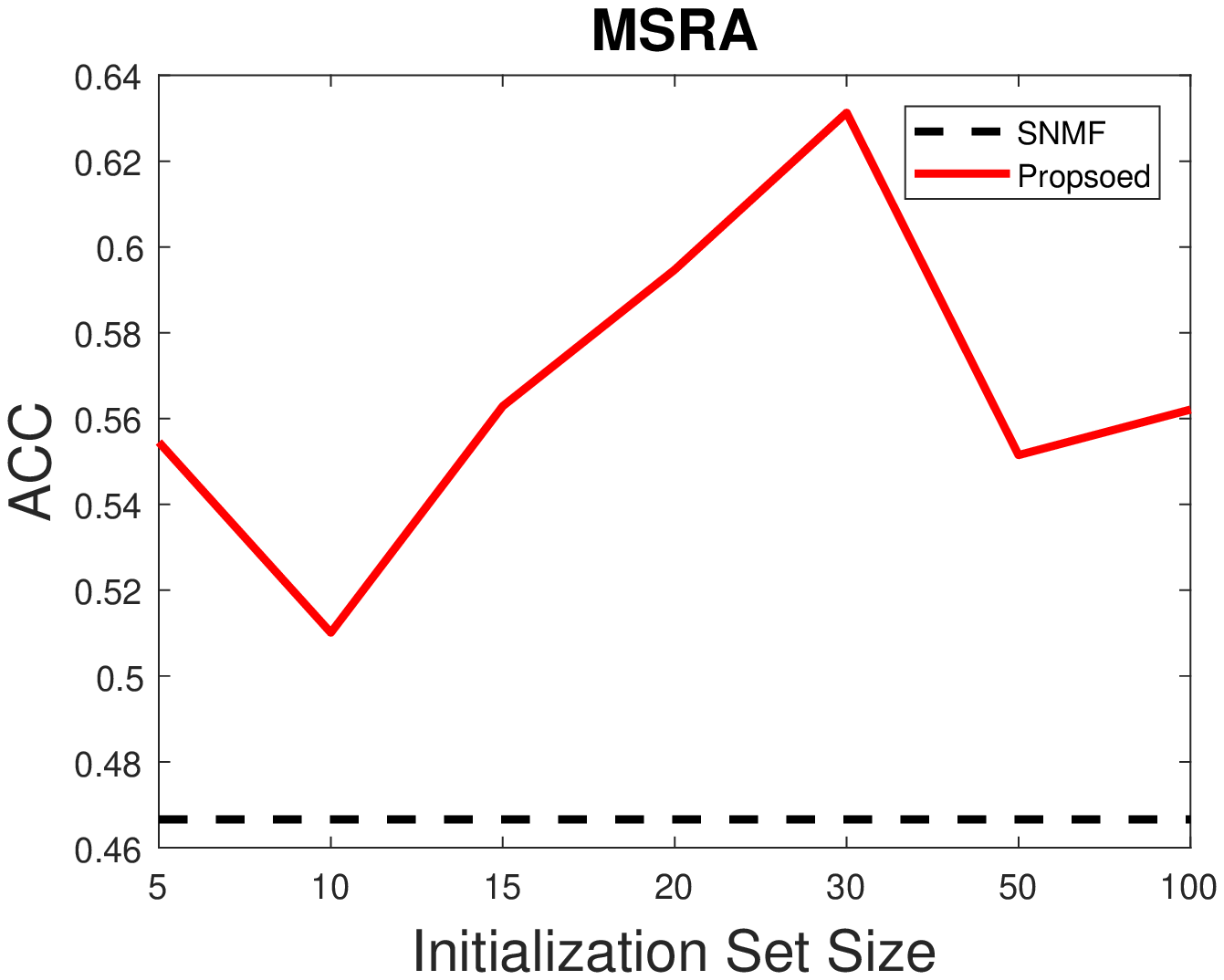,width=3.7cm}}
		%		\vspace{-0.5cm}
		%       \centerline{(a) Result 1}\medskip
	\end{minipage}
	\caption{The ACC of the proposed model with different sizes of the initialization set on all the datasets. The black dash line refers to  the ACC of SNMF as a reference.  }
	\label{fig:en}
\end{figure*}

\begin{figure*}[t]
	\begin{minipage}[b]{0.195\linewidth}
		\centering
		\centerline{\epsfig{figure=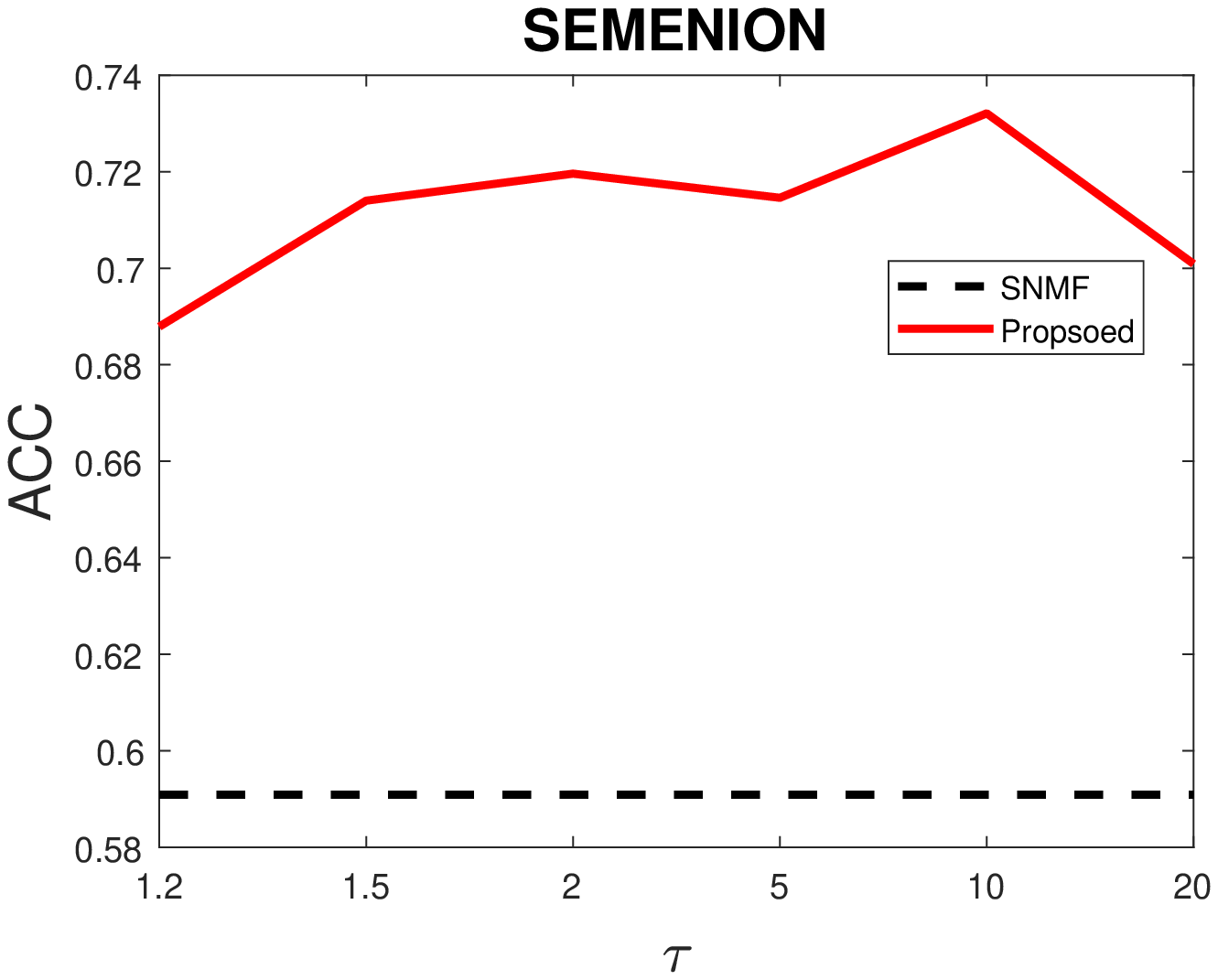,width=3.7cm}}
		%		\vspace{-0.1cm}
		%       \centerline{(a) Result 1}\medskip
	\end{minipage}
	\begin{minipage}[b]{0.195\linewidth}
		\centering
		\centerline{\epsfig{figure=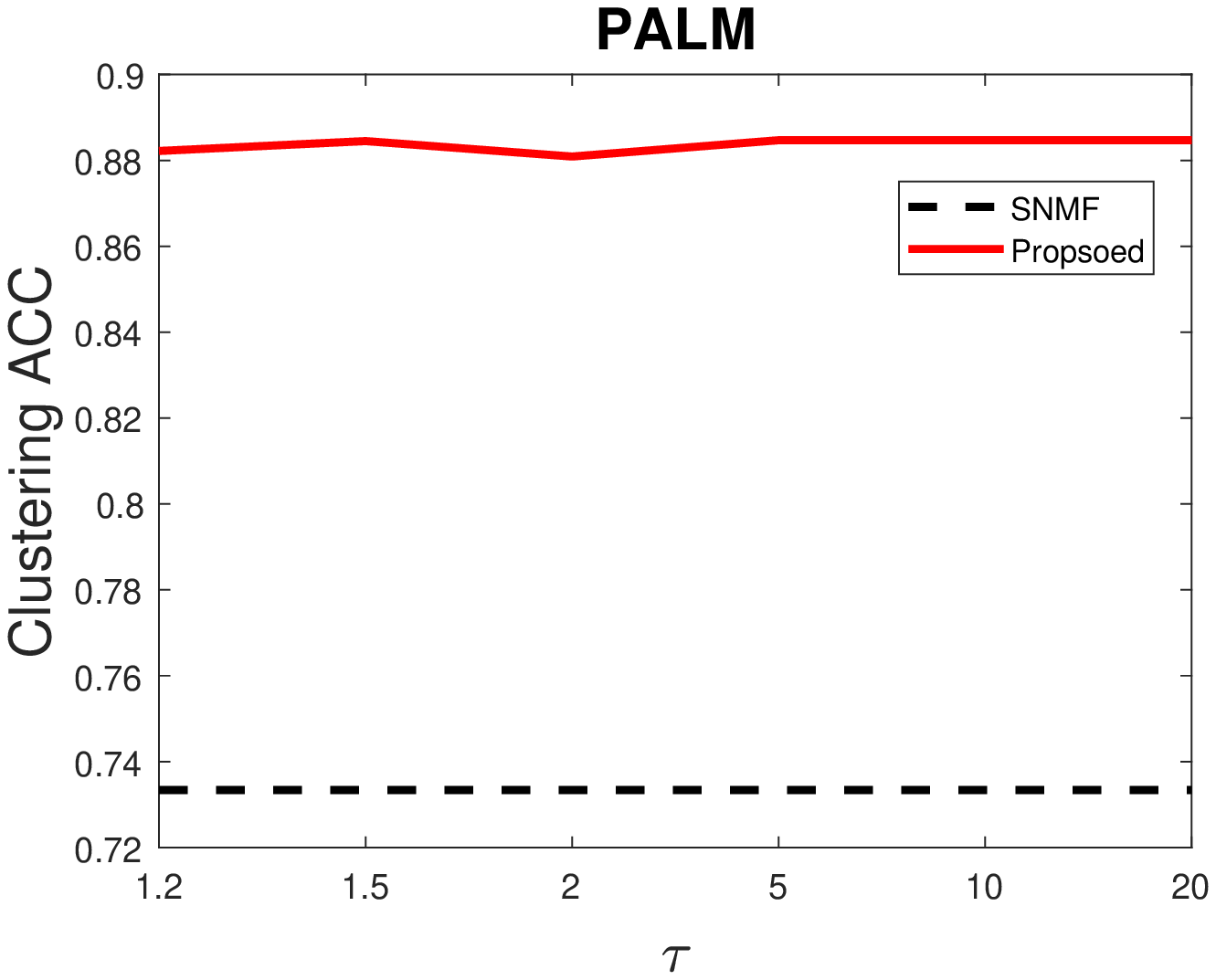,width=3.7cm}}
		%		\vspace{-0.1cm}
		%       \centerline{(a) Result 1}\medskip
	\end{minipage}
	\begin{minipage}[b]{0.195\linewidth}
		\centering
		\centerline{\epsfig{figure=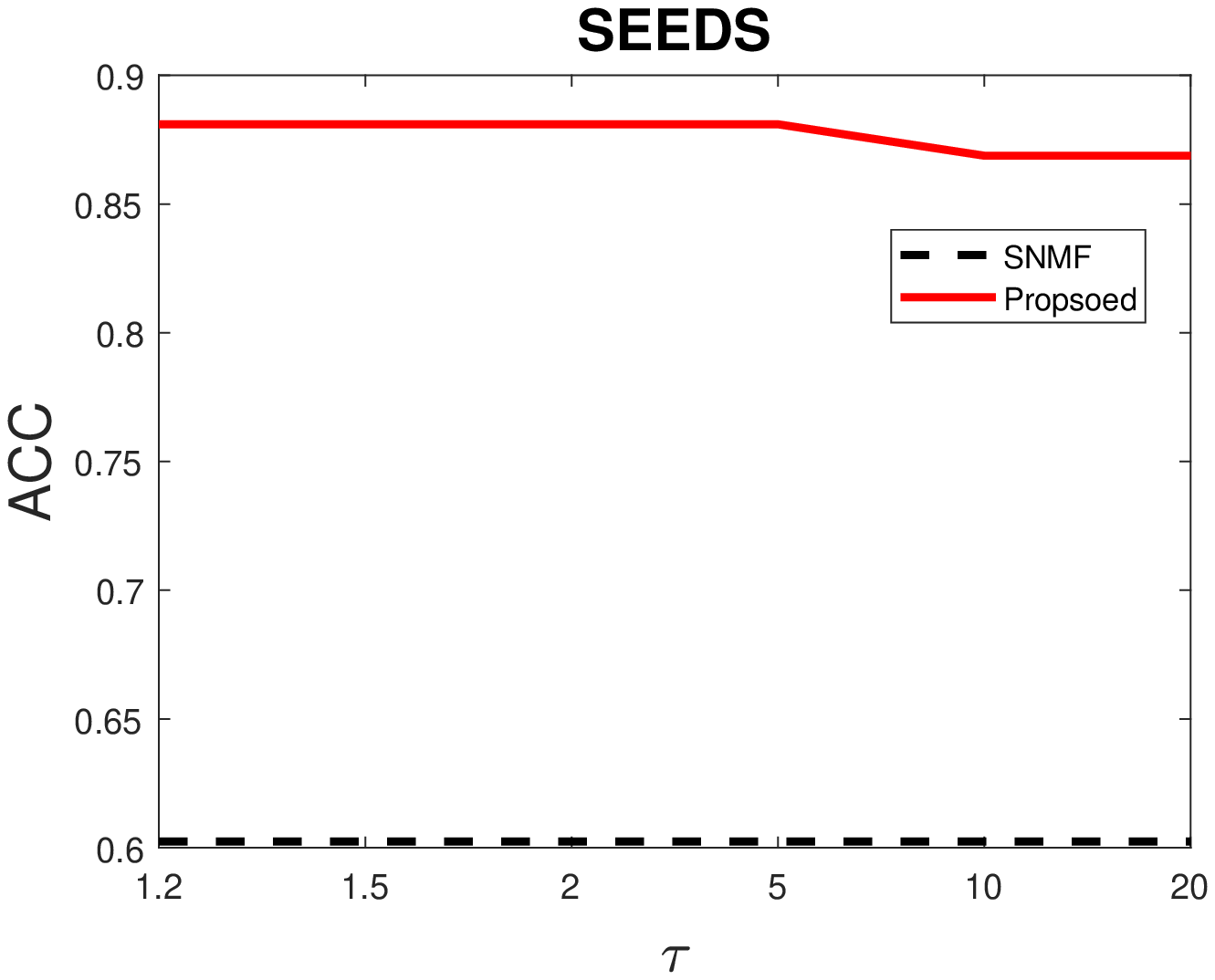,width=3.7cm}}
		%		\vspace{-0.1cm}
		%       \centerline{(a) Result 1}\medskip
	\end{minipage}
	\begin{minipage}[b]{0.195\linewidth}
		\centering
		\centerline{\epsfig{figure=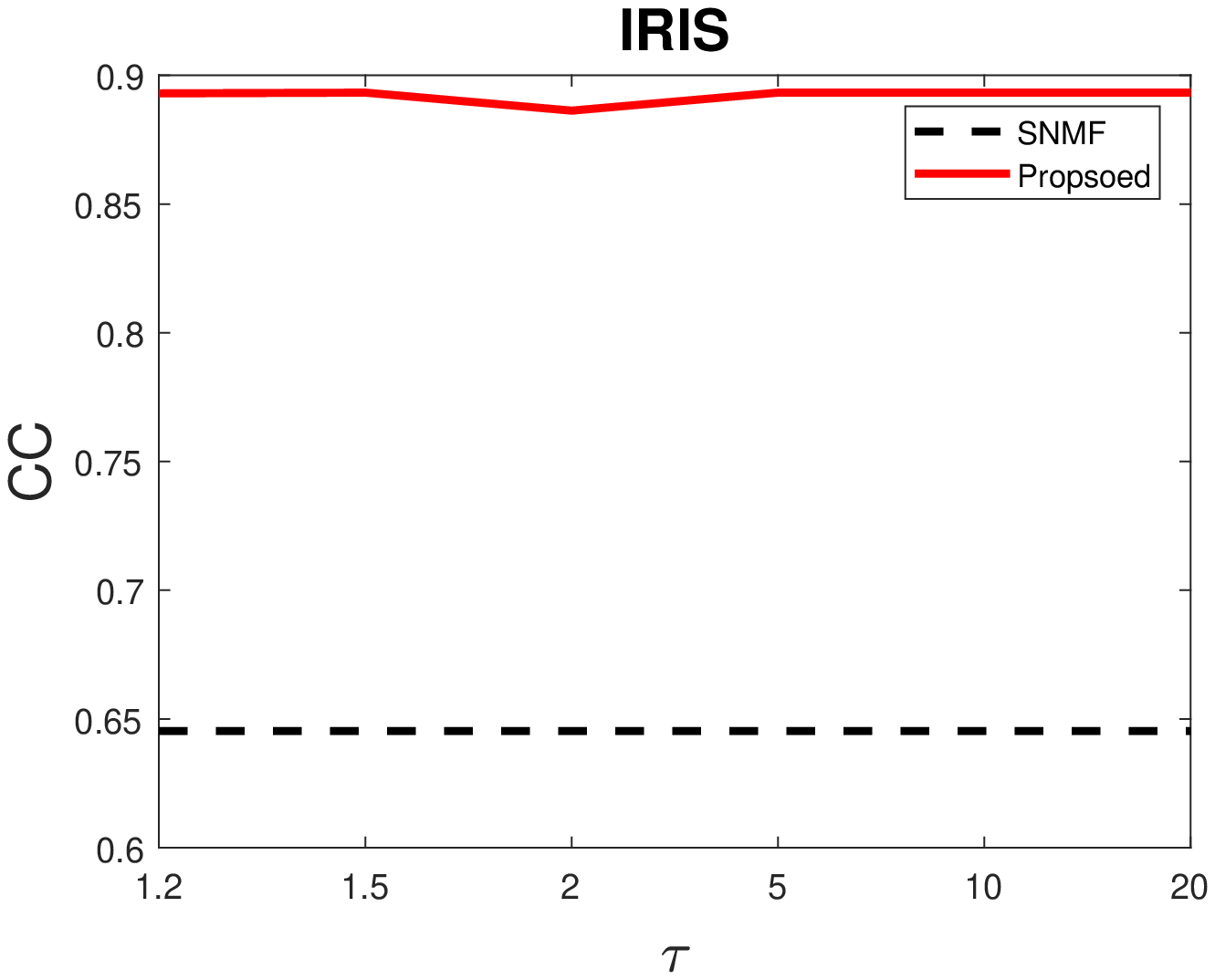,width=3.7cm}}
		%		\vspace{-0.5cm}
		%       \centerline{(a) Result 1}\medskip
	\end{minipage}
	\begin{minipage}[b]{0.195\linewidth}
		\centering
		\centerline{\epsfig{figure=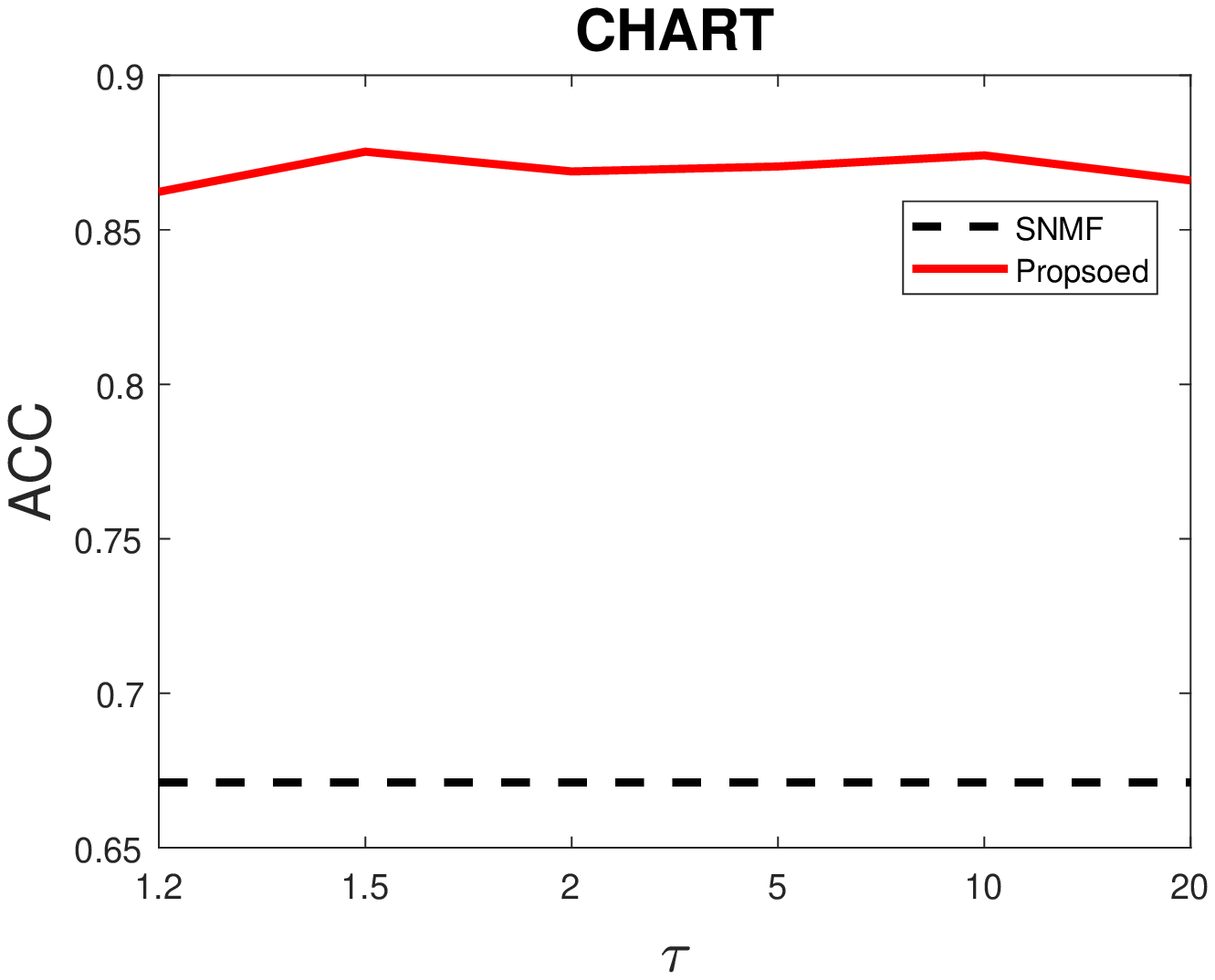,width=3.7cm}}
		%		\vspace{-0.5cm}
		%       \centerline{(a) Result 1}\medskip
	\end{minipage}\\
	\begin{minipage}[b]{0.195\linewidth}
		\centering
		\centerline{\epsfig{figure=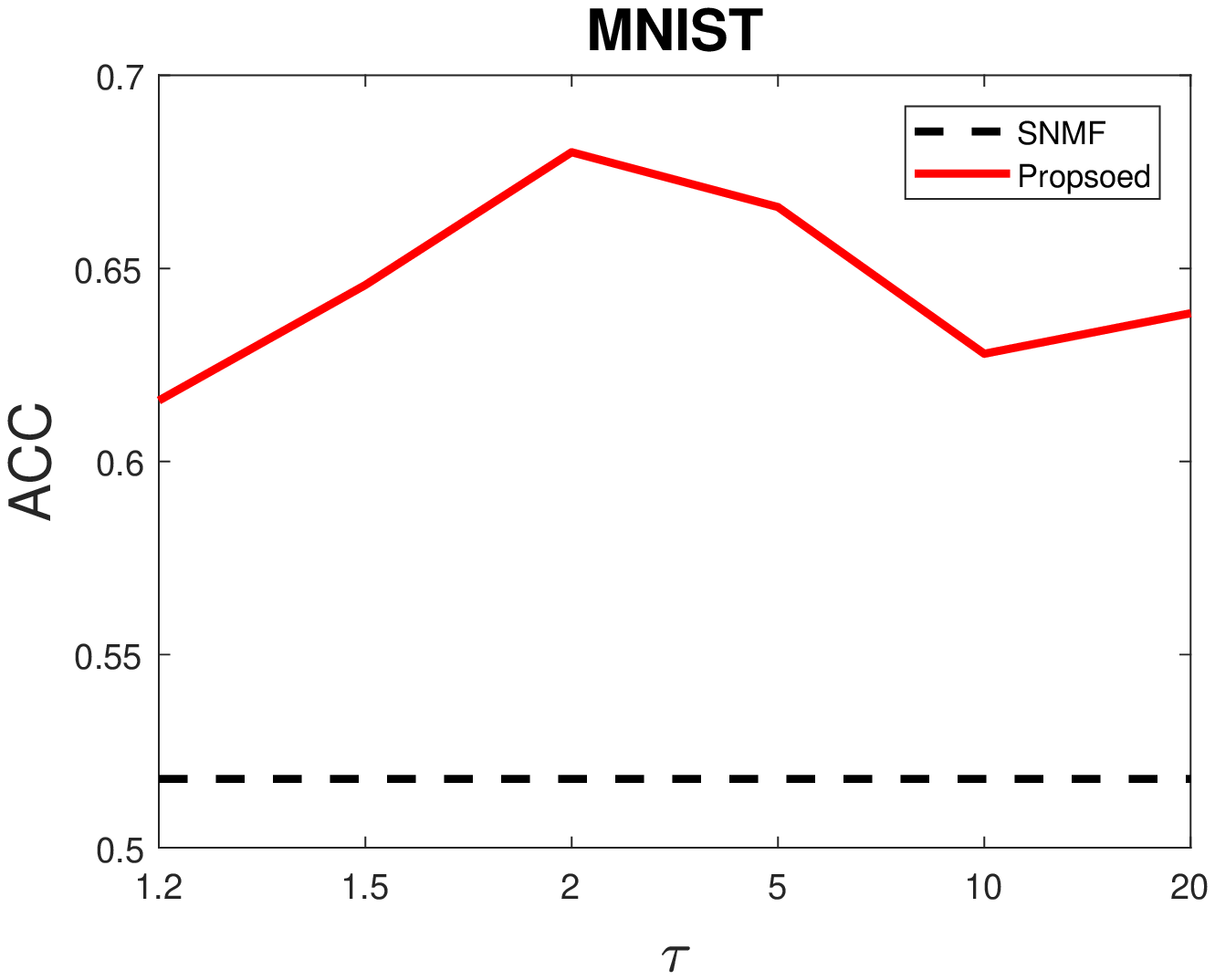,width=3.7cm}}
		%		\vspace{-0.5cm}
		%       \centerline{(a) Result 1}\medskip
	\end{minipage}
	\begin{minipage}[b]{0.195\linewidth}
		\centering
		\centerline{\epsfig{figure=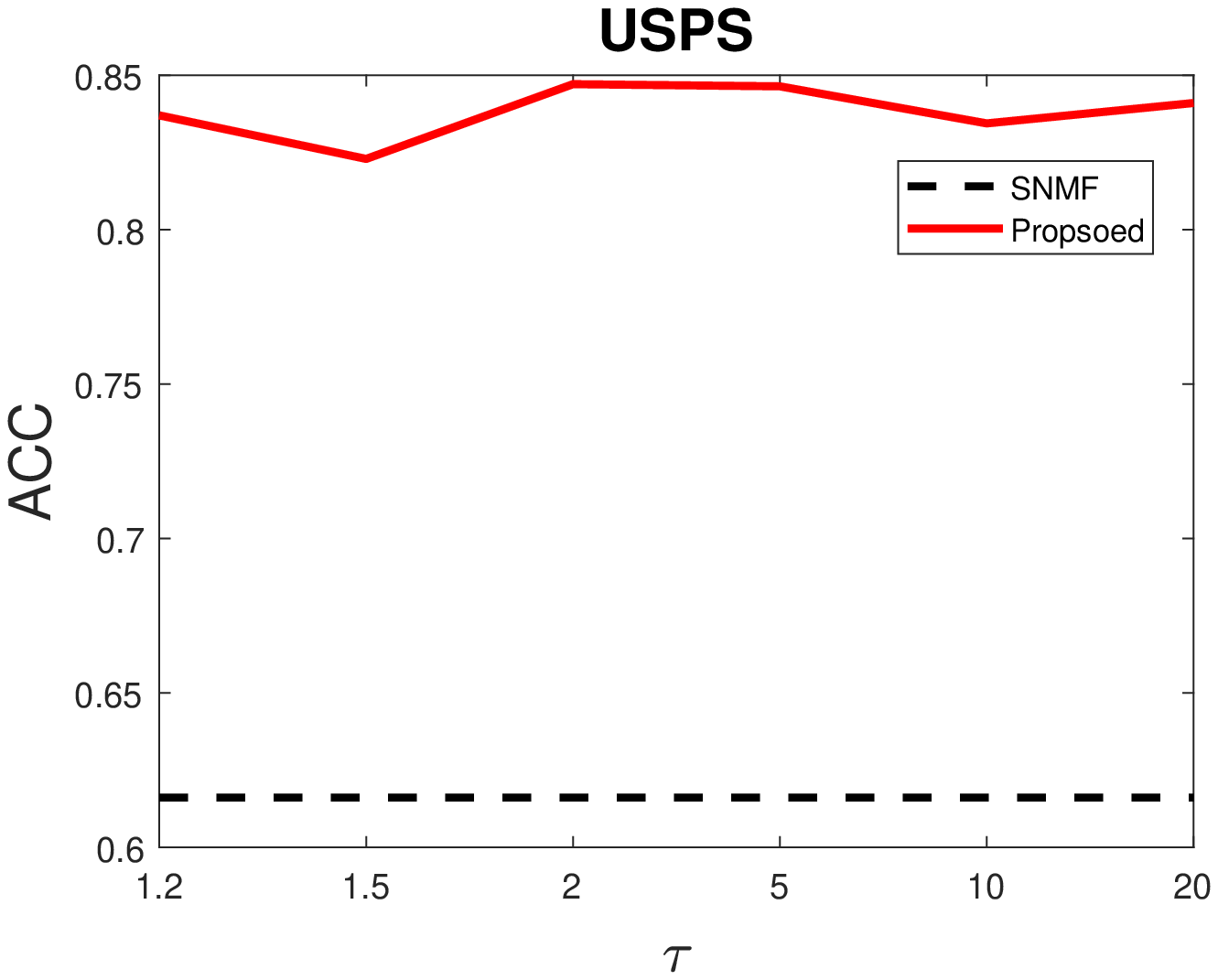,width=3.7cm}}
		%		\vspace{-0.5cm}
		%       \centerline{(a) Result 1}\medskip
	\end{minipage}
	\begin{minipage}[b]{0.195\linewidth}
		\centering
		\centerline{\epsfig{figure=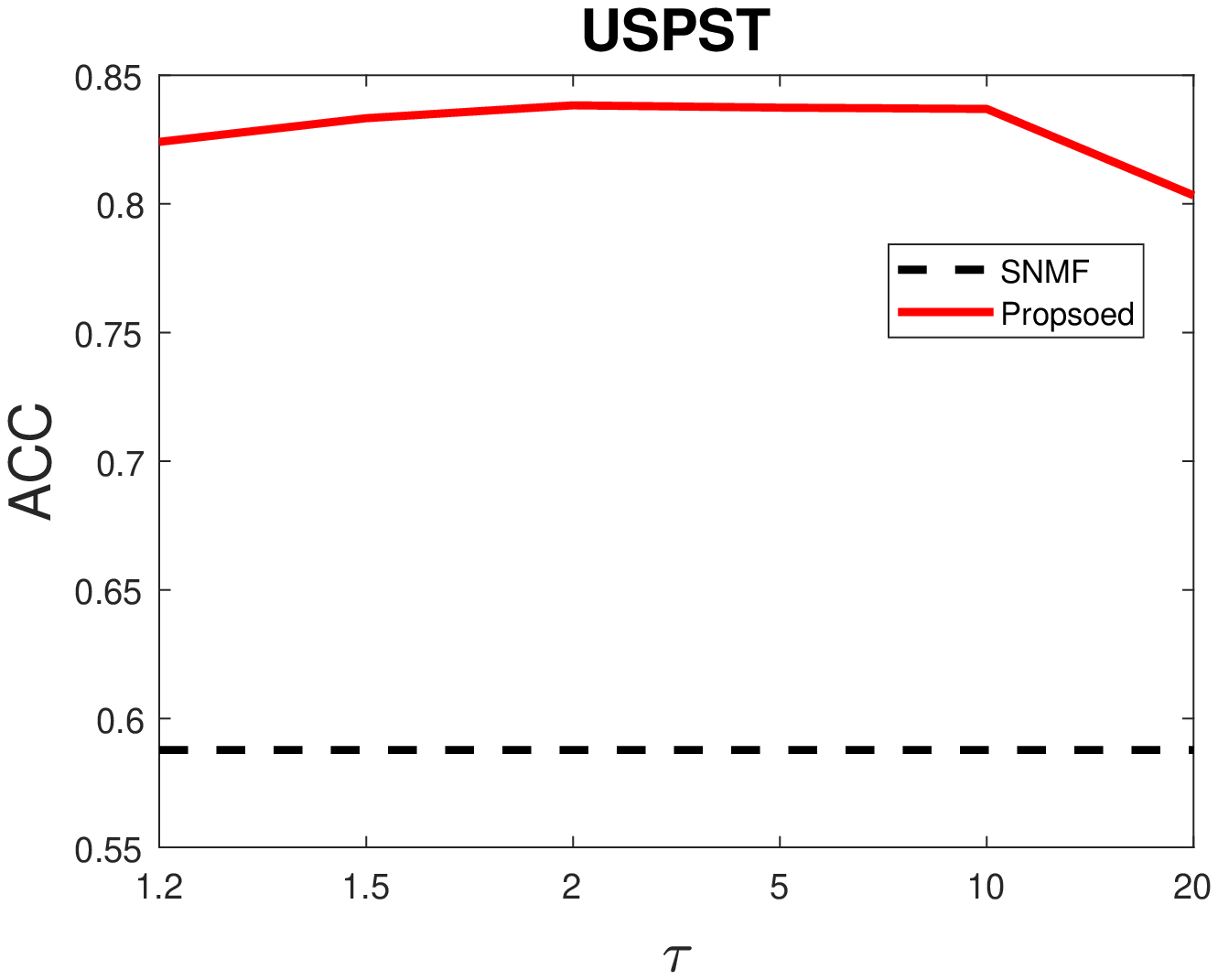,width=3.7cm}}
		%		\vspace{-0.5cm}
		%       \centerline{(a) Result 1}\medskip
	\end{minipage}
	\begin{minipage}[b]{0.195\linewidth}
		\centering
		\centerline{\epsfig{figure=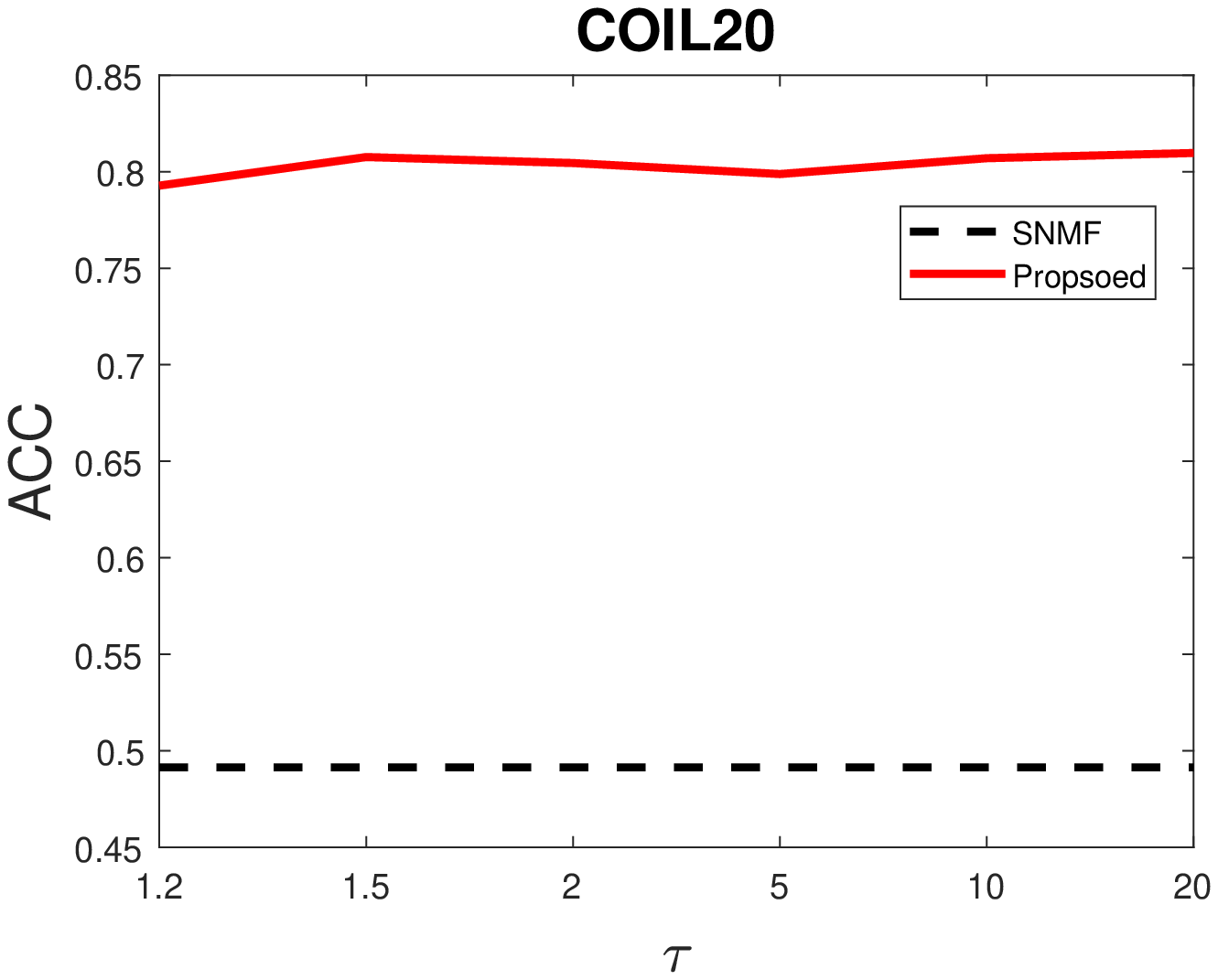,width=3.7cm}}
		%		\vspace{-0.5cm}
		%       \centerline{(a) Result 1}\medskip
	\end{minipage}
	\begin{minipage}[b]{0.195\linewidth}
		\centering
		\centerline{\epsfig{figure=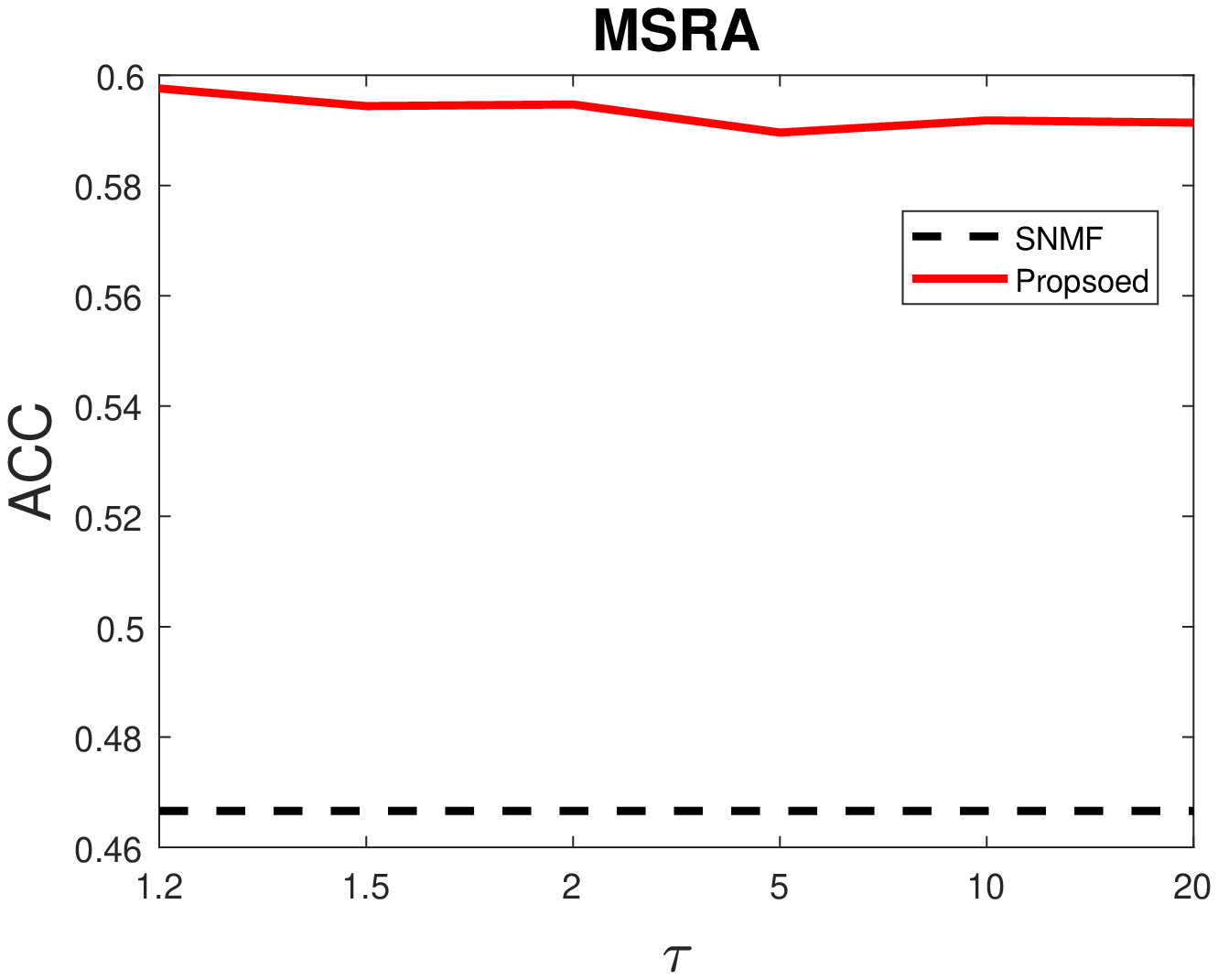,width=3.7cm}}
		%		\vspace{-0.5cm}
		%       \centerline{(a) Result 1}\medskip
	\end{minipage}
	\caption{The influence of the value of $\tau$ on the ACC of the proposed model. 
		We also plot the ACC of SNMF as a reference. }
	\label{fig:tau}
\end{figure*}

\begin{table*}[t]
	\caption{Ablation Study of the Proposed Model}%\smallskip
	\centering
	\resizebox{2\columnwidth}{!}{
		\smallskip\begin{tabular}{l| c c c c c c c c c c |c}
			\hline\hline
			\textbf{ACC}	&	MNIST	&	USPST	&	MSRA	&	SEMENION	&	IRIS	&	CHART	&	SEEDS	&	USPS	&	PALM	&	COIL20 & \textbf{Average}	\\
			SNMF	&$	0.518\searrow	$&$	0.588\searrow	$&$	0.467\searrow	$&$	0.591\searrow	$&$	0.645\searrow	$&$	0.671\searrow	$&$	0.602\searrow	$&$	0.616\searrow	$&$	0.733\searrow	$&$	0.491\searrow	$ & $0.592\searrow$ 
			 
			\\
			SOFT	&$	0.557\searrow	$&$	0.644\searrow	$&$	0.577\searrow	$&$	0.637\searrow	$&$	0.892\nearrow	$&$	0.649\searrow	$&$	0.653\searrow	$&$	0.649\searrow	$&$	0.846\searrow	$&$	0.641\searrow	$ & $0.674\searrow$ 
			\\
			w/o $\bm{\alpha}$	&$	0.637\searrow	$&$	0.817\searrow	$&$	0.596\nearrow	$&$	0.726\nearrow	$&$	0.718\searrow	$&$	0.720\searrow	$&$	0.812\searrow	$&$	0.833\searrow	$&$	0.875\searrow	$&$	0.815\nearrow	$ & $0.755\searrow$ \\
			Proposed	&$	0.680	$&$	0.838	$&$	0.595	$&$	0.720	$&$	0.886	$&$	0.869	$&$	0.881	$&$	0.847	$&$	0.881	$&$	0.805	$ & 
			0.800\\
			\hline
			\textbf{NMI}	&	MNIST	&	USPST	&	MSRA	&	SEMENION	&	IRIS	&	CHART	&	SEEDS	&	USPS	&	PALM	&	COIL20& \textbf{Average}	\\
			SNMF	&$	0.523\searrow	$&$	0.645\searrow	$&$	0.567\searrow	$&$	0.579\searrow	$&$	0.388\searrow	$&$	0.678\searrow	$&$	0.311\searrow	$&$	0.653\searrow	$&$	0.922\nearrow	$&$	0.657\searrow	$ & $0.592\searrow$ 
			\\
			SOFT	&$	0.558\searrow	$&$	0.750\searrow	$&$	0.671\searrow	$&$	0.650\searrow	$&$	0.761\searrow	$&$	0.692\searrow	$&$	0.405\searrow	$&$	0.717\searrow	$&$	0.923\nearrow	$&$	0.803\searrow	$ & $0.693\searrow$ 
			 \\
			w/o $\bm{\alpha}$	&$	0.578\searrow	$&$	0.808\nearrow	$&$	0.666\searrow	$&$	0.672\nearrow	$&$	0.582\searrow	$&$	0.682\searrow	$&$	0.559\searrow	$&$	0.787\searrow	$&$	0.916\searrow	$&$	0.853\searrow	$ & $0.710\searrow$ 
			\\
			Proposed	&$	0.606	$&$	0.805	$&$	0.697	$&$	0.667	$&$	0.769	$&$	0.828	$&$	0.668	$&$	0.794	$&$	0.917	$&$	0.857	$ & $0.761$\\
			\hline
			\textbf{PUR}	&	MNIST	&	USPST	&	MSRA	&	SEMENION	&	IRIS	&	CHART	&	SEEDS	&	USPS	&	PALM	&	COIL20& \textbf{Average}	\\
			SNMF	&$	0.569\searrow	$&$	0.667\searrow	$&$	0.502\searrow	$&$	0.639\searrow	$&$	0.664\searrow	$&$	0.707\searrow	$&$	0.616\searrow	$&$	0.674\searrow	$&$	0.793\searrow	$&$	0.532\searrow	$ & $0.636\searrow$ 
			
			\\
			SOFT	&$	0.601\searrow	$&$	0.765\searrow	$&$	0.595\searrow	$&$	0.700\searrow	$&$	0.892\nearrow	$&$	0.700\searrow	$&$	0.672\searrow	$&$	0.711\searrow	$&$	0.877\searrow	$&$	0.717\searrow	$ & $0.723\searrow$ 
		\\
			w/o $\bm{\alpha}$	&$	0.642\searrow	$&$	0.856\searrow	$&$	0.608\searrow	$&$	0.731\cdots	$&$	0.718\searrow	$&$	0.745\searrow	$&$	0.812\searrow	$&$	0.833\searrow	$&$	0.884\searrow	$&$	0.817\searrow	$ & 	$0.765\searrow$ 
			\\
			Proposed	&$	0.684	$&$	0.858	$&$	0.629	$&$	0.731	$&$	0.886	$&$	0.870	$&$	0.881	$&$	0.847	$&$	0.886	$&$	0.816	$ & $0.809$ \\
			\hline
			\textbf{ARI}	&	MNIST	&	USPST	&	MSRA	&	SEMENION	&	IRIS	&	CHART	&	SEEDS	&	USPS	&	PALM	&	COIL20& \textbf{Average}	\\
			SNMF	&$	0.376\searrow	$&$	0.488\searrow	$&$	0.337\searrow	$&$	0.435\searrow	$&$	0.346\searrow	$&$	0.557\searrow	$&$	0.279\searrow	$&$	0.505\searrow	$&$	0.723\searrow	$&$	0.386\searrow	$ & $0.443\searrow$ 
			 
			\\
			SOFT	&$	0.421\searrow	$&$	0.603\searrow	$&$	0.465\searrow	$&$	0.515\searrow	$&$	0.729\nearrow	$&$	0.573\searrow	$&$	0.388\searrow	$&$	0.594\searrow	$&$	0.822\searrow	$&$	0.573\searrow	$ & $0.568\searrow$ 
			\\
			w/o $\bm{\alpha}$	&$	0.476\searrow	$&$	0.766\searrow	$&$	0.483\searrow	$&$	0.582\nearrow	$&$	0.508\searrow	$&$	0.581\searrow	$&$	0.552\searrow	$&$	0.724\searrow	$&$	0.850\searrow	$&$	0.760\cdots	$ & $0.628\searrow$ 
			\\
			Proposed	&$	0.506	$&$	0.778	$&$	0.488	$&$	0.573	$&$	0.722	$&$	0.763	$&$	0.689	$&$	0.731	$&$	0.854	$&$	0.760	$ & $0.686$\\
			\hline
			\textbf{F1-score}	&	MNIST	&	USPST	&	MSRA	&	SEMENION	&	IRIS	&	CHART	&	SEEDS	&	USPS	&	PALM	&	COIL20& \textbf{Average}	\\
			SNMF	&$	0.447\searrow	$&$	0.546\searrow	$&$	0.400\searrow	$&$	0.497\searrow	$&$	0.579\searrow	$&$	0.641\searrow	$&$	0.531\searrow	$&$	0.560\searrow	$&$	0.726\searrow	$&$	0.421\searrow	$ & $0.535\searrow$ 
			
			\\
			SOFT	&$	0.490\searrow	$&$	0.648\searrow	$&$	0.516\searrow	$&$	0.568\searrow	$&$	0.820\nearrow	$&$	0.656\searrow	$&$	0.615\searrow	$&$	0.641\searrow	$&$	0.824\searrow	$&$	0.600\searrow	$ & $0.638\searrow$ 
		 \\
			w/o $\bm{\alpha}$	&$	0.531\searrow	$&$	0.791\searrow	$&$	0.530\searrow	$&$	0.626\nearrow	$&$	0.673\searrow	$&$	0.654\searrow	$&$	0.704\searrow	$&$	0.752\searrow	$&$	0.851\searrow	$&$	0.773\cdots	$ & 	$0.688\searrow$ 
			\\
			Proposed	&$	0.558	$&$	0.802	$&$	0.533	$&$	0.619	$&$	0.816	$&$	0.803	$&$	0.793	$&$	0.758	$&$	0.856	$&$	0.773	$ & $0.731$\\
			\hline\hline
		\end{tabular}
	}
	\label{table-abl}
\end{table*}

\subsection{Comparison of  Clustering Performance} 
Tables II-XI show the clustering performance of all the methods on the 10 datasets, where the best performance under each metric is bold, and the second best is underlined. From Tables II-XI,  we can observe that 
\begin{itemize}
	\item Our method significantly outperforms SNMF,  especially our method improves ACC up to $0.31$ on COIL20. 
%	The improvements of ACC of our method over SNMF are at least larger than $0.12$, and half of them are larger than $0.23$. On COIL20, the ACC of our method increases $0.31$ compared with that of SNMF.  
	The improvements of our method over SNMF in terms of  other metrics are also significant, e.g., the ARI increases from $0.279$ to $0.688$ on SEEDS. Moreover, the proposed method has a smaller std than SNMF on all the datasets, indicating that the proposed model is more robust to the initialization than SNMF. 
	Note that both SNMF and the proposed method adopt the same affinity matrix as input.
	%The noteworthy improvement substantiates that the proposed method is a simple yet very  effective method to enhance the traditional SNMF.
	\item Compared with the advanced graph clustering methods, the advantage of the proposed one is also significant. For example, on CHART and USPST, our method improves the ACC $0.167$ and  $0.119$, respectively, compared with the best graph clustering model. Moreover, the performance of the proposed method is also superior to that of the ensemble clustering methods. For instance, on IRIS, 
%	the ACC of the proposed model improves $0.166$ 
	our method raises the ACC value from $0.720$ to $0.886$, 
	compared with the best ensemble clustering method.
	\item 
%	Taking all the compared methods into account, our method achieves highly competitive performance with all the metrics. Specifically, 
Taking all the compared methods into account,
	the proposed method achieves the best performance under $44$ out of $50$ cases and the second best performance under 3 out of the remaining $6$ cases, suggesting the highly competitive  clustering ability of our method. 
	%$3$ second highest values in the remaining $6$ cases. 
%	When the proposed method cannot achieve the best performance, the gap with the highest one is quite  tiny, e.g., on IRIS, the NMI of our method is just $0.007$ lower than the highest one. Therefore, we can conclude that the proposed method is extremely effective in clustering. 
	\item The performance of the compared methods is usually not robust to different datasets. For example, LWGP is good at partitioning COIL20, but not on  USPST. ALS favors PALM and COIL20 more than SEMENION and CHART. vBSUM performs much better on IRIS  than MSRA and COIL20. On the contrary, our method consistently produces the best or almost best performance over these 10 datasets, validating its robustness.  
\end{itemize}

\subsection{Influence of the Iteration Number of Algorithm 1} 
We studied how the number of iterations in Algorithm 1 affects clustering performance. 
%Note that Fig. 1 uses the double Y axes to express various indicators, i.e., the left axis with the black line denotes the ACC, and the right axis with the blue line represents the proposed stopping condition, ANMI. 
As shown in  Fig. 1, the ACC value  of the proposed method increases rapidly at the first a few iterations, and then becomes  relatively stable with the increase of iteration. We can also observe that on the majority datasets, the highest ACC can be selected by the proposed stopping criterion, e.g., on IRIS and MNIST. For the cases where the highest ACC is not picked, the proposed criterion can also produce a  satisfied ACC for the proposed method. Besides, the  proposed criterion usually terminates Algorithm 1 within $5$ iterations, which can reduce the computational cost. As a summary, the proposed stopping criterion is both effective and efficient.

\subsection{Influence of the Size of the Set of Initialization}
We also investigated the influence of the size of initialization set (the value of $b$) on the proposed method. 
Fig. \ref{fig:en} shows the ACC of the proposed method with the different sizes of the initialization set on all the datasets. Due to the randomness of the initialization, those curves may fluctuate at some locations. 
However, generally, a larger size of the initialization set  leads to a higher ACC. The trend is particularly evident in COIL20, PLAM, IRIS, and CHART.  
Moreover, 20 different initialization are sufficient for the proposed method to achieve satisfactory performance. %\textcolor{red}{More analysis after seeding the real results}

%The stopping condition has two merits, first, the clustering performance is good, second, improve the efficiency.
\subsection{Influence of the Value of  $\tau$ on Performance}
Here we explored  the impact of the value of $\tau$ on the performance of our method.  
Fig. \ref{fig:tau} shows the ACC values of our method with various $\tau$\footnote{As analyzed in section III, $\tau$ should be larger than 1 to avoid selecting only one basic SNMF. Therefore, the value of $\tau$ was selected from $\{1.2, 1.5, 2, 5, 10, 20\}$. } on all the datasets, where we can find that the curves are quite flat on the majority cases, like SEEDS and COIL20. This observation  demonstrates the robustness of the proposed model against $\tau$.  On SEMENION and MNIST, $\tau$ should be neither too large or too small. Overall, the suggested value of $\tau$ is acceptable for all the datasets. 
%Moreover, we also need to emphasize that the flat curves do not indicate that $\mathbf{\alpha}$ is unimportant. 
Moreover, we also would like to  emphasize that the flat curves do not dispel the importance of $\bm{\alpha}$. 
As will be shown in the next subsection, removing $\bm{\alpha}$ will significantly degrade the clustering performance of our model. 

\subsection{Ablation Study}
In this section, we evaluated the importance of different components of the proposed method. Specifically, w/o $\bm{\alpha}$ denotes our model without the learable  weight  $\bm{\alpha}$, i.e.,
\begin{equation}
\begin{split}
&\min_{\mathbf{V}_m,\mathbf{S},}
\sum_{m=1}^b\left\|\mathbf{S}-\mathbf{V}_m\mathbf{V}_m^\mathsf{T}\right\|_F^2,~{\rm s.t.~}\mathbf{V}_m\geq 0, \forall m.
\end{split}
\end{equation}
SOFT represents the reconstruction of $\mathbf{S}$ in Eq. \eqref{learnS} is replaced with a soft manner, i.e., 
\begin{equation}
\mathbf{S}=\sum_{m=1}^b\alpha_m\mathbf{V}_m\mathbf{V}_m^\mathsf{T}.
\end{equation}
Table  \ref{table-abl} shows the clustering performance of w/o $\bm{\alpha}$ and SOFT as well as SNMF, where $\nearrow$, $\searrow$ and $\cdots$ indicate the value of the associated metric of the compared methods is larger than, smaller than and equivalent to our S$^3$NMF, respectively.  
The last column of Table  \ref{table-abl} presents the average value of each method over all the datasets. From Table \ref{table-abl} we can observe that both SOFT and w/o $\bm{\alpha}$ are superior to SNMF but inferior to our method. 
%Specifically, only in quite limited cases ($13/150$), the proposed model does not achieve the highest value. 
Especially,  according to the average values in the last column,  it is obvious that our method
outperforms SNMF, SOFT, and w/o $\bm{\alpha}$ to a large extent. This observation demonstrates that both the hard manner in the construction of $\mathbf{S}$ and the self-weighing scheme contribute to the proposed model. The advantage of the hard manner for constructing $\mathbf{S}$ over the soft manner is credited to that  the proposed model could receive stronger feedback from the clustering result to guide the affinity matrix  construction.  $\bm{\alpha}$ is important because the contribution of each base SNMF can be adaptively balanced based on its quality.

\section{Conclusion}

In this paper, by taking advantage of the characteristic that SNMF is sensitive to initialization, we have presented novel self-supervised SNMF (S$^3$NMF) for clustering. Without relying on any additional information except multiple random nonnegative initialization matrices, our S$^3$NMF significantly
outperforms the state-of-the-art graph clustering methods and ensemble clustering methods.
We formulated S$^3$NMF as a self-weighted nonnegative constrained optimization problem,  and proposed  an alternating iterative optimization method to solve it and proved the convergence theoretically. 
Moreover, we have presented a stopping criterion to terminate the optimization process, where its effectiveness is validated by the experiments. 
Additionally, our model is insensitive to its hyper-parameters, and the recommended values for the 
hyper-parameters empirically work very well, which validate its practicability.

\appendices
%\section{Proof of the First Zonklar Equation}
\section{}
In the appendix, we prove that Eq. \eqref{auxiliary-V} is a valid auxiliary function of Eq. \eqref{V-sub}. The objective function of Eq. \eqref{V-sub} can be expanded as
\begin{equation}
\begin{split}
&(\mathbf{\alpha}_m)^\tau\left\|\mathbf{S}-\mathbf{V}_m\mathbf{V}_m^\mathsf{T}\right\|_F^2\\
&=(\mathbf{\alpha}_m)^\tau {\rm Tr}\left(\mathbf{SS}^\mathsf{T}-2\mathbf{SV}_m\mathbf{V}_m^\mathsf{T}+\mathbf{V}_m\mathbf{V}_m^\mathsf{T}\mathbf{V}_m\mathbf{V}_m^\mathsf{T}\right).
\end{split}
\label{app-v-loss}
\end{equation}  
%\begin{equation}
%\begin{split}
%&\mathbf{\alpha}_i^\tau\left\|\mathbf{S}-\mathbf{V}_i\mathbf{V}_i^\mathsf{T}\right\|_F^2\\
%&=\mathbf{\alpha}_i^\tau {\rm Tr}\left(\mathbf{SS}^\mathsf{T}-2\mathbf{SV}_i\mathbf{V}_i^\mathsf{T}+\mathbf{V}_i\mathbf{V}_i^\mathsf{T}\mathbf{V}_i\mathbf{V}_i^\mathsf{T}\right)\\
%&\propto {\rm Tr}\left(-2\mathbf{SV}_i\mathbf{V}_i^\mathsf{T}+\mathbf{V}_i\mathbf{V}_i^\mathsf{T}\mathbf{V}_i\mathbf{V}_i^\mathsf{T}\right)
%\end{split}
%\end{equation}
The first term $\mathbf{S}\mathbf{S}^\mathsf{T}$ in the brackets is a constant. 
For the second term, we can expand it as 
\begin{equation}
\begin{split}
&{\rm Tr}\left(\mathbf{SV}_m\mathbf{V}_m^\mathsf{T}\right)=\sum_i\left(\mathbf{SV}_m\mathbf{V}_m^\mathsf{T}\right)_{ii}=\sum_{ijk}\mathbf{S}_{ik}\mathbf{V}_{m_{ij}}\mathbf{V}_{m_{kj}}.
\end{split}
\end{equation}
Since for any positive value $x$, we have $x>1+{\rm log} (x)$. Let $\frac{\mathbf{V}_{m_{ij}}\mathbf{V}_{m_{kj}}}{\mathbf{V}^t_{m_{ij}}\mathbf{V}^t_{m_{kj}}}=x$, we have:
\begin{equation}
{\rm Tr}\left(\mathbf{SV}_m\mathbf{V}_m^\mathsf{T}\right)\geq\sum_{ijk}\mathbf{S}_{ik}\mathbf{V}^t_{m_{ij}}\mathbf{V}^t_{m_{kj}}\left(1+{\rm log}\frac{\mathbf{V}_{m_{ij}}\mathbf{V}_{m_{kj}}}{\mathbf{V}^t_{m_{ij}}\mathbf{V}^t_{m_{kj}}}\right).
\label{aux-2}
\end{equation}
%\begin{equation}
%\begin{split}
%&{\rm Tr}\left(\mathbf{SV}_m\mathbf{V}_m^\mathsf{T}\right)=\sum_i\left(\mathbf{SV}_m\mathbf{V}_m^\mathsf{T}\right)_{ii}\\
%&=\sum_{ik}\mathbf{S}_{ik}\left(\mathbf{V}_m\mathbf{V}_m^\mathsf{T}\right)_{ki}=\sum_{ijk}\mathbf{S}_{ik}\mathbf{V}_{m_{ij}}\mathbf{V}_{m_{kj}}
%\end{split}
%\end{equation}
For the third term,
 we have 
 \begin{equation}
 \begin{split}
 &{\rm Tr}\left(\mathbf{V}_m\mathbf{V}_m^\mathsf{T}\mathbf{V}_m\mathbf{V}_m^\mathsf{T}\right)=\sum_i\left(\mathbf{V}_m\mathbf{V}_m^\mathsf{T}\mathbf{V}_m\mathbf{V}_m^\mathsf{T}\right)_{ii}=\\
 &\sum_{ik}\left(\mathbf{V}_m\mathbf{V}_m^\mathsf{T}\right)_{ik}\left(\mathbf{V}_m\mathbf{V}_m^\mathsf{T}\right)_{ki}=\sum_{ijk}\left(\mathbf{V}_m\mathbf{V}_m^\mathsf{T}\right)_{ik}\mathbf{V}_{m_{kj}}\mathbf{V}_{m_{ij}}
 \end{split}
 \end{equation}
 and 
 \begin{equation}
 {\rm Tr}\left(\mathbf{V}_m\mathbf{V}_m^\mathsf{T}\mathbf{V}_m\mathbf{V}_m^\mathsf{T}\right)\leq\sum_{ijk}\left(\mathbf{V}^t_m\mathbf{V}_m^{t^\mathsf{T}}\right)_{ik}\mathbf{V}_{m_{kj}}\frac{(\mathbf{V}_{m_{ij}})^4}{(\mathbf{V}_{m_{ij}}^{t})^3},
 \label{aux-4}
 \end{equation}
where the proof of the   inequality in Eq. \eqref{aux-4} can be found in \cite{9072553}. 
% 
%\begin{equation}
%\begin{split}
%{\rm Tr}\left(\mathbf{V}_m\mathbf{V}_m^\mathsf{T}\mathbf{V}_m\mathbf{V}_m^\mathsf{T}\right)=\sum_i\left(\mathbf{V}_m\mathbf{V}_m^\mathsf{T}\mathbf{V}_m\mathbf{V}_m^\mathsf{T}\right)_{ii}\\
%%=\sum_{ik}\left(\mathbf{V}_m\mathbf{V}_m^\mathsf{T}\right)_{ik}\left(\mathbf{V}_m\mathbf{V}_m^\mathsf{T}\right)_{ki}\\
%=\sum_{ijk}\left(\mathbf{V}_m\mathbf{V}_m^\mathsf{T}\right)_{ik}\mathbf{V}_{m_{kj}}\mathbf{V}_{m_{ij}}. 
%\end{split}
%\end{equation}
% , and Eq. (25) can be found in Jia's paper.
%
Taking Eq. \eqref{aux-2} and Eq. \eqref{aux-4} into account, 
we have $g(\mathbf{V}_{m})\geq(\mathbf{\alpha}_m)^\tau\left\|\mathbf{S}-\mathbf{V}_m^t\mathbf{V}_m^{t^\mathsf{T}}\right\|_F^2$. Moreover, when $\mathbf{V}_{m}=\mathbf{V}^t_{m}$, we have $g(\mathbf{V}^t_{m})=(\mathbf{\alpha}_m)^\tau\left\|\mathbf{S}-\mathbf{V}_m^t\mathbf{V}_m^{t^\mathsf{T}}\right\|_F^2$. The two conditions in \textbf{Definition 1} are satisfied, and thus  $g(\mathbf{V}_{m})$ is an auxiliary function of the $\mathbf{V}_{m}$-subproblem. 

%Since for any positive value $x$, we have x>1+log (x), we have Eq. 28.

%\input{authorbibliography}
\bibliographystyle{IEEEtran}
\bibliography{IEEEabrv,bib}

\end{document}